\documentclass[10pt,journal,compsoc]{IEEEtran}
\ifCLASSOPTIONcompsoc
  \usepackage[nocompress]{cite}
\else
  \usepackage{cite}
\fi
\usepackage{hyperref} 
\hypersetup{
    colorlinks=true,
    linkcolor=red,
}
\usepackage{amssymb}
\usepackage{bbding}
\usepackage{amsmath}
\usepackage{booktabs}

\ifCLASSINFOpdf
   \usepackage[pdftex]{graphicx}
\else
\fi

\usepackage{stfloats}
\usepackage{color}
\usepackage[dvipsnames]{xcolor}

\usepackage{multirow}

\usepackage{caption}
\usepackage{subcaption}
\usepackage{amsmath}
\usepackage{cprotect}

\usepackage{amssymb}
\usepackage{pifont}

\begin{document}
%
\title{
Visual Instruction Tuning towards General-Purpose Multimodal Model: A Survey
}

\author{
Jiaxing~Huang$^\dagger$, Jingyi~Zhang$^\dagger$,
Kai Jiang,
Han Qiu
and~Shijian~Lu$^*$
\IEEEcompsocitemizethanks{\IEEEcompsocthanksitem All authors are with the School of Computer Science and Engineering, Nanyang Technological University, Singapore.\protect
\IEEEcompsocthanksitem $\dagger$ denotes equal contribution; $*$ denotes corresponding author.}
}

%
%

\markboth{Journal of \LaTeX\ Class Files,
December~2023}%
{Shell \MakeLowercase{\textit{et al.}}: Bare Demo of IEEEtran.cls for Computer Society Journals}
%



\IEEEtitleabstractindextext{%
\begin{abstract}
Traditional computer vision generally solves each single task independently by a dedicated model with the task instruction implicitly designed in the model architecture, arising two limitations: 
(1) it leads to task-specific models, which require multiple models for different tasks and restrict the potential synergies from diverse tasks;
(2) it leads to a pre-defined and fixed model interface that has limited interactivity and adaptability in following user' task instructions.
To address them, Visual Instruction Tuning (VIT) has been intensively studied recently, which finetunes a large vision model with language as task instructions, aiming to learn from a wide range of vision tasks described by language instructions a general-purpose multimodal model that can follow arbitrary instructions and thus solve arbitrary tasks specified by the user.
This work aims to provide a systematic review of visual instruction tuning, covering (1) the background that presents computer vision task paradigms and the development of VIT;
(2) the foundations of VIT that introduce commonly used network architectures, visual instruction tuning frameworks and objectives, and evaluation setups and tasks;
(3) the commonly used datasets in visual instruction tuning and evaluation;
(4) the review of existing VIT methods that categorizes them with a taxonomy according to both the studied vision task and the method design and highlights the major contributions, strengths, and shortcomings of them;
(5) the comparison and discussion of VIT methods over various instruction-following benchmarks;
(6) several challenges, open directions and possible future works in visual instruction tuning research.

\end{abstract}

}

\maketitle

\IEEEdisplaynontitleabstractindextext

\IEEEpeerreviewmaketitle

\IEEEraisesectionheading{\section{Introduction}\label{sec:introduction}}

Computer vision has been a long-standing challenge in artificial intelligence, which aims to enable computers, machines or systems to perceive, analyze, comprehend and interact with the visual world like human beings~\cite{voulodimos2018deep,zhang2020empowering}.
With the development of deep neural networks~\cite{krizhevsky2017imagenet, simonyan2014very,he2016deep}, computer vision research has achieved great successes in a spectrum of tasks, such as discriminative vision tasks (e.g., image classification and segmentation, object detection, etc.) and generative vision tasks (e.g., image generation, image editing, etc.).

Nevertheless, in this line of research, each vision task is generally solved independently by a dedicated vision model, where the task instruction is implicitly considered and designed in the model architecture, such as segmentation heads for mask prediction, detection heads for box prediction, image captioning heads for descriptive text generation and image generation decoder for generating RGB images.
This gives rise to two inherent limitations: (1) it leads to vision models that are task-specific, which requires training and using multiple models for different tasks and restrict the potential synergies from diverse tasks; (2) it results in vision models that typically have a pre-defined and fixed interface, leading to limited interactivity and adaptability in following users' task instructions.

\begin{figure}[!t]
    \centering
    \includegraphics[width=0.42\textwidth]{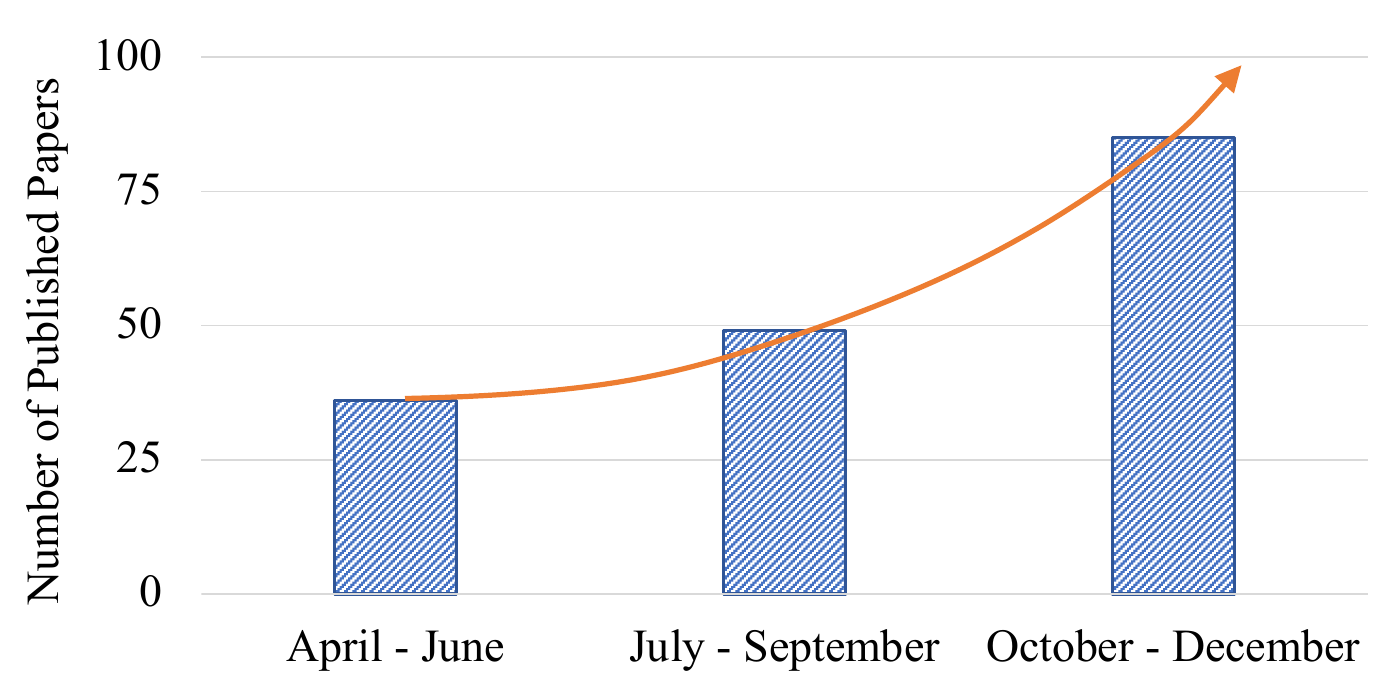}
    \caption{The number of publications on visual instruction tuning in 2023. It has shown exponential growth since the pioneering work LLaVA in April 2023. Data collected from Google Scholar. 
    }
    \label{fig:trend}
\end{figure}

Recently, instruction tuning has demonstrated great effectiveness in fine-tuning large language models (LLMs) towards general-purpose LLMs.
In instruction tuning, natural languages are used to explicitly represent various task instructions and guide the end-to-end trainable model to understand and switch to the task of interest.
In this way, the model can be fine-tuned with a broad range of tasks described by natural language instructions, ultimately leading to a general-purpose model that can follow arbitrary instructions and solve arbitrary tasks specified by the user~\cite{zheng2023judging,OpenAI2020ChatGPT,openai2023gpt4}.

Inspired by the success in natural language processing, visual instruction tuning has been proposed, which fine-tunes large vision models with language as task instructions, aiming to build a general-purpose multimodal model (or called general-purpose vision-language model).
Specifically, visual instruction tuning constructs a universal interface that takes both visual and language inputs, where the language input works as task instructions which guide the model to understand the task of interest, process the visual input accordingly and return the expected output.
With this universal interface, the model can be fine-tuned with a wide of vision tasks using visual instruction tuning data (i.e., a triplet of data consisting of visual input, language instruction input and the corresponding output), resulting in a general-purpose multimodal model that accepts arbitrary language instruction inputs and visual inputs and can thus solve arbitrary vision tasks.
For example, given a natural image as the visual input, the output of the general-purpose multimodal model could be a detailed image description, a set of bounding boxes, or a modified image if the language instruction input asks to ``describe the image"", ``locate objects in the image'', or ``modify the style of the image''.

The benefits of visual instruction tuning are threefold: (1) it constructs a universal vision task interface with language as task instructions, which allows the model to learn and solve a wide range of vision tasks, benefiting from the synergies from diverse tasks;
(2) it enables the model to accept arbitrary task instructions from the user, ultimately forming an intelligent model with strong interactivity and adaptability in following the user's intent;
(3) it is computationally efficient as it can leverage the off-the-shelf pre-trained large vision model and large language model, and combine and fine-tune them to ultimately construct a general-purpose multimodal model.

Despite the significant interest in visual instruction tuning for constructing a general-purpose multimodal model, as evidenced by the considerable number of recent papers illustrated as illustrated in Figure~\ref{fig:trend}, the research community is short of a comprehensive survey that can help sort out existing visual instruction tuning methods, the facing challenges, as well as future research directions.

Despite the significant interest in visual instruction tuning for constructing a general-purpose multimodal model, as evident from the numerous recent publications as illustrated in Figure~\ref{fig:trend}, the research community lacks a systematic survey that can help comprehensively organize current visual instruction tuning approaches, the existing research challenges and potential research directions for future studies.
We strive to address this void via conducting a comprehensive survey of visual instruction tuning studies over a diverse range of vision tasks, ranging from discriminative image tasks (e.g., image classification and segmentation) to generative image tasks (e.g., image generation and editing), complex image reasoning tasks (e.g., visual question answering and visual assistant), video tasks, medical vision tasks, 3D vision tasks, etc.
The survey is performed from different perspectives, ranging from background to foundations, datasets, methodology, benchmarks, and current research challenges and open research directions.
We hope this effort will offer a comprehensive overview on what accomplishments we have achieved, what challenges we currently faced, and what we could further achieved in visual instruction tuning research.

We summarizes the main contributions of this work in three aspects.
\textit{First}, it provides a systematic review of visual instruction tuning.
We develop a taxonomy according to both the studied vision task and the method design, and highlight the major contributions, strengths, and shortcomings of existing visual instruction tuning methods.
Unlike other literature reviews that primarily concentrate on the NLP filed or delve into vision-language pre-training, our survey centers on the newly emerging research direction of visual instruction tuning, and systematically organizes the recent methods according to the investigated vision task and the instruction tuning design, offering a comprehensive overview of this promising research direction.
\textit{Second}, it investigates and analyzes the up-to-date advancements of visual instruction tuning, comprising a thorough benchmarking and discussion of existing methods over various instruction-following evaluation datasets.
\textit{Third}, it identifies and discusses several challenges, along with potential directions for future studies in visual instruction tuning research.

The remaining sections of this work are organized as the follows.
Section \ref{Background} introduces the task paradigms in computer vision, the development of visual instruction tuning and several relevant surveys. 
Section \ref{Foundation} investigates the foundations of visual instruction tuning, encompassing commonly used network architectures, visual instruction tuning frameworks and objectives, and evaluation setups and tasks for instruction-tuned general-purpose multimodal models. 
Section \ref{Dataset} provides an overview of widely adopted datasets in visual instruction tuning and the evaluation of instruction-tuned models. 
Section \ref{Sec.VLP} categorizes and reviews various visual instruction tuning methods.

\begin{figure*}[t]
    \centering
    \includegraphics[width=0.9\textwidth]{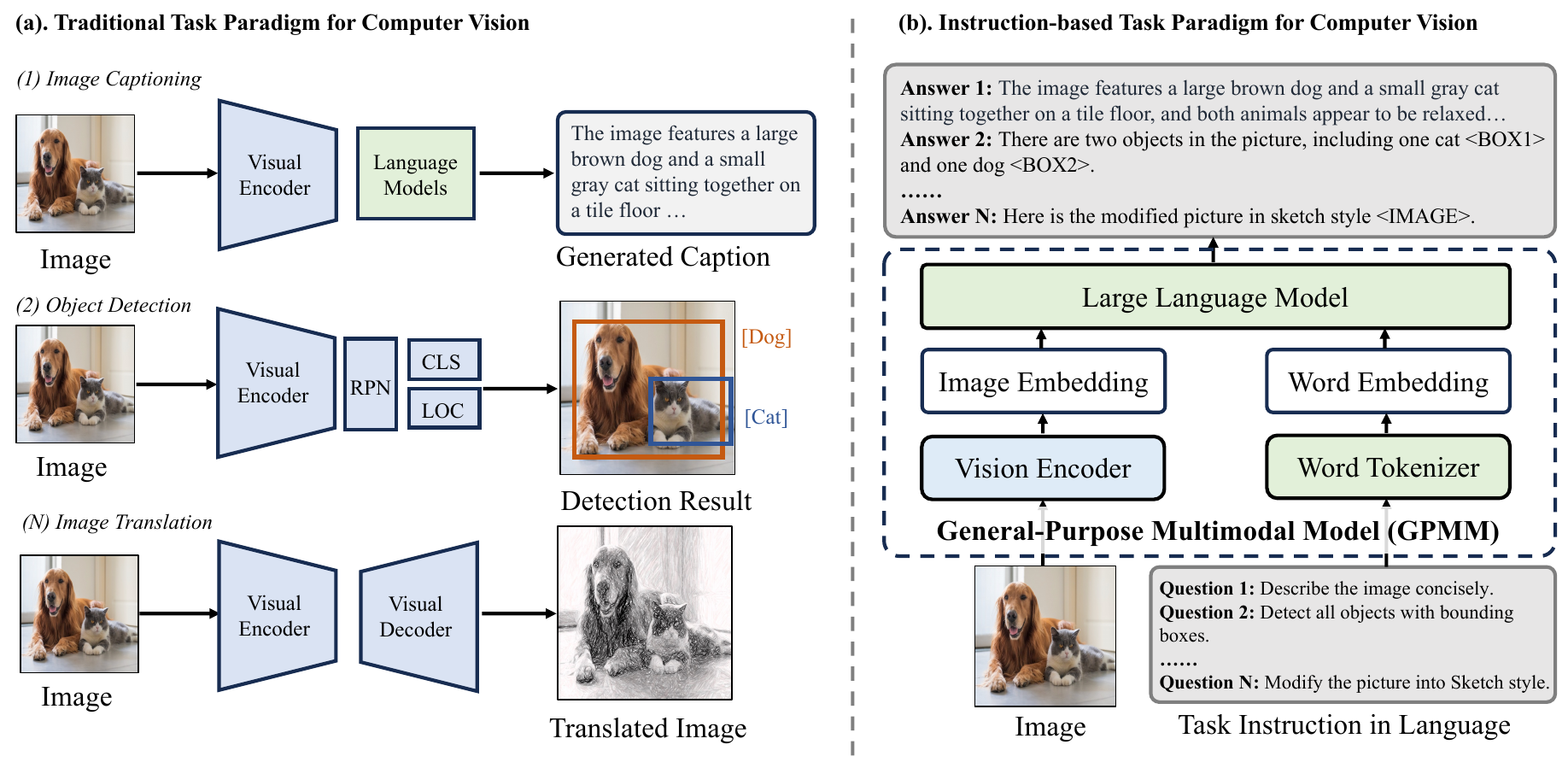}
    \caption{
    Illustrations of traditional task paradigm for computer vision in \textbf{(a)} and instruction-based task paradigm for computer vision in \textbf{(b)}. Compared with the paradigm in \textbf{(a)} that solves each single task independently by a dedicated model with task instruction implicitly designed in the model architecture, the new task paradigm with visual instruction tuning enables a general-purpose multimodal model that can follow arbitrary instructions and thus solve arbitrary tasks specified by the user. 
    }
    \label{fig.intro}
\end{figure*}

\section{Background}\label{Background}

In this section, we present the development of computer vision task paradigm and how it evolves from ``traditional task paradigm'' towards the new ``instruction-based task paradigm".
In addition, we also summarize the development of visual instruction tuning.

\subsection{Task Paradigms for Computer Vision}

The development of computer vision task paradigm can be roughly categorized into two stages: (1) the ``traditional task paradigm'' characterized by a pre-defined and fixed task interface, and (2) the ``instruction-based task paradigm" featuring with an interactive, adaptive and flexible instruction-following task interface.
Subsequently, we delve into a detailed introduction, comparison, and analysis of these two task paradigms.

\subsubsection{Traditional Task Paradigm for Computer Vision}

In traditional computer vision task paradigm, each vision task is generally solved independently by a dedicated vision model, where the task instruction is implicitly considered and designed in the model architecture.
Specifically, upon a feature extraction backbone like ResNet or ViT, traditional computer vision task paradigm generally achieves different vision tasks by designing various task-specific prediction heads, where each prediction head takes the extracted features as input and generates outputs with a pre-defined and fixed format for the given task.
For example, semantic segmentation is generally achieved by a segmentation head that takes image features as input and returns a segmentation mask in a pre-defined format, i.e., $\text{Height} \times \text{Width} \times \text{Number of Categories}$. Object detection is typically accomplished through a detection head that predicts based on the input image features a set of bounding boxes in the pre-defined format, $\{N, x1, y1, x2, y2\}$ where the first term denotes the number of predicted boxes and the last four terms stand for box coordinates.
Image generation is commonly achieved via an image decoding head, which decodes image features into an image in the RGB format.

In summary, traditional computer vision task paradigm implicitly consider the task instruction in the model design.  Therefore, this paradigm generally solves each vision task independently by a dedicated vision model, resulting in that most existing studies in this line of paradigm focus on developing effective model architectures for each of various vision tasks respectively.

As a result, traditional computer vision task paradigm often suffers from two inherent limitations, including (1) it leads to vision models that are task-specific, which requires training and using multiple models for different tasks and restrict the potential synergies from diverse tasks, and (2) it results in vision models that typically have a pre-defined and fixed task interface, leading to limited interactivity and adaptability in following users’ task instructions, as shown in Figures~\ref{fig.intro}.

\subsubsection{Instruction-based Task Paradigm for Computer Vision}

Driven by the successes in natural language processing, a new instruction-based task paradigm has been proposed, which introduces visual instruction tuning that fine-tunes large vision models with language as task instructions, ultimately building a general-purpose multimodal model (or called general-purpose vision-language model), as shown in Fig. 2.
In visual instruction tuning, it first constructs a universal interface that takes both visual and language inputs, where the language input works as task instructions which guide the model to understand the task of interest, process the visual input accordingly and return the expected output.
With such a universal interface, the model can learn a wide of vision tasks described by natural language instructions, ultimately forming a general-purpose multimodal model that accepts arbitrary language instruction inputs and visual inputs and can thus solve arbitrary vision tasks.

Compared with the traditional computer vision task paradigm that considers and designs the task instruction implicitly in the model architecture, this new paradigm explicitly represent various vision task instructions in natural languages, enabling the model to understand and learn a wide range of vision tasks and ultimately can accept arbitrary language instruction inputs and visual inputs and solve arbitrary vision tasks.

\subsection{Development of Visual Instruction Tuning}

Visual instruction tuning studies have made great progresses since the pioneer work of LLaVA.
We summarize the development of visual instruction tuning from three aspects
: (1) \textit{Task Instructions: from ``unilingual instructions'' to ``multilingual instructions''.} 
(2) \textit{Visual inputs: from ``a single type of visual input'' to ``multiple types of visual input''.} 
(3) \textit{Task difficulty: from simple to complex tasks.}

\begin{figure}[ht]
    \centering
    \includegraphics[width=0.46\textwidth]{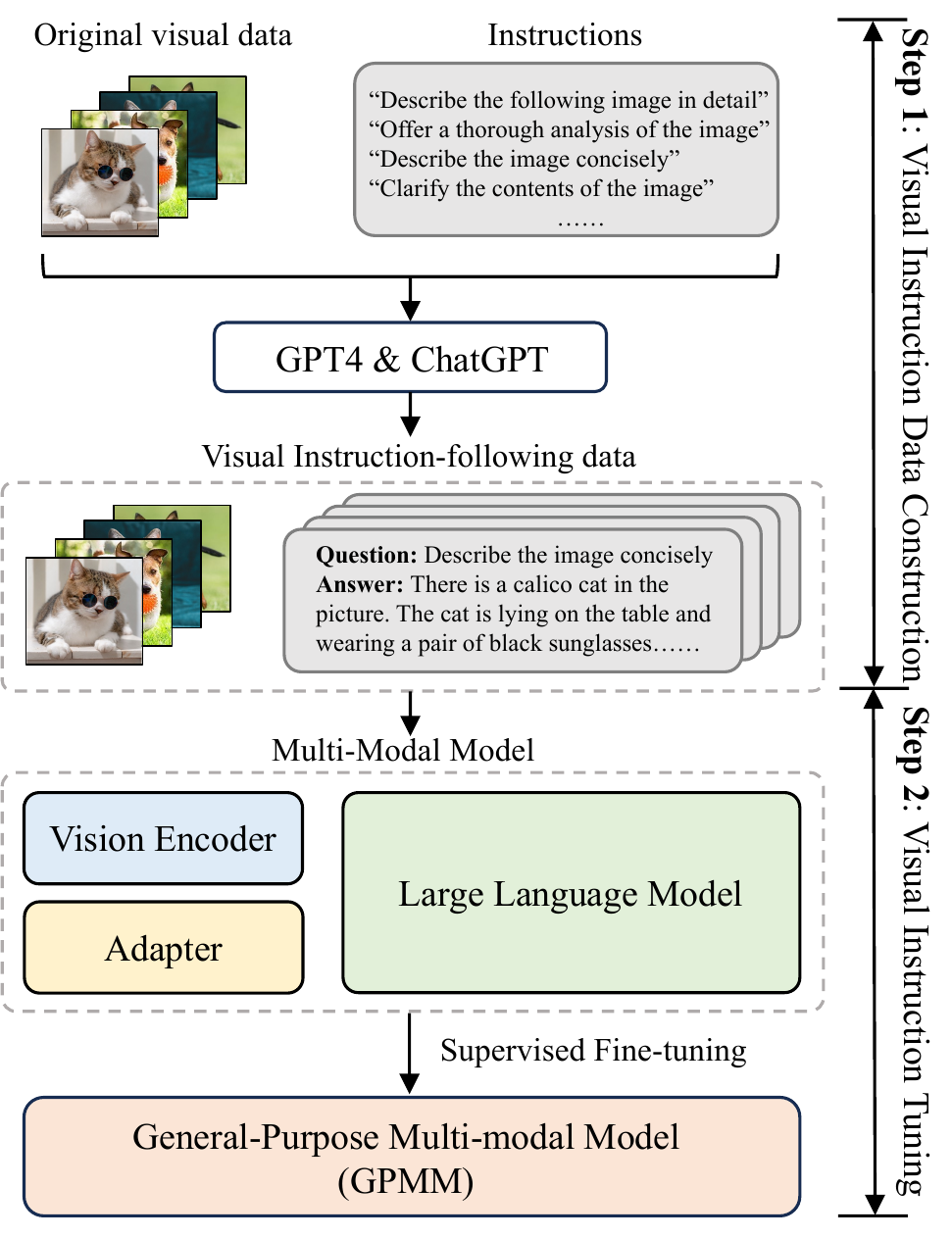}
    \caption{Pipeline of visual instruction tuning.
    }
    \label{fig:pipeline}
\end{figure}

\section{Visual Instruction Tuning Foundations}\label{Foundation}

Visual instruction tuning~\cite{liu2023visual} aims to fine-tune large vision models with visual instruction-following data, targeting general-purpose multimodal model (GPMM).
The pipeline of visual instruction tuning generally consists of two stages, i.e., visual instruction-following data construction and visual instruction tuning as illustrated in Figure~\ref{fig:pipeline}.
This section introduces the foundation of visual instruction tuning, including common ways for constructing visual instruction-following data, network architectures for encoding image and text data, visual instruction-tuning framework, objective and downstream tasks for evaluations.

\begin{figure}[ht]
    \centering
    \includegraphics[width=0.48\textwidth]{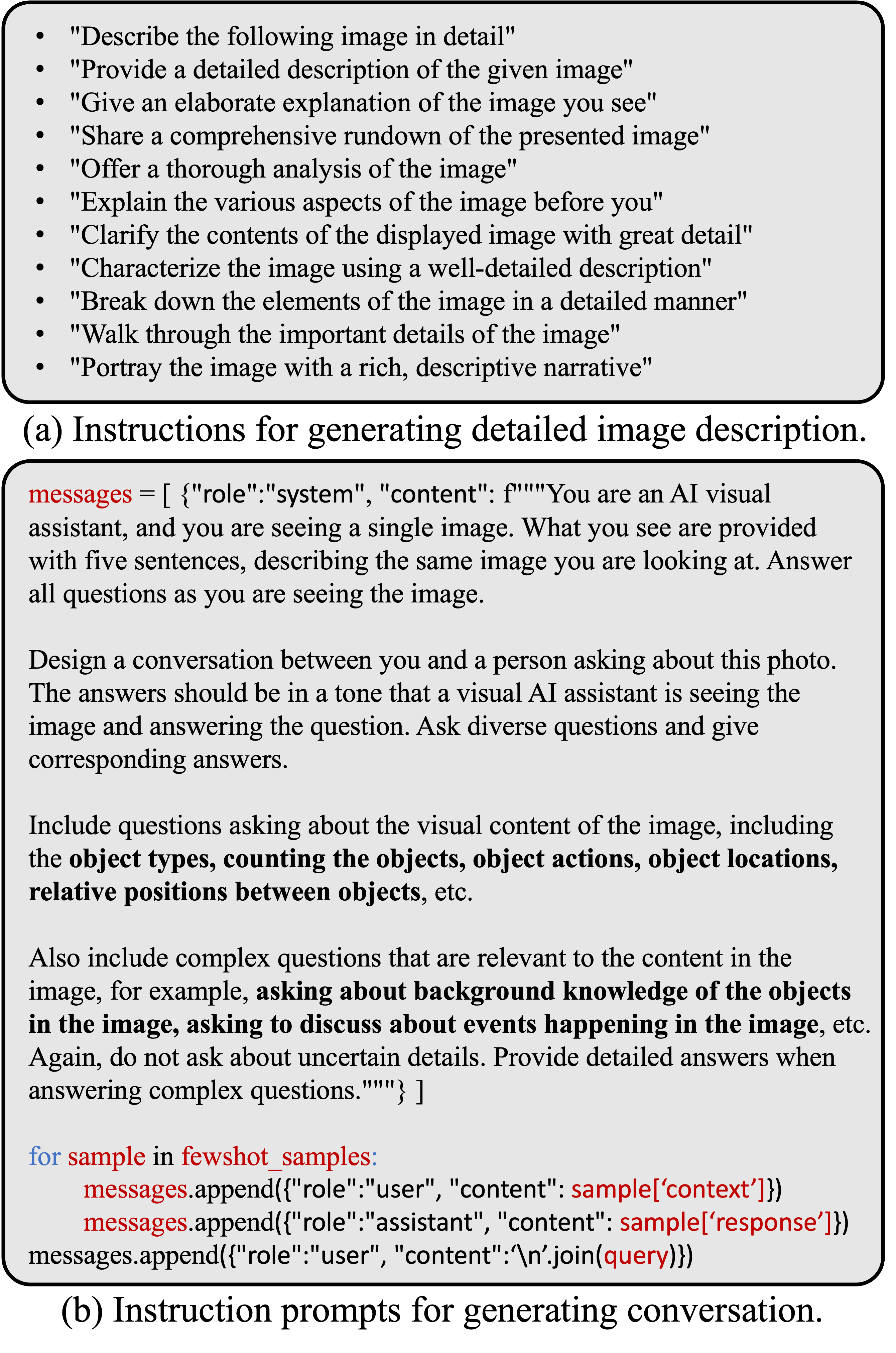}
    \caption{Instructions used in LLaVA-Instruct-158k~\cite{liu2023visual}. The content is from LLaVA~\cite{liu2023visual}.
    }
    \label{fig:instructions}
\end{figure}

\subsection{Visual Instruction-Following Data Construction}

Visual instruction-following data typically have the format of \{\verb|Instruction|, \verb|Input|, \verb|Output|\}, where \verb|Instruction| denotes instruction questions, \verb|Input| denotes input image and text pairs (i.e., \verb|Input| = \{\verb|Image|, \verb|Text|\}) and \verb|Output| denotes the response following the given instruction.
Visual instruction-following data is generally expanded from public multimodal data, such as image-text pairs~\cite{radford2021learning,schuhmann2021laion},  augmented via the application of large language models~\cite{OpenAI2020ChatGPT,openai2023gpt4}.
Specifically, given an image and its associated text \{\verb|Image|, \verb|Text|\}, several \verb|Instruction| questions are created aimed at guiding the model to describe the image's content, as illustrated in Figure~\ref{fig:instructions}.
The accumulation of such instructions is generally achieved through two primary methods: first, through manual composition~\cite{liu2023visual}; and second, by employing large language models to generate instructions based on a set of initial seed prompts~\cite{li2023mimicit}.
Then, the created instructions are fed to LLMs with the image-text pair to obtain the visual instruction-following data:

\begin{equation}
\begin{split}
&\textbf{Human:} \ \texttt{Instruction}, \ \texttt{Image <STOP>} \ \texttt{\textbackslash n} \\ 
&\textbf{Assistant:} \ \texttt{Text <STOP> \ \textbackslash n}.
\end{split}
\end{equation}

To enhance the diversity and improve the quality of both instructions and responses, recent studies have focused on two strategies: firstly, integrating additional contextual information, such as location data and bounding boxes, to facilitate detailed image comprehension; secondly, designing multiple types of instruction-following data such as single-turn descriptions and multi-turn conversations. Specifically, single-turn descriptions are typically generated by prompting large language models (LLMs) with a series of questions as illustrated in Figure~\ref{fig:instructions} (a). 
Different from the single-turn descriptions, multi-turn conversations require \textbf{Human} keep asking questions about the given image such as the object category, object location and object actions and \textbf{Assistant} answers the questions over several iterations as in Figure~\ref{fig:instructions} (b), where fine-tuning model with multi-turn conversations largely equips the model with strong chat capability.

\subsection{Network Architectures}

Visual instruction tuning utilizes a multimodal model to extract features from image and text components in visual instruction-following data. This model generally includes a vision encoder and a large language model as its core components. The section introduces the deep neural networks that are commonly employed in the field of visual instruction tuning.

\subsubsection{Architectures for Vision Learning}

Transformers have gained considerable attention in vision learning due to their effectiveness and versatility. 
Vision Transformer (ViT) is commonly employed for image feature extraction, employing a sequence of Transformer blocks, each consisting of a multi-head self-attention layer and a feed-forward network. 
In practical application, different pre-trained versions of ViT are utilized. For instance, CLIP-pre-trained ViT is used for broad image understanding~\cite{liu2023visual}, while SAM-pre-trained ViT is favored for more detailed, fine-grained image analysis~\cite{zhao2023bubogpt}.

In video feature learning, ViT is extended with additional temporal encoders to effectively model time-related information. For example, Valley~\cite{luo2023valley} introduces a temporal modeling component to capture the dynamic aspects of input videos.

For 3D image feature learning, as in the case with PointCloud data, specialized models like Point-BERT~\cite{yu2021point} and PointNet~\cite{qi2017pointnet} are employed. These models are designed to effectively extract features from PointCloud data, facilitating a deeper understanding of 3D spaces.

\subsubsection{Architectures for Language Learning}

For text feature learning, transformer-based large language models (LLMs) are prevalent. 
Specifically, the Transformer~\cite{vaswani2017attention} adopts an encoder-decoder architecture. The encoder comprises 6 blocks, each incorporating a multi-head self-attention layer and a multi-layer perceptron (MLP). Similarly, the decoder consists of 6 blocks, each including a multi-head attention layer, a masked multi-head layer, and an MLP.
Building upon the standard Transformer architecture, LLaMA~\cite{touvron2023llama} has emerged as a prominent choice for text feature extraction due to its proficiency across a range of language tasks.
Based on LLaMA~\cite{touvron2023llama}, several instruction-tuned LLMs, such as Vicuna~\cite{zheng2023judging} and Guanaco~\cite{dettmers2023qlora}, are also leveraged for extracting text features.

\subsubsection{Architectures for Audio Learning}

For extracting audio features, transformer-based architecture has been adopted. For example, Whisper~\cite{radford2023robust}, which is a general-purpose speech recognition model, has been adopted for learning audio features.

\begin{figure}[ht]
    \centering
    \includegraphics[width=0.42\textwidth]{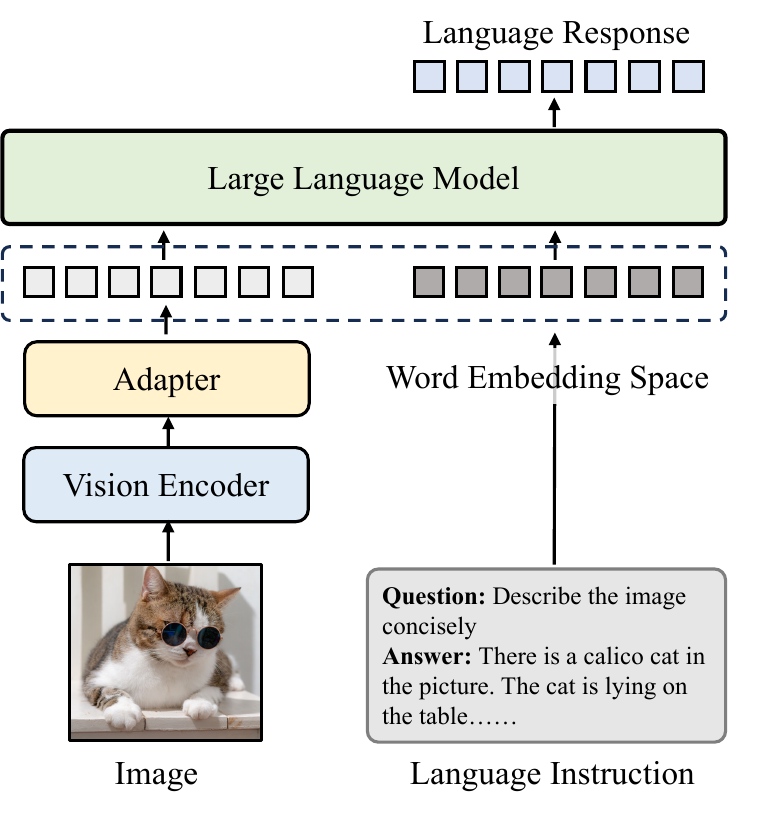}
    \caption{Illustration of visual instruction tuning framework.}
    \label{fig:framework}
\end{figure}

\subsection{Visual Instruction Tuning Framework}

The widely-adopted framework for visual instruction tuning is illustrated as in Figure~\ref{fig:framework}, which generally consists of a vision encoder, a large language model (LLM) and a adapter.
In this framework, the vision encoder is adopted for extracting features from images. The adapter then serves as a bridge, translating these image features into the word embedding space, thereby facilitating the LLM's interpretation of the vision encoder's outputs. The adapter is often designed to be lightweight and cost-effective, such as a few linear layers~\cite{liu2023visual}, to ensure efficient multimodal integration. Subsequently, the LLM processes the combined text and image embeddings to generate the expected language response.

\subsection{Visual Instruction Tuning Objective}
\label{sec:pre_obj}

Given the constructed visual instruction-following data, the multimodal model 
is fine-tuned in a full-supervised manner.
Specifically, the multimodal model  
is trained to predicted each token in the output sequentially based on the instruction and input image.

\subsection{Evaluation Setups and Tasks}\label{sec.eval}

In this section, we present commonly used setups and tasks in general-purpose multimodal model evaluation. 
The setups include \textit{human evaluation}, \textit{GPT-4 evaluation} and \textit{traditional quantitative evaluation} and the tasks used for \textit{traditional quantitative evaluation} include discriminative tasks (e.g., image classification, object detection), generative tasks (e.g., image generation) and complex image reasoning tasks (e.g., VQA, image captioning and visual assistant).

\subsubsection{Human Evaluation}

Since the objective of visual instruction tuning to enhance the capability of multimodal model to understand the human instructions effectively and accurately, human evaluation is vital for assessing the tuned multimodal models, specifically for tasks that require a high level of understanding and could not be easily quantified by traditional metrics.
Specifically, human evaluation enables to assess the tuned model from various aspects, such as relevance that whether the model's response is relevant to the given instruction, coherence that if the text is logically consistent and well-organized and fluency that if the generated response is natural and correctly follows the grammatical rules.

\subsubsection{GPT-4 Evaluation}

Although human evaluation is beneficial and helpful, it is always time-consuming and costly. Inspired by the strong capability of GPT-4~\cite{openai2023gpt4} in understanding human instructions, some studies adopt GPT-4 as an alternative for measuring the quality of the model’s generated responses.
Specifically, GPT-4 evaluates the model from various aspects, including helpfulness, relevance, accuracy, and level of detail, and then assign an overall score ranging from 1 to 10, where higher scores reflect better performances.
In addition, GPT-4 will be required to give a detailed explanation for the evaluations, enabling better understand the capability of the tuned model.

\subsubsection{Quantitative Metric Evaluation}

In addition to human and GPT-4 evaluation, various downstream tasks are adopted for quantitative evaluation, including discriminative tasks, generative tasks and complex image reasoning tasks.
For evaluating the discrimination capability of the model, several image recognition tasks are adopted such as image classification~\cite{deng2009imagenet,nilsback2008automated}, object detection~\cite{lin2014microsoft,gupta2019lvis}, segmentation~\cite{kazemzadeh2014referitgame,kazemzadeh2014referitgame} and visual grounding~\cite{kazemzadeh2014referitgame}.
For evaluating the model's capability in generating image or video, image generation~\cite{chen2015microsoft}, pointcloud generation~\cite{deitke2023objaverse} and video generation~\cite{perazzi2016benchmark} tasks are adopted.
Besides, various tasks including visual question answering~\cite{mani2020point,marino2019ok} and image captioning~\cite{young2014image} are leveraged for assessing the model's capability in complex image reasoning.
Recently, several visual assistant benchmarks~\cite{li2023evaluating,fu2023mme,liu2023mmbench,liu2023mmbench,li2023seed,liu2023visual,yu2023mm} are proposed for comprehensively assessing the instruction-tuned model.
For example, MMBench~\cite{liu2023mmbench} is designed for robustly and accurately evaluating the various abilities of multimodal models by assessing the model from 20 different aspects, such as logic reasoning and  fine-grained perception.
SeedBench~\cite{li2023seed} enables comprehensive assessment by incorporating 12 evaluation tasks spanning from the understanding of the image to the comprehension of the video.

\section{Datasets}\label{Dataset}

This section summarizes the widely adopted datasets for visual instruction tuning and evaluations.

\subsection{Datasets for Visual Instruction Tuning}

For Visual Instruction Tuning, multiple multimodal instruction-following datasets were collected. According to the data type, instruction-following datasets can be categorized into single-turn dataset and multi-turn dataset, as detailed in Table~\ref{tab:sum_VLM_data}.
 
\subsubsection{Single-turn}

\begin{itemize}

    \item \textbf{MiniGPT-4~\cite{zhu2023minigpt}} curates an image description dataset that contains 3439 image-text pairs for instruction fine-tuning. MiniGPT-4 randomly selects 5000 images from the Conceptual Caption dataset~\cite{changpinyo2021conceptual,sharma2018conceptual} and prompts its pre-trained VLM model to generate detailed descriptions for each image. The generated descriptions are then refined and filtered both manually and by using ChatGPT, resulting in 3439 high-quality image-text pairs.

    \item \textbf{Clotho-Detail~\cite{zhao2023bubogpt}} is an audio-text instruction dataset that contains 3938 audio-text pairs with an average length of 52.7 words for description. Clotho-Detail is extended from Clotho~\cite{drossos2020clotho}, by using GPT-4 to aggregate its original short captions into long descriptions.

    \item \textbf{VGGSS-Instruction~\cite{zhao2023bubogpt}} is an image-audio-text triple-modality instruction dataset. It adopts a group of fixed templates to wrap the original labels of VGGSS~\cite{chen2021localizing} into descriptions. The dataset contains 5158 image-audio-text pairs where the audio is only related to a certain region in the image.
    
    \item \textbf{DetGPT~\cite{pi2023detgpt}} curate an instruction tuning dataset for reasoning-based object detection. Using short captions and category names of existing objects for each image as prompts, DetGPT uses ChatGPT to generate a long description, as well as several query-answer pairs for each image. The result instruction tuning dataset contains 5000 images and around 30000 query-answer pairs.
    
    \item \textbf{MultiInstruct~\cite{xu2022multiinstruct}} build a comprehensive instruction dataset that covers 62 diverse multimodal tasks from 10 broad categories, such VQA, Image-text matching, grounded generation, and so on. These tasks include 34 existing tasks derived from 21 public dataset and 28 new tasks extended from them. Each task is equipped with 5 instruction templates to prompt the model to perform the specific task.
    
    \item \textbf{Shikra-RD~\cite{chen2023shikra}} is an instruction-tuning dataset for the task of referential dialogue, which contains 5922 question-answer pairs. It resorts to GPT-4 to generate referential question-answer pairs based on the bounding box and description annotations of Flickr30K dataset, where the object coordinate may appear in both the questions or answers for referential region understanding.
    
    \item \textbf{MGVLID~\cite{zhao2023chatspot}} is a multi-grained vision-language instruction-following dataset, involving both image-level and region-level instruction data. For image-level instruction data, MGVLID collects commonly used Question-Answering, image captioning, and object detection datasets, and converts their annotations into a unified instruction format. For the region-level instruction data, MGVLID uses various instruction templates to refine region-text pairs, collected from existing regional-level tasks such as object detection and OCR, into question-answer pairs.

    \item \textbf{AS-1B~\cite{wang2023all}} is a large region-text dataset that contains 1.2 billion region-text pairs extracted from 11 million images. Each region is annotated with a semantic tag, several question-answer pairs, and a detailed caption for the comprehensive description, resulting in a total of 3.5 million distinct semantic tags for the entire dataset.
    
    \item \textbf{MM-IT~\cite{moon2023anymal}} is a multimodal instruction-tuning dataset that contains 60k manually annotated data and 150k synthetic data for diverse modalities including image, video, and audio.

    \item \textbf{LRV-Instruction~\cite{liu2023mitigating}} is a large-scale robust visual-instruction dataset that contains 400K instructions generated by GPT-4, involving 16 vision-language tasks. In addition to positive question-answer pairs, LRV introduces negative instructions, which may involve manipulations or incorrect content, to improve the robustness of LLMs.

    \item \textbf{VisIT-Bench~\cite{bitton2023visit}} is a visual instruction benchmark that contains 592 test instances, covering tasks from basic recognition to game playing and creative generation.

    \item \textbf{T2M~\cite{wu2023next}} is a text-to-multimodal instruction dataset that contains 14.7k instances. The target is to generate corresponding multimodal contents given text captions.
    
    \item \textbf{ChiMed-VL-Instruction~\cite{liu2023qilin}} is a Chinese medicine vision language instruction dataset that contains 479k question-answer pairs.

    \item \textbf{Valley-Instruct-73k~\cite{luo2023valley}} is a video instruction dataset that contains 73k instruction data, including 37k conversation pairs, 26k reasoning QA pairs and 10k description pairs.
    
    \item \textbf{MACAW-LLM~\cite{lyu2023macaw}} is a multimodal instruction dataset that consists of 69K image instruction pairs generated from COCO image captions~\cite{lin2014microsoft} and 50K video instruction pairs generated from Charades~\cite{sigurdsson2016much} and AVSD~\cite{alamri2019audio}.

\end{itemize}

\begin{table*}[!t]
    \setlength\tabcolsep{6pt}
    \centering
    \caption{Summary of visual instruction tuning datasets. 
    }
    \resizebox{0.95\linewidth}{!}{
    \begin{tabular}{l|l|l|l|l|l}
    \toprule[1pt]
        \textbf{Data Type} & \textbf{Dataset}  & \textbf{Size} &\textbf{Modality} & \textbf{Language} & \textbf{Construction} \\
        \midrule
        Single-Turn Instruction & MiniGPT-4~\cite{zhu2023minigpt} & 3.5K & Image,Text & English & GPT-3.5-generated \\
        & Clotho-Detail~\cite{zhao2023bubogpt} & 3.9K & Text,Audio & English & GPT-4-generated \\
        & VGGSS-Instruction~\cite{zhao2023bubogpt} & 5.2K & Image,Text,Audio & English & GPT-4-generated \\
        & DetGPT~\cite{pi2023detgpt} & 30K &Image,Text&English & GPT-4-generated \\
        &MultiInstruct~\cite{xu2022multiinstruct}& - & Image,Text&English& Manual Annotation \\
        &Shikra-RD~\cite{chen2023shikra}& 5.9K & Image,Text&English &GPT-4-generated\\
        &MGVLID~\cite{zhao2023chatspot} & 3M & Image,Text&English& GPT-4-generated\\
        & AS-1B~\cite{wang2023all}&1B & Image,Text&English & Model-generated\\ 
        & MM-IT~\cite{moon2023anymal} &210K& Image,Text&English& Manual Annotation/LLaMA-2-generated \\
        &LRV-Instruction~\cite{liu2023mitigating} & 400K& Image,Text&English& GPT-4-generated \\
        &VisIT-Bench~\cite{bitton2023visit} & - & Image,Text &English& GPT-4-generated \\
        &T2M~\cite{wu2023next} &14.7K & Image,Video,Text,Audio& English &GPT-4-generated \\
        &ChiMed-VL-Instruction~\cite{liu2023qilin} & 469K & Image,Text & Chinese & GPT-3.5-generated\\
        &Valley-Instruct-73k~\cite{luo2023valley} & 73K & Video,Text & English & GPT-3.5-generated \\
        &MACAW-LLM~\cite{lyu2023macaw} & 119K& Image,Video,Text& English & GPT-3.5-turbo-generated \\
       
        \midrule
        Multi-Turn Instruction  & LLaVA-Instruct-158k~\cite{liu2023visual}  & 158K & Image,Text & English & ChatGPT-generated\\ 
        & GPT4RoI~\cite{zhang2023gpt4roi} & - & Image,Text & English & - \\
        &MultiModal-GPT~\cite{gong2023multimodal}  & - & Image,Text & English & GPT-4-generated \\
        &MIMIC-IT~\cite{li2023otter}  & 2.8M & Image,Video,Text & Multiple languages &ChatGPT-generated \\
        &SVIT~\cite{zhao2023svit} & 4.2M &Image,Text&English &GPT-4-generated \\
        &PF-1M~\cite{chen2023visual} & 975k &Image,Text&English & Self-Instructed \\
        &ILuvUI~\cite{jiang2023iluvui}& 353K &Image,Text&English&GPT-3.5-generated\\
        &StableLLaVA~\cite{li2023stablellava}  &126K & Image,Text&English & StableDiffusion \& ChatGPT-generated  \\
        &X-LLM~\cite{chen2305bootstrapping} & 10K &Image,Video,Text  & Chinese,English & ChatGPT-generated\\ 
        &GPT4Tools~\cite{yang2023gpt4tools} &71K&Image,Text&English&GPT-3.5-generated \\
        &LLaVAR~\cite{zhang2023llavar} & 16K&Image,Text&English&GPT-3.5-generated \\
        & PVIT~\cite{chen2023position} & 22K &Image,Text & English & GPT-3.5-generated  \\
        & SparklesDialogue~\cite{huang2023sparkles} & 6.4K & Image,Text&English &GPT-4-generated \\
        & GRIT~\cite{you2023ferret} & 1.1M & Image,Text&English &GPT-4-generated \\
        &VIGC-LLaVA~\cite{wang2023vigc} & 1.8M &Image,Text&English& Model-generated \\ 
        & M$^3$IT~\cite{li2023m} & 2.4M & Image,Video,Text & 80 Languages & - \\
        & LLaVA-Med~\cite{li2023llava} & 60K & Image,Text&English &GPT-4-generated \\  
        &Mosit~\cite{wu2023next} & 5K &Image,Video,Text,Audio &English & GPT-4-generated\\
        &PointLLM~\cite{xu2023pointllm} & 730K & PointCloud,Text &English &  GPT-4-generated\\
        & TEXTBIND~\cite{li2023textbind} & 25.6K& Image,Text&English &GPT-4-generated \\  
        & MULTIS~\cite{zhao2023chatbridge} & 4.6M &Image,Video,Text,Audio &English & Models and ChatGPT-generated \\
        &LAMM~\cite{yin2023lamm} & 196K & Image,PointCloud,Text &English& GPT-4-generated \\
        &VideoChat~\cite{li2023videochat} & 11K & Video,Text & English &GPT-4-generated \\
        &Video-ChatGPT~\cite{maaz2023video} & 100K & Video,Text & English & Human Crafted, Model \& GPT-3.5-generated \\
        &OphGLM~\cite{gao2023ophglm} & 20K & Image,Text& English & GPT-3.5-generated \\     
    \bottomrule[1pt]
    \end{tabular}
    }
\label{tab:sum_VLM_data}
\end{table*}

\subsubsection{Multi-turn}

\begin{itemize}
    \item \textbf{LLaVA-Instruct-158k~\cite{liu2023visual}} contains 158 image-text instruction data, including 58k conversation data asking about the visual content of the image, 23k description data, and 77k complex reasoning data where the question may involve multi-step reasoning process.
    
    \item \textbf{GPT4RoI~\cite{zhang2023gpt4roi} } convert Visual Genome region caption annotations~\cite{krishna2017visual}, RefCOCOg~\cite{mao2016generation}, Flcker30k~\cite{Plummer_2015_ICCV}, and Visual Commonsense Reasoning~\cite{Zellers_2019_CVPR} into instruction data for single/multiple region understanding, and leverage LLaVA-Instruct-158k~\cite{liu2023visual} supplemented with bounding box annotations to improve the capability of multi-round conversation.
    
    \item \textbf{MultiModal-GPT~\cite{gong2023multimodal} } employs a unified instruction template to construct instruction data for both language-only data such as Dolly 15k and Alpaca GPT4~\cite{peng2023instruction} and language-vision data including LLaVA~\cite{liu2023visual}, Mini-GPT4~\cite{zhu2023minigpt}, A-OKVQA~\cite{marino2019ok}, COCO Caption~\cite{chen2015microsoft}, and OCR VQA~\cite{mishra2019ocr}.

    \item \textbf{MIMIC-IT~\cite{li2023mimicit}} is an instruction dataset that contains 2.8 million multimodal instruction-response pairs for language, image, and video understanding. It contains 502k video clips and 8.1 million images, supporting eight languages including English, Chinese, Spanish, Japanese, French, German, Korean, and Arabic.

    \item \textbf{SVIT~\cite{zhao2023svit}} is a large instruction dataset that contains 4.2 million visual instruction data. It comprises 1.6 million conversation QA pairs, 1.6 complex reasoning QA pairs, 1.0 million referring QA pairs, and 106k image description data, supporting comprehensive capability of visual understanding and reasoning. 
    
    \item \textbf{PF-1M~\cite{chen2023visual}} contains 975k instruction-response data. It collects 37 image captioning and VQA datasets, then uses its pre-trained Polite Flamingo~\cite{chen2023visual} to rewrite their original annotations into a unified instruction-answer format, and clean the data on both rule-based and model-based filters, obtaining 975k high-quality instruction data.
    
    \item \textbf{ILuvUI~\cite{jiang2023iluvui}} is instruction dataset for UI tasks, i.e. UI element detection or multi-step UI navigation and planning. It contains 224K conversations, 32K consise description data, 32k detailed description dta, 32k logical reasoning data, 32k potential actions, and 1k UI transition data.
    
    \item \textbf{StableLLaVA~\cite{li2023stablellava}} is a synthetic image-dialogue dataset. It uses Chatgpt to generate image prompts, then cooperates with StableDiffusion~\cite{rombach2022high} to generate the corresponding image, and additionally employs Chatgpt to generate descriptions based on the same image prompts, resulting in 126K image-dialogue pairs.
    
    \item \textbf{X-LLM~\cite{chen2305bootstrapping}} construct a multimodal instruction data including about 10k samples that are selected and transformed from MiniGPT-4~\cite{zhu2023minigpt}, AISHELL-2~\cite{du2018aishell}, VSDial-CN, and ActivityNet Caps~\cite{krishna2017dense}.
    
    \item \textbf{GPT4Tools~\cite{yang2023gpt4tools}} curate a instruction dataset to enable LLMs to use multimodal tools. It contains 71.4K instruction-following data involving 23 tools for image generation and image understanding for the training set, 1170 data which share the same tools involved in the training data as validation set, and 652 samples including 8 new tools as test set.
    
    \item \textbf{LLaVAR~\cite{zhang2023llavar}} construct 16K multi-turn conversation data for text-rich image understanding, by prompt GPT-4 with OCR data and image captions.

    \item \textbf{PVIT~\cite{chen2023position}} build an image-region-language instruction dataset. It contains 146k single-turn instruction data converted from VQA datasets, 86k instruction data for five specific tasks~(i.e., small object recognition, same-category object discrimination, object relationship based reasoning, object attribute based reasoning, and optical character recognition) on object understanding, and 22k general instruction data generated by prompting ChatGPT with image description and in-centext examples.
    
    \item \textbf{SparklesDialogue~\cite{huang2023sparkles}} is instruction dataset for conversations involving multiple images. It comprises of two parts, SparklesDialogueCC and SparklesDialogueVG. SparklesDialogueCC, generated based on Conceptual Captions~\cite{sharma2018conceptual}, contains 4.5k dialogues, each of which consists of at least two images and two round of conversation. And SparklesDialogueVG is built from Visual Genome~\cite{krishna2017visual} and includes 2k dialogue. Each dialogue contains at least threeimages across two turns.
    
    \item \textbf{GRIT~\cite{you2023ferret}} is large instruction dataset for referring and grounding tasks. It contains 1.1 million instruction data for image reasoning and understanding which are converted from public dataset or generated via ChatGPT and GPT-4, and 130k negative data to improve model robustness and reduce object hallucination.
    
    \item \textbf{VIGC-LLaVA~\cite{wang2023vigc}} is an instruction dataset autonomously generated by VLLMs through the Visual Instruction Generation and Correction~(VIGC) framework~\cite{wang2023vigc}. It contains 36.7k instruction data generated from COCO dataset~\cite{lin2014microsoft} and 1.8 million instruction data from Objects365~\cite{shao2019objects365}.
    
    \item \textbf{M$^3$IT~\cite{li2023m}} is a multimodal, multilingual instruction tuning dataset that contains 2.4 million instances. It involves 40 visual-language tasks and 400 munnually written instruction templates, with seven tasks translated into 80 languages.
    
    \item \textbf{LLaVA-Med~\cite{li2023llava}} curate a biamedical instruction dataset by prompte GPT-4 to generate multi-round conversations. It contains $60, 000$ image-text pairs with 5 medical image modalities, including CXR (chest X-ray), CT (computed tomography), MRI (magnetic resonance imaging), histopathology, and gross (i.e., macroscopic) pathology.
    
    \item \textbf{Mosit~\cite{wu2023next}} is a modality-switching instruction tuning dataset that supports complex multimodal inputs and outputs for multi-round instruction conversation. Each conversation in Mosit consists 3-7 rounds~(question-answer pairs) where either question or answer may include multimodal content~(text, image, audio, and video) at either the question or the answer. It contains a total of 5k dialogues.

    \item \textbf{PointLLM~\cite{xu2023pointllm}} constructs large point-text instruction dataset that contains 660k description data and 70k complex instruction. It leverages GPT-4 to convert the 3D object captioning dataset Cap3D~\cite{luo2023scalable} into instruction following dataset.
    
    \item \textbf{TEXTBIND~\cite{li2023textbind}} curates a instruction dataset containing 25.6k conversation for image understanding. which is achieved by applying its proposed TEXTBIND, an annotation-free frmawork for improving the multi-turn instruction following capability of LLMs, to GPT4 and the CC3M~\cite{sharma2018conceptual} dataset.
    
    \item \textbf{MULTIS~\cite{zhao2023chatbridge}} is a multimodal instruction-tuning dataset that contains 4.4 million task-specific samples that are converted from public question-answering and captioning dataset using ChatGPT, and 209k multimodal chat samples involving conversations, descriptions and complex reasoning on multiple modalities including text, image, audio, and video.
    
    \item \textbf{LAMM~\cite{yin2023lamm}} includes 186k text-image instruction pairs, and 10k text-pointcloud instruction pairs. It contains four types of instruction data, including daily conversation, factual knowledge dialogue about knowledge and content reasoning, detailed description, and visual task dialogues. The task dialogues involve most vision tasks for both 2D and 3D vision, such as captioning, scene graph recognition, classification, detection, counting and OCR.
    
    \item \textbf{VideoChat~\cite{li2023videochat}} is a video-centric instruction dataset build from WebVid-10M~\cite{bain2021frozen} using ChatGPT. It contains 7K video descriptions and 4k video conversations.
    
    \item \textbf{Video-ChatGPT~\cite{maaz2023video}} is a video-based instruction daaset containing 100k video-instruction pairs annotated by human annotators, off-the-shelf models, and GPT3.5. It covers various data types such as detailed descriptions, summarizations, question-answer pairs, and conversations, .etc.
    
    \item \textbf{OphGLM~\cite{gao2023ophglm}} is a ophthalmic instruction dataset comprising of 20k dialogs related to ophthalmic diseases.
    
\end{itemize}

\subsection{Datasets for Instruction-tuned Model Evaluation}
With visual instruction tuning, we can build general-purpose multimodal models that can solve various vision tasks according to users' instructions.
Various datasets have been adopted in Instruction-tuned models evaluations, including datasets for discriminative image tasks (e.g., image classification~\cite{fei2004learning,deng2009imagenet,parkhi2012cats,nilsback2008automated,krause2013collecting}, object detection~\cite{lin2014microsoft,gupta2019lvis,everingham2010pascal}, image segmentation~\cite{kazemzadeh2014referitgame,kazemzadeh2014referitgame,mao2016generation}, visual grounding~\cite{kazemzadeh2014referitgame}), generative image tasks (e.g., image generation~\cite{chen2015microsoft}), complex image reasoning tasks (e.g., visual question answering~\cite{mani2020point,goyal2017making,marino2019ok,hudson2019gqa,lu2022learn,gurari2018vizwiz,singh2019towards}, image captioning~\cite{chen2015microsoft,young2014image,krishna2017visual,agrawal2019nocaps}, visual assistant~\cite{li2023evaluating,fu2023mme,liu2023mmbench,li2023seed,liu2023visual,yu2023mm}), video tasks (e.g., video generation~\cite{kim2019audiocaps,perazzi2016benchmark}, video captioning~\cite{kim2019audiocaps,wang2019vatex}, video VQA~\cite{xu2017video,jang2017tgif,yu2019activitynet}), medical vision tasks (e.g., medical VQA~\cite{lau2018dataset,liu2021slake,he2020pathological}, medical classification~\cite{li2019diagnostic}, medical segmentation~\cite{li2019diagnostic}), document vision tasks (e.g., document VQA~\cite{mathew2021docvqa,masry2022chartqa,chen2019tabfact,tanaka2021visualmrc}) and 3D vision tasks (e.g., pointcloud classification~\cite{wu20153d,deitke2023objaverse}, pointcloud generation~\cite{deitke2023objaverse}, pointcloud VQA~\cite{azuma2022scanqa}, pointcloud detection~\cite{dai2017scannet}).

\begin{table*}[!t]
\setlength\tabcolsep{6pt}
\centering
\caption{Summary of visual instruction tuning methods (Part 1).
}
\resizebox{0.99\linewidth}{!}{
\begin{tabular}{l|l|p{2.5cm}|p{2.5cm}|p{8.5cm}}
\toprule[1pt]
\multirow{2}{*}{\textbf{Task}} &\multirow{2}{*}{\textbf{Method}} & \multicolumn{2}{c|}{\textbf{Base Model}}  & \multirow{2}{*}{\textbf{Tuning Data}}  \\
\cmidrule{3-4}
&&\textbf{Vision Encoder}&\textbf{Language Encoder}& \\
\midrule
\multicolumn{5}{c}{\textbf{Instruction-based Image Learning for Discriminative Tasks}} \\
\midrule
 Image Classification & Instruction-ViT~\cite{xiao2023instruction} & CLIP ViT & CLIP Transformer \\
\midrule
  {Image Segmentation} &LISA~\cite{lai2023lisa} & CLIP ViT, SAM ViT & Vicuna & ADE20K, COCO, LVIS, RefCOCO, RefCOCO+, RefCOCOg, RefCLEF, LLaVA-Instruct-158K \\
 \midrule
  Object Detection & VisionLLM~\cite{wang2023visionllm}&ResNet,InternImage-H&  BERT-Base &COCO2017, RefCOCO, RefCOCO+, RefCOCOg, COCO Caption, COCO Caption  \\
 &DetGPT~\cite{pi2023detgpt}&BLIP-2&13B Vicuna &SBU, LAION, Conceptual Caption, Query-answer Instruction Dataset \\
 &Shikra~\cite{chen2023shikra} &ViT-L/14&Vicuna-7/13B&LLaVA-Pretraining, Flickr30K Entities, RefCOCO, Visual Gemone, Visual-7W, RefCOCO, RefCOCO+, RefCOCOg, VQAv2, PointQA-Local/Twice, LLaVA-Instruct-150K, VCR, Shikra-RD \\
 &ChatSpot~\cite{zhao2023chatspot}&CLIP ViT-L/14 &Vicuna-7B &Multi-Grained Vision-Language Instruction-following Dataset \\
 &ASM~\cite{wang2023all}&ViT-g/14&Husky-7B&The All-Seeing Dataset (AS-1B) \\
 &PVIT~\cite{chen2023position}&RegionCLIP ResNet50x4 & LLaVA-7B& GQA, VCR\\
 \midrule
 Visual Grounding& BuboGPT~\cite{zhao2023bubogpt} & SAM ViT & Vicuna & MiniGPT-3.5K, LLaVA-Instruct-158K, Clotho-Detail \\
&FERRET~\cite{you2023ferret} & CLIP-ViT-L/14@336p &Vicuna&Ground-and-Refer Instruction-Tuning dataset \\
&GLaMM~\cite{rasheed2023glamm}&CLIP ViT-H/14, SAM ViT&Vicuna  &Grounding-anything Dataset\\
\midrule
\multicolumn{5}{c}{\textbf{Instruction-based Image Learning for Generative Tasks}} \\
\midrule
 Image Generation &GPT4Tools~\cite{yang2023gpt4tools}&Q-Former&LLaMA, Vicuna, OPT & GPT4Tools Dataset \\
&TEXTBIND~\cite{li2023textbind}&BLIP2&Stable Diffusion XL&LLaVA, MiniGPT-4, MultiInstruct, Platypus, Shikra \\
\midrule
 Image Editing & LLaVA-Interactive~\cite{chen2023llava} \\
\midrule
\multicolumn{5}{c}{\textbf{Instruction-based Image Learning for Complex Reasoning Tasks}} \\
\midrule
Image Captioning&GPT4RoI~\cite{zhang2023gpt4roi} & CLIP ViT & Vicuna & LLaVA-Instruct-158K, RefCOCOg, VG, Flicker30k \\ 
 & MiniGPT-4~\cite{zhu2023minigpt} & EvaCLIP ViT & Vicuna & CC3M, CC12M, SBU, LAION 400M, MiniGPT-3.5K \\
&Clever Flamingo~\cite{chen2023visual} &OpenFlamingo ViT &LLaMA& PF-1M\\
&DreamLLM~\cite{dong2023dreamllm}&CLIP-L/14  &Vicuna-7B &LAION400M, LAION-COCO, MMC4, BLIP-LAION, LLaVAPretrain, LLaVAInstruct, InstructMMC4, Instruct-BLIP-LAION  \\
&AnyMAL~\cite{moon2023anymal}&CLIP ViT-L, ViT-G, DinoV2 &Vicuna &LAION-2B Dataset, AudioSet, AudioCaps, CLOTHO, Ego4D  \\
\midrule
 Visual Question Answering&LaVIN~\cite{luo2023cheap} & CLIP ViT &LLaMA & ScienceQA, Alphaca-52k, LLaVA-158k\\
&SCITUNE~\cite{horawalavithana2023scitune}&CLIP ViT &LLaVA &SciCap datasets \\
&MultiInstruct~\cite{xu2022multiinstruct}&OFA&OFA&VQAv2, Visual7w, GQA, OK-VQA, Visual Genome, MSCOCO, RefCOCO, COCO-Text, TDIUC, IQA, VAW, MOCHEG, WikiHow\\
&LMEye~\cite{li2023lmeye} & BLIP-2&LLaMA-7b/13b &SemArt Dataset \\
&VPG-C~\cite{li2023fine}&EVA-CLIP &Vicuna-7B &DEMON  \\
&BLIVA~\cite{hu2023bliva}& EVA-CLIP-ViT-G/14 & LLaVA&MSCOCO, TextCaps, VQAv2, OKVQA, A-OKVQA, OCR-VQA, LLaVA-Instruct-150K \\
&MiniGPT-v2~\cite{chen2023minigpt}&EVA& LLaMA2-chat (7B)  &  LAION, CC3M, SBU, GRIT-20M, COCO caption,  Text Captions, RefCOCO, RefCOCO+, RefCOCOg, GQA, VQA-v2, OCR-VQA, OK-VQA, AOK-VQA, Flickr30k Dataset, Unnatural Instruction Dataset   \\
&mPLUG-Owl2~\cite{ye2023mplug}&ViT-L/14 &LLaMA-2-7B &VQAv2, GQA, OKVQA, OCRVQA, A-OKVQA, COCO, TextCaps \\
&InstructBlip~\cite{dai2023instructblip} &&&COCO Caption, Web CapFilt, NoCaps, Flickr30K, TextCaps, VQAv2, VizWiz, GQA, Visual Spatial Reasoning, IconQA, OKVQA, A-OKVQA, ScienceQA, Visual Dialog, OCR-VQA, TextVQA, HatefulMemes, LLaVA-Instruct-150K, MSVD-QA, MSRVTT-QA, iVQA \\
&InternLM-XComposer~\cite{zhang2023internlm}&EVA-CLIP & InternLM & In-house Data, LLaVA-150k, Alpaca-en\&zh, ShareGPT-en\&zh, Oasst-en\&zh, LRV \\
\midrule
Visual Assistant & LLaVa~\cite{liu2023visual} & CLIP ViT & Vicuna & CC3M Concept-balanced 595K, LLaVA-Instruct-158K \\ 
(Visual Chatbot)
&LLaMA-Adapter-V2~\cite{gao2023llama} & CLIP ViT & LLaMA&GPT-4-LLM, COCO \\
&Otter~\cite{li2023otter} & CLIP ViT & MPT & MIMIC-IT \\
&MultiModal-GPT~\cite{gong2023multimodal} &CLIP ViT &LLaMA& LLaVA-Instruct-158K, MiniGPT-3.5K, A-OKVQA, COCO Caption, OCR VQA\\
&LLaVA-1.5~\cite{liu2023improved} &  CLIP ViT & Vicuna &LLaVA, ShareGPT, VQAv2, GQA, OKVQA, OCRVQA, A-OKVQA, TextCaps, RefCOCO, VG \\
&SVIT~\cite{zhao2023svit} & CLIP ViT & Vicuna& SVIT-4.2M \\
&ILuvUI~\cite{jiang2023iluvui} &  CLIP ViT & Vicuna& CC3M Concept-balanced 595K, LLaVA-Instruct-158K \\
&AssistGPT~\cite{gao2023assistgpt} &BLIP2, Gounding Dino, Google OCR & Vicuna& A-OKVQA, NExT-QA\\
&StableLLaVA~\cite{li2023stablellava}&CLIP-ViT-L/14 & LLaMA & Synthesized Image-Dialogue Dataset \\
&X-LLM~\cite{chen2305bootstrapping}&Blip-2 & Vicuna& CC3M, COCO, VG-Caps, Flickr30k, SBU, AI-Caps, Wukong, MSRVTT, AISHELL-1, AISHELL-2, VSDial-CN, AISHELL-2, VSDial-CN, MiniGPT-4, AISHELL-2, VSDial-CN, ActivityNet Caps\\ 
&PandaGPT~\cite{su2023pandagpt}&ImageBind &Vicuna-13B & Image-language Instruction-following Dataset\\
&LAMM~\cite{yin2023lamm}&CLIP ViT-L/14 &Vicuna & Language-Assisted Multi-Modal Instruction-Tuning Dataset \\
&LLaVAR~\cite{zhang2023llavar}&CLIP-ViT-L/14, CLIP-ViT-L/14-336 &Vicuna-13B &LAION-5B \\
&Qwen-VL~\cite{bai2023qwen}&ViT & Qwen-7B&LAION-en\&zh, DataComp, Coyo, CC12M\&3M, SBU, COCO, In-house Data, GQA, VGQA, VQAv2, DVQA, OCR-VQA, DocVQA, TextVQA, ChartQA, AI2D, GRIT, Visual Genome, RefCOCO, RefCOCO+, RefCOCOg, SynthDoG-en\&zh, Common Crawl pdf\&HTML \\
&Sparkles~\cite{huang2023sparkles}&BLIP-2, EVA-ViT& MiniGPT-4 & SparklesDialogue, SparklesDialogueCC, SparklesDialogueVG\\
&CogVLM~\cite{wang2023cogvlm}&EVA2-CLIP-E & Vicuna-7Bv-1.5&VQAv2, TextVQA \\
&SEED-LLaMA~\cite{ge2023making}&ViT &LLaMA&JourneyDB, DiffusionDB, LAION-Aesthetics, VIST, Instructpix2pix, MagicBrush, LLaVA, LLaVAR, GSD, VSR, MagicBrush, TextCaps, VQAv2, OKVQA, A-OKVQA, GQA, VizWiz, TextVQA, OCR-VQA, Video-ChatGPT, ActivityNet, Next-QA, MSVD, MSR-VTT, iVQA \\
&OtterHD~\cite{li2023otterhd}&Fuyu-8B&Fuyu-8B& LLaVA-Instruct, VQAv2, GQA, OKVQA, OCRVQA, A-OKVQA, COCO-GOI, COCO-Caption, TextQA, RefCOCO, COCOITM, ImageNet, LLaVA-RLHF \\
&ImageBind-LLM~\cite{han2023imagebind}&ImageBind & LLaMA& COCO, CC3M, CC12M, SBU, LAION-2B, COYO, MMC4  \\
\bottomrule[1pt]
\end{tabular}
}
\label{tab:sum_VLM_1}
\end{table*}

\begin{table*}[!t]
    \setlength\tabcolsep{6pt}
    \centering
    \caption{Summary of visual instruction tuning methods (Part 2).}
    \resizebox{0.99\linewidth}{!}{
    \begin{tabular}{l|l|p{2.5cm}|p{2.5cm}|p{8.5cm}}
    \toprule[1pt]
        \multirow{2}{*}{\textbf{Task}}&\multirow{2}{*}{\textbf{Method}} & \multicolumn{2}{c|}{\textbf{Base Model}}  & \multirow{2}{*}{\textbf{Tuning Data}}  \\
        \cmidrule{3-4}
        &&\textbf{Vision Encoder}&\textbf{Language Encoder}& \\
        \midrule
        \multicolumn{5}{c}{\textbf{ Instruction-based Video Learning}} \\
        \midrule
         Visual Assistant & NExT-GPT~\cite{wu2023next}&ImageBind & Vicuna &`Text+X' — `Text' Data, `Text' — `Text+X' Data, MosIT Data   \\
         (Visual Chatbot) 
        &EmbodiedGPT~\cite{mu2023embodiedgpt}&ViT-G/14, ResNet50&LLaMA-7B& EgoCOT Dataset, EgoVQA Dataset\\
        & ChatBridge~\cite{zhao2023chatbridge} &ViT-G&Vicuna  &MULTimodal InStruction tuning Dataset\\
        & VideoChat~\cite{li2023videochat} &ViT-G & StableVicuna& Video-centric Multimodal Instruction Data\\
        & Video-ChatGPT~\cite{maaz2023video}&CLIP ViT-L/14&Vicuna language decoder &ActivityNet-200 dataset \\
        & Video-LLaMA~\cite{zhang2023video}& ViTG/14 from EVA-CLIP &Vicuna/LLaMA&Webvid-2M, CC595k, Image-Detail-description Dataset,  Image-instruction Dataset, Video-instruction Dataset \\
        & VALLEY~\cite{luo2023valley}&CLIP ViT-L/14&Stable-Vicuna& Jukinmedia Dataset\\
        & MACAW-LLM~\cite{lyu2023macaw}&CLIP-ViT-B/16 &LLaMA-7B &Macaw-LLM Instruction Dataset \\
        \midrule
        \multicolumn{5}{c}{\textbf{Instruction-based Medical Vision Learning}} \\
        \midrule
        Visual Question Answering & PMC-VQA~\cite{zhang2023pmc}&PMC-CLIP ResNet-50 &LLaMA, PMC-LLaMA, PubMedBERT, LLaMA-ENC, PMC-LLaMA-ENC &PMC-VQA Dataset \\
        \midrule
         Visual Assistant  & LLaVA-Med~\cite{li2023llava} \\
        (Visual Chatbot) &Qilin-Med-VL~\cite{liu2023qilin}&ViT & ChineseLLaMA2-13B-Chat & ChiMed-VL \\
        &OphGLM~\cite{gao2023ophglm}&GLM-130B&ChatGLM & Ophthalmology Dataset \\
         \midrule
        \multicolumn{5}{c}{\textbf{Instruction-based Document Vision Learning}} \\
        \midrule
        Visual Assistant & mPLUG-DocOwl~\cite{ye2023mplugdoc}& CLIP ViT-L/14 &Vicuna&ChartQA, DocVQA, InfographicsVQA, WikiTableQuestions, TextVQA, VisualMRC, DeepForm, Kleister Charity, TabFact, TextCaps \\
        (Visual Chatbot)&mPLUG-PaperOwl~\cite{hu2023mplug}&ViT-L/14 &LLaMA-7B &M-Paper Dataset\\
        \midrule
        \multicolumn{5}{c}{\textbf{Instruction-based 3D Vision Learning}} \\
        \midrule
         Visual Assistant & PointLLM~\cite{xu2023pointllm}&ULIP-2, ULIP-2 &Vicuna &Cap3D  \\
        (Visual Chatbot) & LAMM~\cite{yin2023lamm} &CLIP ViT-L/14 &Vicuna & Language-Assisted Multi-Modal Instruction-Tuning Dataset\\
        \bottomrule[1pt]
    \end{tabular}
    }
    \label{tab:sum_VLM_2}
\end{table*}

\begin{figure}[!t]
    \centering
    \includegraphics[width=0.49\textwidth]{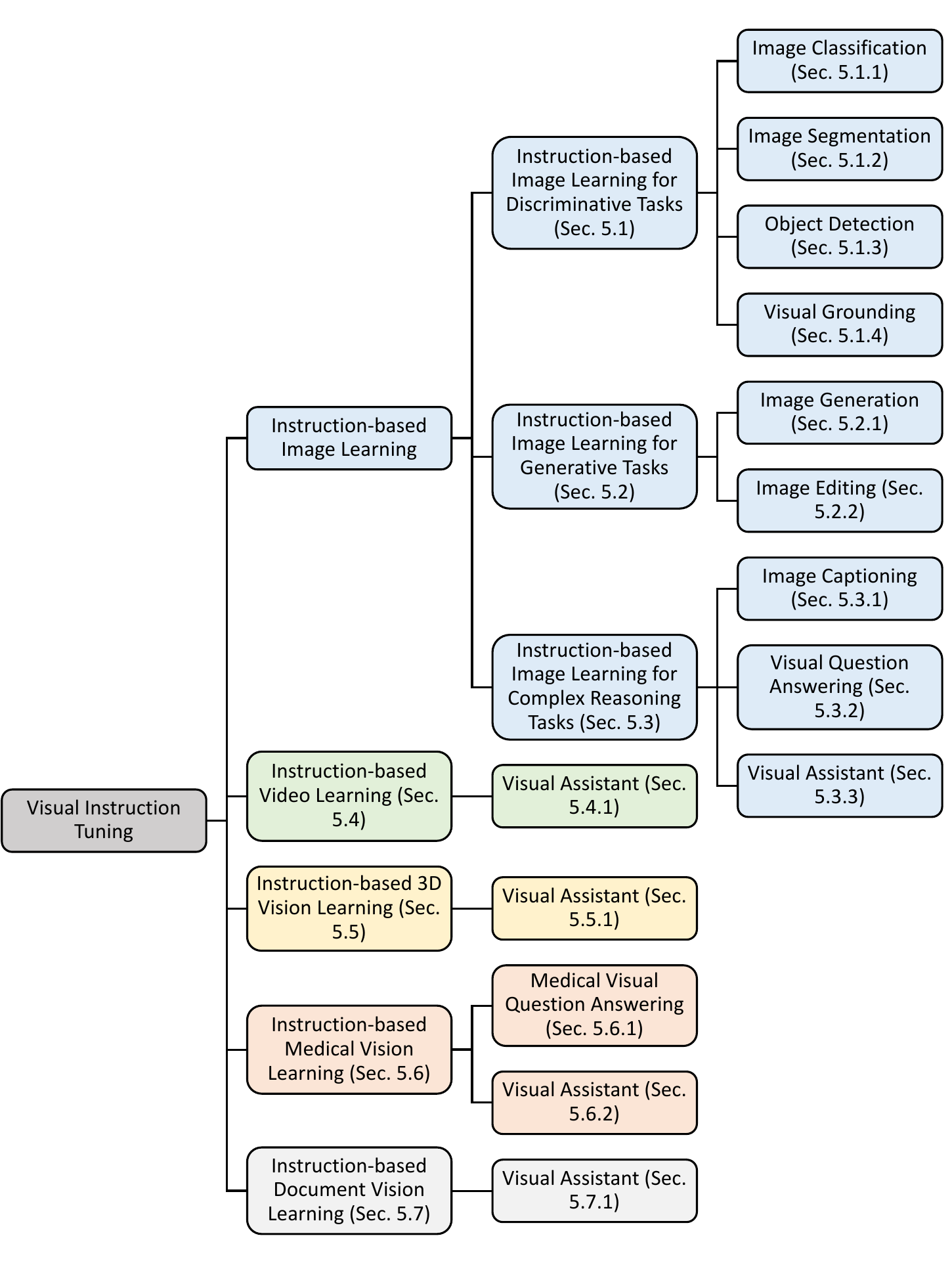}
    \caption{ Typology of visual instruction tuning.}
    \label{fig:arch}
\end{figure}

\section{Visual Instruction Tuning}\label{Sec.VLP}

Visual instruction tuning towards general-purpose multimodal models has been explored for various vision tasks, including discriminative tasks, generative tasks, complex image reasoning tasks, video tasks, medical vision tasks, document vision tasks, and 3D vision tasks as illustrated in Table~\ref{fig:arch}. 
This section reviews them with the above-mentioned tasks listed in Tables~\ref{tab:sum_VLM_1} and~\ref{tab:sum_VLM_2}.

\subsection{Instruction-based Image Learning for Discriminative Tasks}
Instruction-based image learning for discriminative tasks has been widely explored for general-purpose multimodal models, which construct instruction datasets and tuning methods for learning discriminative multimodal features.

\subsubsection{Image Classification}
\begin{figure}[ht]
    \centering
    \includegraphics[width=0.35\textwidth]{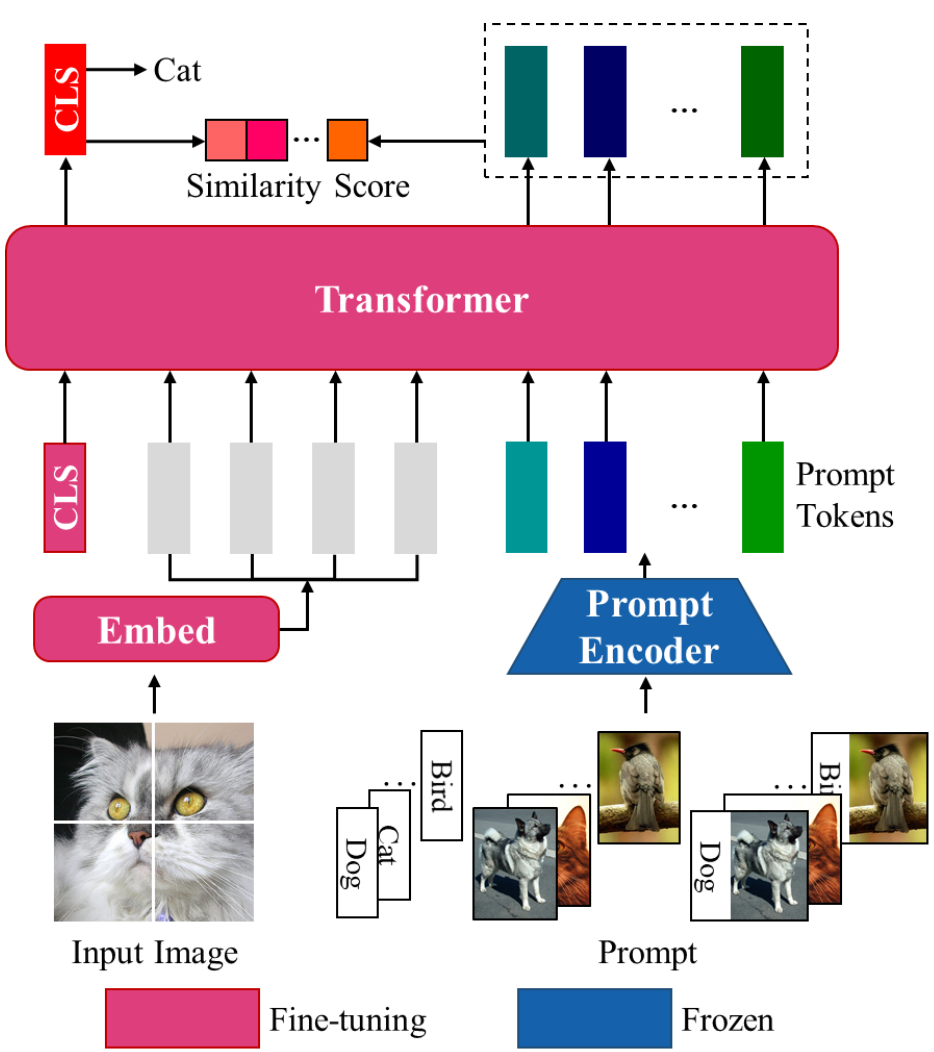}
    \caption{Illustration of the  Insturction-ViT~\cite{xiao2023instruction}. 
    Figure is from~\cite{xiao2023instruction}.}
    \label{fig:instruction_ViT}
\end{figure}

In this task, visual instruction tuning~\cite{xiao2023instruction} aims to learn multimodal category information for image classification by specifically designed instruction tuning methods and datasets.
For example, 
Instruction-ViT introduces the instruction tuning method into the vision transformer (ViT) via employing and fusing the multimodal prompts (in texts and images) that carry class-related information for guiding model fine-tuning as shown in Figure~\ref{fig:instruction_ViT}.
Specifically, Instruction-ViT leverages the self-attention mechanisms of the transformer to combine the multimodal prompts and input image. 
Then it uses a learnable [CLS] token to represent global image features and a series of prompt tokens to represent prompt features to complete the downstream task of classification, where the similarity between [CLS] token and prompt tokens have been utilized to guide model fine-tuning. 
The innovative instruction tuning method of fusing multimodal prompts improves accuracy and domain adaptation ability for image classification networks.

\subsubsection{Image Segmentation}
Image segmentation aims to partition a digital image into multiple segments or regions to simplify or change the representation of an image into something that is more semantic and easier to analyze.
In general-purpose multimodal models with visual instruction tuning, image segmentation involves using the multimodal instructions and expressions to guide the model to reason and comprehend users' intents, segmenting regions in images.

\begin{figure}[ht]
    \centering
    \includegraphics[width=0.48\textwidth]{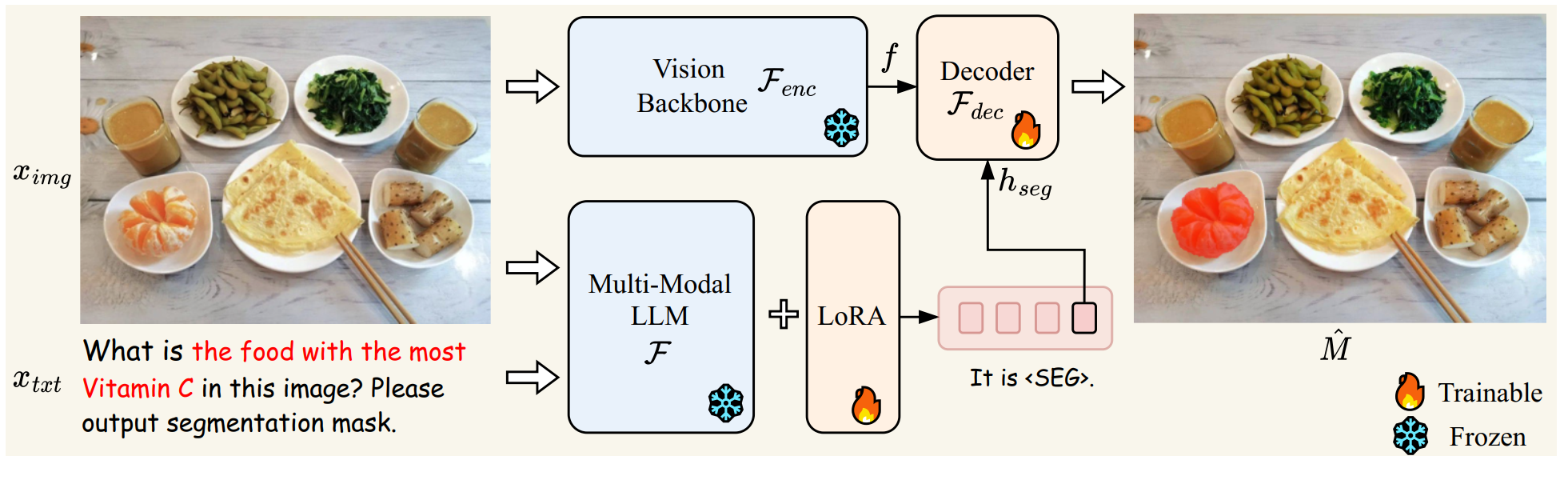}
    \caption{Illustration of the  Large Language Instructed Segmentation Assistant (LISA)~\cite{lai2023lisa}. 
    Figure is from~\cite{lai2023lisa}.}
    \label{fig:LISA}
\end{figure}

Large Language Instructed Segmentation Assistant (LISA)~\cite{lai2023lisa} first proposed a new “reasoning segmentation” task, which aims to generate a segmentation prediction according to a free-form query text that involves complex reasoning. 
Unlike the vanilla referring segmentation task, query texts in reasoning segmentation are more intricate and may involve complex vision and language reasoning or world knowledge. 
This task requires the model to possess the ability to reason the user-specified text queries and the image jointly and produce the expected segmentation predictions. 
As shown in Figure~\ref{fig:LISA}, LISA designed a multimodal Large Language Model (LLM) named LISA to produce segmentation masks based on complex and implicit query texts. 
LISA incorporates a new token, represented as <SEG`>, to signify the request for the segmentation output. 
Using the embedding-as-mask paradigm, LISA has been empowered with segmentation abilities and gains advantages through end-to-end training. 
Thus, the model can handle various scenarios, such as complex reasoning, explanatory answers, and multi-round conversations. 
In addition, LISA has demonstrated strong zero-shot segmentation ability when trained exclusively with reasoning-free data and can be further enhanced via fine-tuning over reasoning segmentation image-instruction pairs.

\subsubsection{Object Detection}
Object detection aims to identify and locate the objects in a given image or video frame.  
In general-purpose multimodal models with visual instruction tuning, object detection involves using visual instructions to guide the model in identifying and localizing objects within an image.

\begin{figure}[t]
    \centering
    \includegraphics[width=0.48\textwidth]{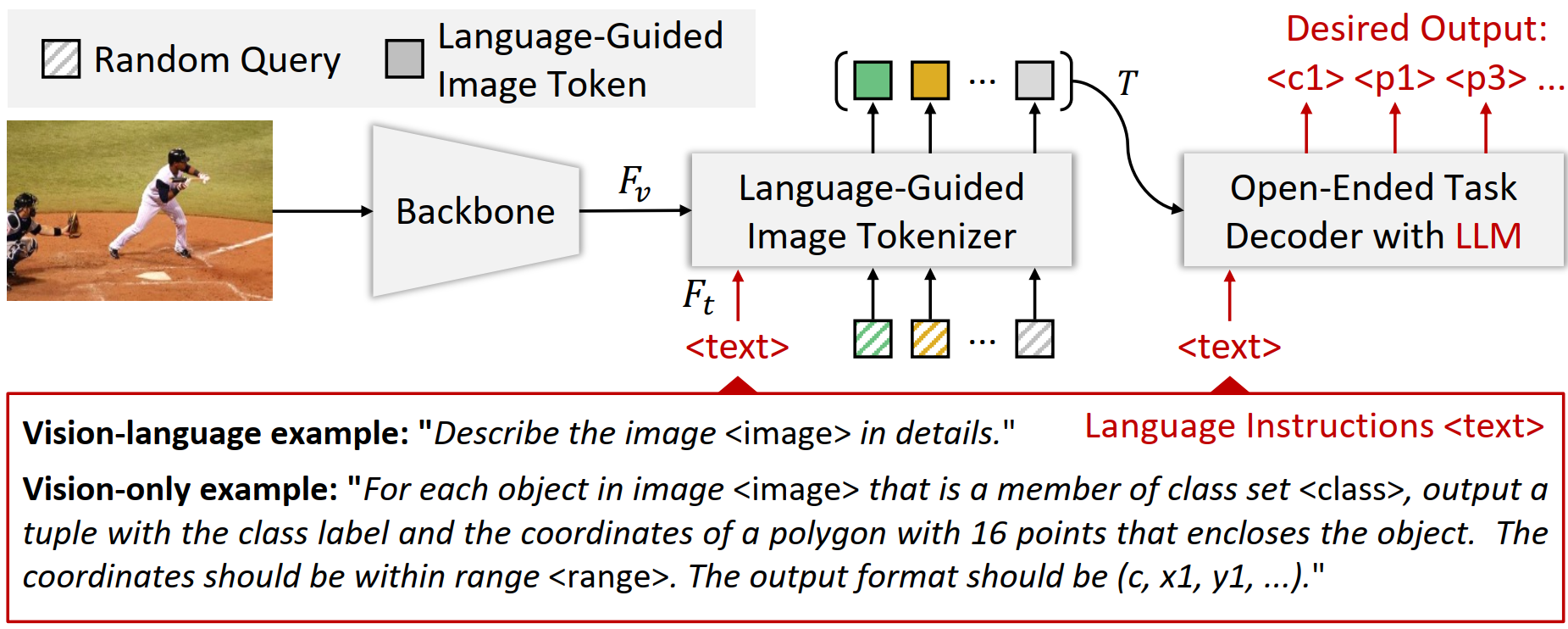}
    \caption{Illustration of the  VisionLLM~\cite{wang2023visionllm}. Figure is from~\cite{wang2023visionllm}.}
    \label{fig:VisionLLM}
\end{figure}

In VisionLMM~\cite{wang2023visionllm}, object detection is one of the vision-centric tasks that the framework is designed to address. 
It leverages LLMs to handle object detection in an instruction-based way which is open-ended and customizable, allowing for the flexible definition and management of object detection tasks using language instructions. 
As shown in Figure~\ref{fig:VisionLLM}, VisionLMM consists of 3 core designs. The first is the language instructions that unify a diverse range of vision tasks and enable flexible task configuration. The second is the Instruction-Aware Image Tokenizer that extracts the required visual information according to the provided language instructions for effective comprehension and parsing of the visual input.
The third one is the LLM-based open-task decoder. It takes inputs the extracted visual embeddings and language instruction embeddings and generates the expected results for various vision tasks. 
VisionLLM enables instruction-based task configuration, such as fine-grained object detection and coarse-grained object detection, and achieves an mAP of over 60\% on the COCO dataset, which places it on par with detection-specific models. 

\begin{figure}[t]
    \centering
    \includegraphics[width=0.48\textwidth]{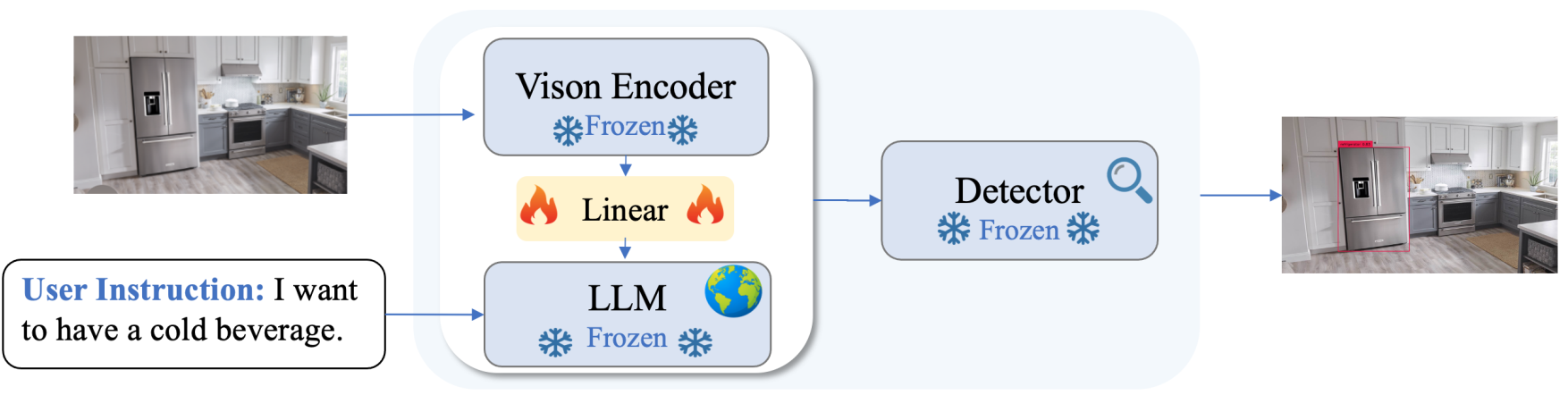}
    \caption{Illustration of the  DetGPT~\cite{pi2023detgpt}. 
    Figure is from~\cite{pi2023detgpt}.}
    \label{fig:OD_DetGPT}
\end{figure}

DetGPT~\cite{pi2023detgpt} introduced a new paradigm for object detection called reasoning-based object detection, which enables the system to reason users' task instructions and visual inputs jointly, allowing it to understand and follow users' intents and conduct object detection accordingly, even if the use's task instruction does not explicitly mention the object.
This paradigm aims to address the limitations of conventional object detection systems by allowing users to use natural languages to express their intents, and the model can reason users' intents and detect the object of interest.
DetGPT involves a two-stage approach for reasoning-based object detection. In the first stage, a multimodal model is used for comprehending the input image, which predicts the related object descriptions that fit the detection instructions specified by users. In the second stage, based on the predicted object descriptions, an open-vocabulary detector is then employed to generate the detection predictions. 
As shown in Figure~\ref{fig:OD_DetGPT}, DetGPT consists of an image encoder for visual feature extraction, and a cross-modal mapping module that maps visual features to the aligned image-text feature space. 
Additionally, it employs a pre-trained large language model to comprehend and reason the visual features and the language instructions jointly, ultimately determining which of the objects could fulfill users' instructions. 
The open-vocabulary object detector then locates the target objects among the results from the multimodal model.

Shikra~\cite{chen2023shikra} focuses on addressing the absence of natural referential ability in current Multimodal Large Language Models (MLLMs) by introducing a unified model capable of handling inputs and outputs of spatial coordinates in natural language form. 
Shikra aims to enable referential dialogue, which is an essential component of everyday human communication and possesses extensive practical applications. 
It is designed to handle tasks related to spatial coordination, such as REC, PointQA, VQA, and Image Captioning, without the need for extra vocabularies, position encoders, or external plug-in models.
Shikra's architecture comprises a vision encoder, an alignment layer, and a Large Language Model (LLM). It uses a pre-trained Vision Transformer as the visual encoder, an alignment layer to align visual and language information, and a large language model to process natural language inputs and generate responses. The design is intentionally simple, without the need for additional vocabularies, position encoders, or external plug-in models.

\begin{figure}[t]
    \centering
    \includegraphics[width=0.48\textwidth]{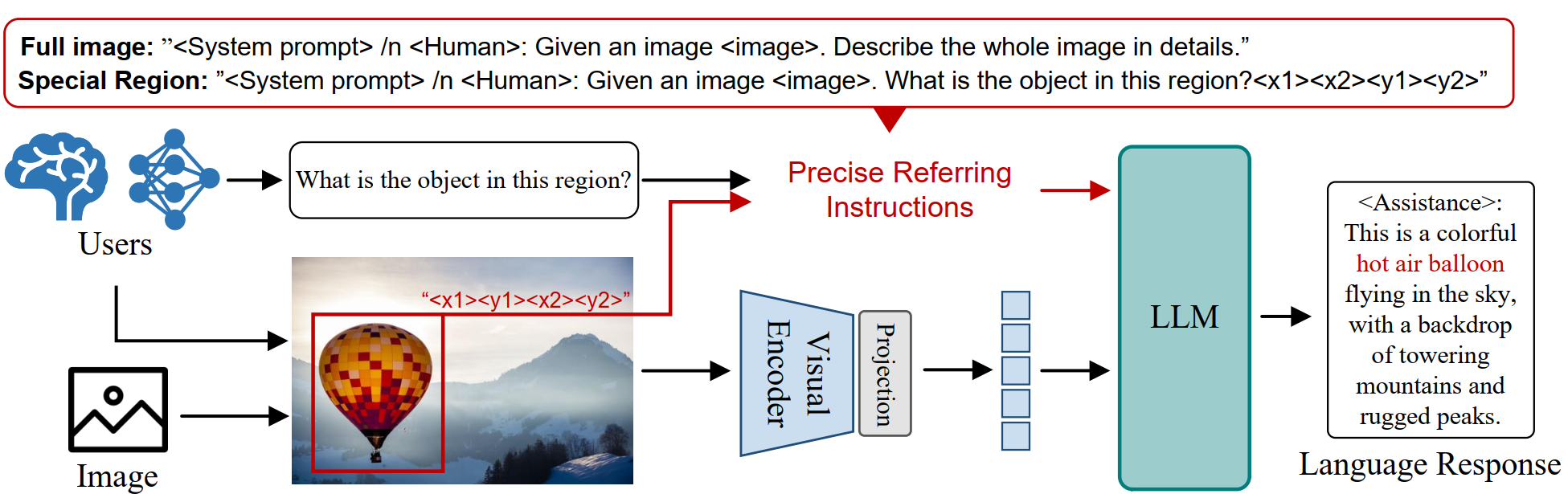}
    \caption{Illustration of the ChatSpot~\cite{zhao2023chatspot}. 
    Figure is from~\cite{zhao2023chatspot}.}
    \label{fig:OD_ChatSpot}
\end{figure}

ChatSpot~\cite{zhao2023chatspot} propose precise referring instruction tuning, which aims to enable multimodal large language models (MLLMs) to support fine-grained interaction. It focuses on utilizing a diverse range of prompts, like points and bounding boxes, as the location prompts to indicate the specific regions of interest (RoIs) in images.
Precise referring instruction tuning improves the flexibility and user-friendliness of the interaction with MLLMs, particularly in the context of vision-language tasks.
As illustrated in Figure~\ref{fig:OD_ChatSpot}, the proposed unified end-to-end multimodal large language model, ChatSpot, comprises 3 main designs: an image encoder, a decoder-only large language model (LLM), and a modality alignment block. The image encoder processes visual inputs, while the LLM handles language understanding and generation. The modality alignment block aligns visual tokens with the language semantic space, enabling seamless integration of vision and language modalities for diverse forms of interaction, including mouse-clicking, drawing boxes, and native language input.
ChatSpot exhibits promising performance on a series of designed evaluation tasks.

\begin{figure}[t]
    \centering
    \includegraphics[width=0.48\textwidth]{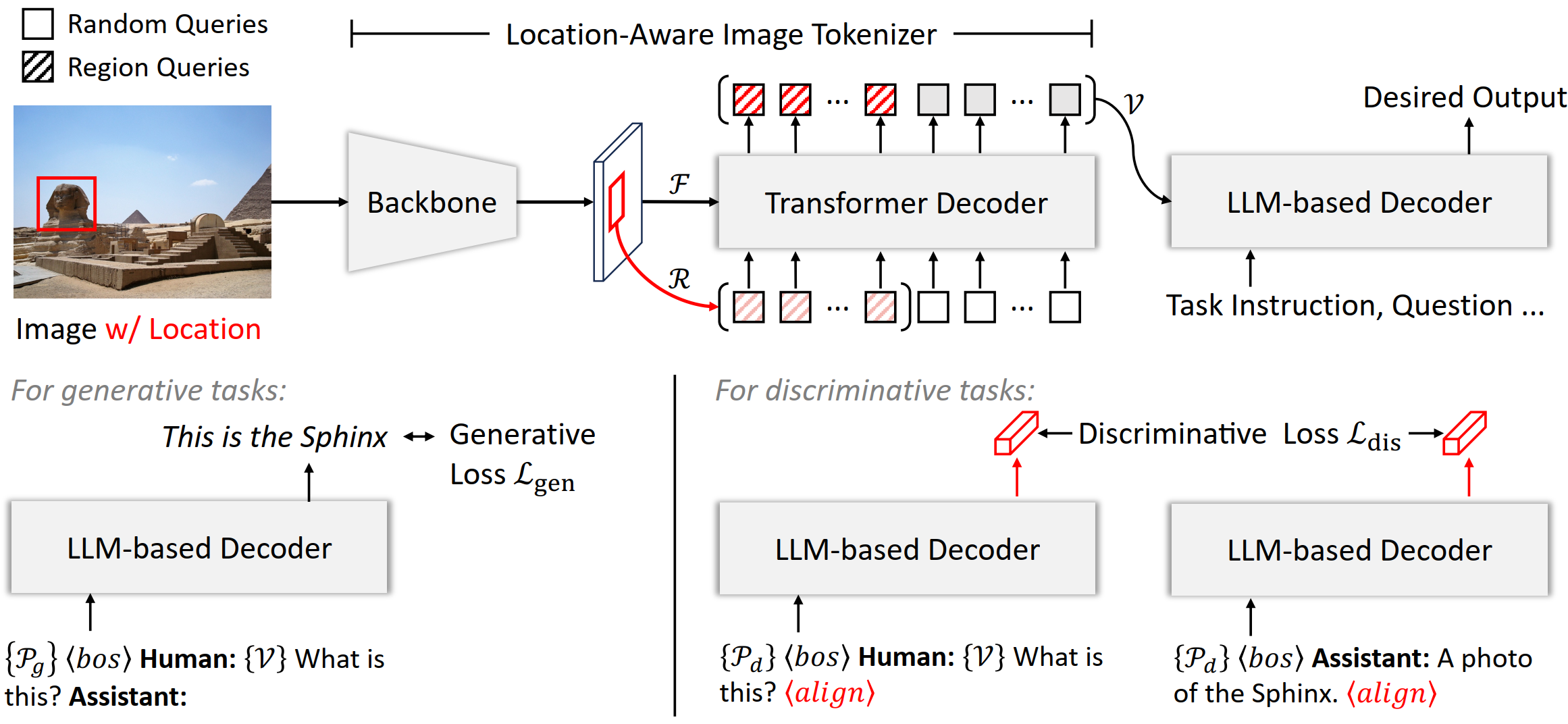}
    \caption{Illustration of the All-Seeing Model (ASM)~\cite{wang2023all}. 
    Figure is from~\cite{wang2023all}.}
    \label{fig:OD_ASM}
\end{figure}

All-Seeing (AS) Project~\cite{wang2023all}, which contributes a large-scale dataset, named AS-1B, for open-world panoptic visual perception as well as the All-Seeing Model, a universal vision-language model capable of recognizing and understanding context in arbitrary regions. 
As shown in Figure~\ref{fig:OD_ASM}, The All-Seeing Model (ASM) consists of two modules including a position-aware image tokenizer and an LLM-based decoder.
The first module encodes image conditioned the location information represented as bounding boxes, masks, and points, which empowers ASM with the  location ability. 
As the second module inherits world knowledge and reasoning ability from the pre-trained LLMs, it can provide a robust foundation for visual perception. 
Additionally, ASM designs a special prompt to enable the model to switch to and handle generative or discriminative vision tasks accordingly.
The ASM model demonstrates remarkable zero-shot performance in various vision and language tasks, including regional retrieval, recognition, captioning, and question-answering, and is evaluated on representative vision and vision-language tasks.

\begin{figure}[t]
    \centering
    \includegraphics[width=0.48\textwidth]{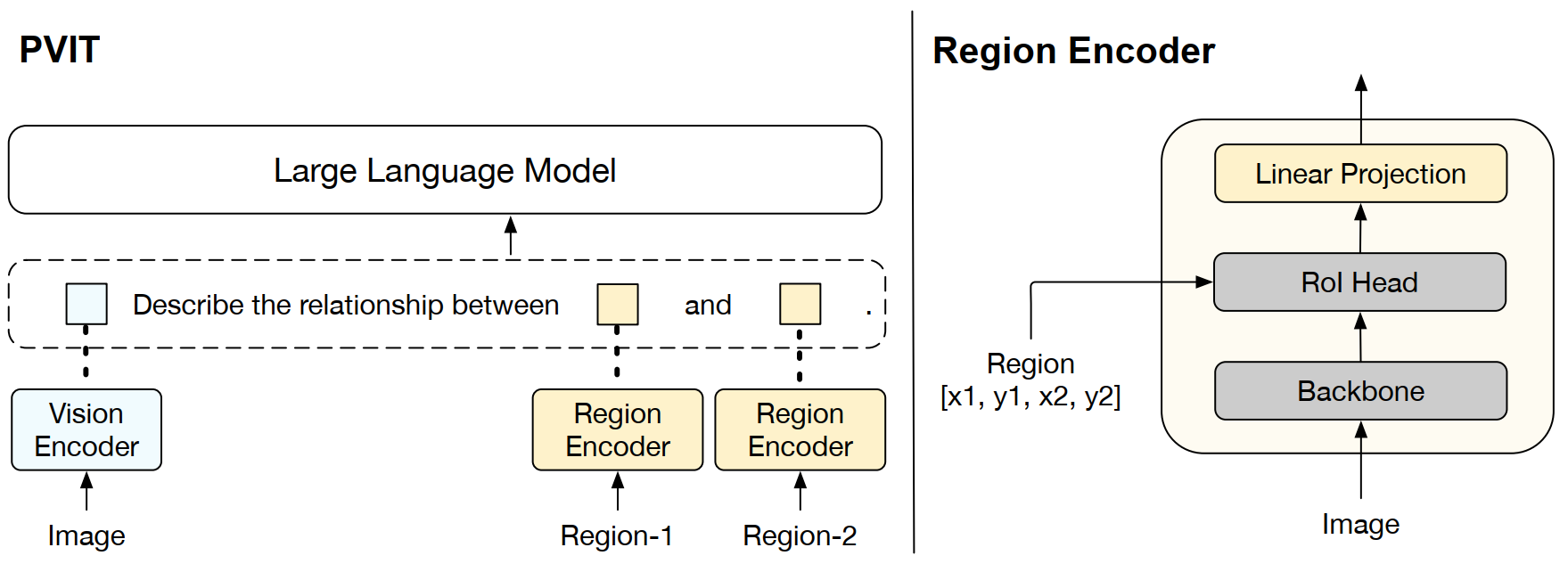}
    \caption{Illustration of the Position-enhanced Visual Instruction Tuning (PVIT)~\cite{chen2023position}. 
    Figure is from~\cite{chen2023position}.}
    \label{fig:OD_PVIT}
\end{figure}

PVIT~\cite{chen2023position} introduces Position-enhanced Visual Instruction Tuning (PVIT), which extends the capabilities of Multimodal Large Language Models (MLLMs) by integrating an additional region-level vision encoder. The proposed method also includes a region-level instruction data construction scheme and an evaluation dataset to facilitate the training and evaluation of PVIT.
The model architecture of PVIT is illustrated in Figure~\ref{fig:OD_PVIT}, it consists of three primary components: a vision encoder, a region encoder, and a large language model (LLM). The model processes an input image together with instructions containing embedded regions and generates corresponding responses. The region encoder is responsible for extracting region-level features from the image and regions, which are then integrated into the large language model for fine-grained multimodal instruction tuning.
The stage training strategy of PVIT involves an initial stage where a linear projection layer is trained to align region features with the embedding space of the large language model (LLM). In the second stage, the model is fine-tuned with region-level instruction data to adapt to complex fine-grained instructions. This approach allows the model to first learn to understand region features and then enhance its capabilities in following instructions that contain regions.

\subsubsection{Visual Grounding}
Object detection involves identifying and locating objects within an image by classifying and locating them. In contrast, visual grounding goes a step further by linking specific regions or objects within an image to textual descriptions, requiring comprehension of the context and semantics of both the visual scene and the associated language. 
Visual instruction tuning for visual grounding aims to enable the system to understand finer-grained context, attributes, and the relationships between objects as described in the text, effectively bridging the gap between visual perception and linguistic representation.

\begin{figure}[t]
    \centering
    \includegraphics[width=0.48\textwidth]{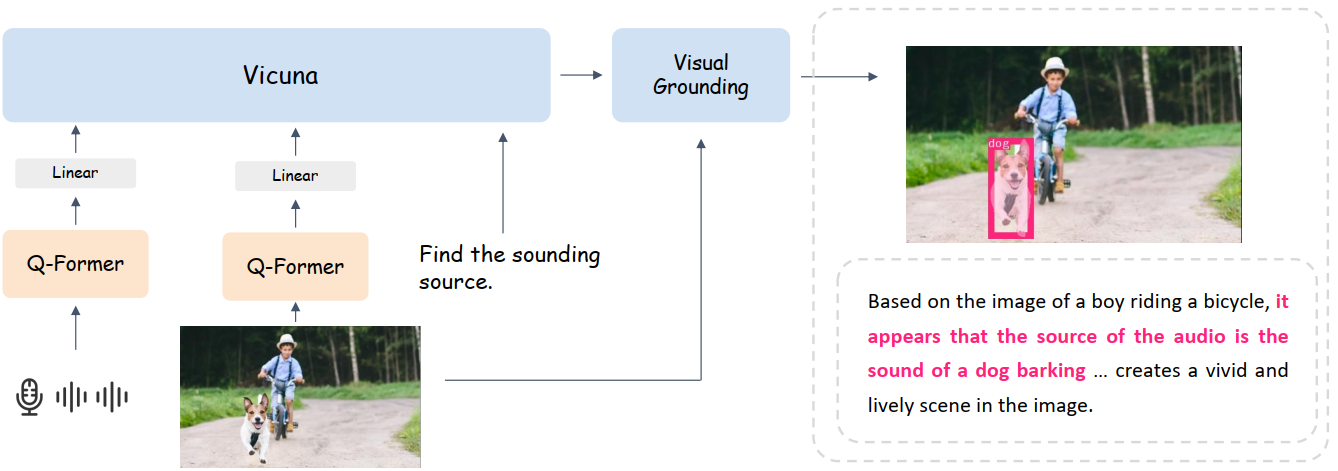}
    \caption{Illustration of the BuboGPT~\cite{zhao2023bubogpt}. 
    Figure is from~\cite{zhao2023bubogpt}.}
    \label{fig:VG_BuboGPT}
\end{figure}

BuboGPT~\cite{zhao2023bubogpt}, a multimodal language model with visual grounding capabilities, enabe a fine-grained understanding of visual objects and other modalities. It proposes an off-the-shelf visual grounding pipeline and a two-stage training scheme for joint multimodal understanding. Additionally, the paper constructs a high-quality multimodal instruction-tuning dataset, facilitating the model's ability to recognize and respond to arbitrary combinations of input modalities.
As shown in Figure~\ref{fig:VG_BuboGPT}, the model architecture of BuboGPT consists of a multimodal language model that integrates visual grounding capabilities. It employs a visual grounding pipeline with tagging, grounding, and entity-matching modules to establish fine-grained relations between visual objects and other modalities. Additionally, BuboGPT uses a two-stage training scheme to align vision and audio features with language and performs multimodal instruction tuning on a high-quality dataset to enable joint multimodal understanding.

\begin{figure}[t]
    \centering
    \includegraphics[width=0.48\textwidth]{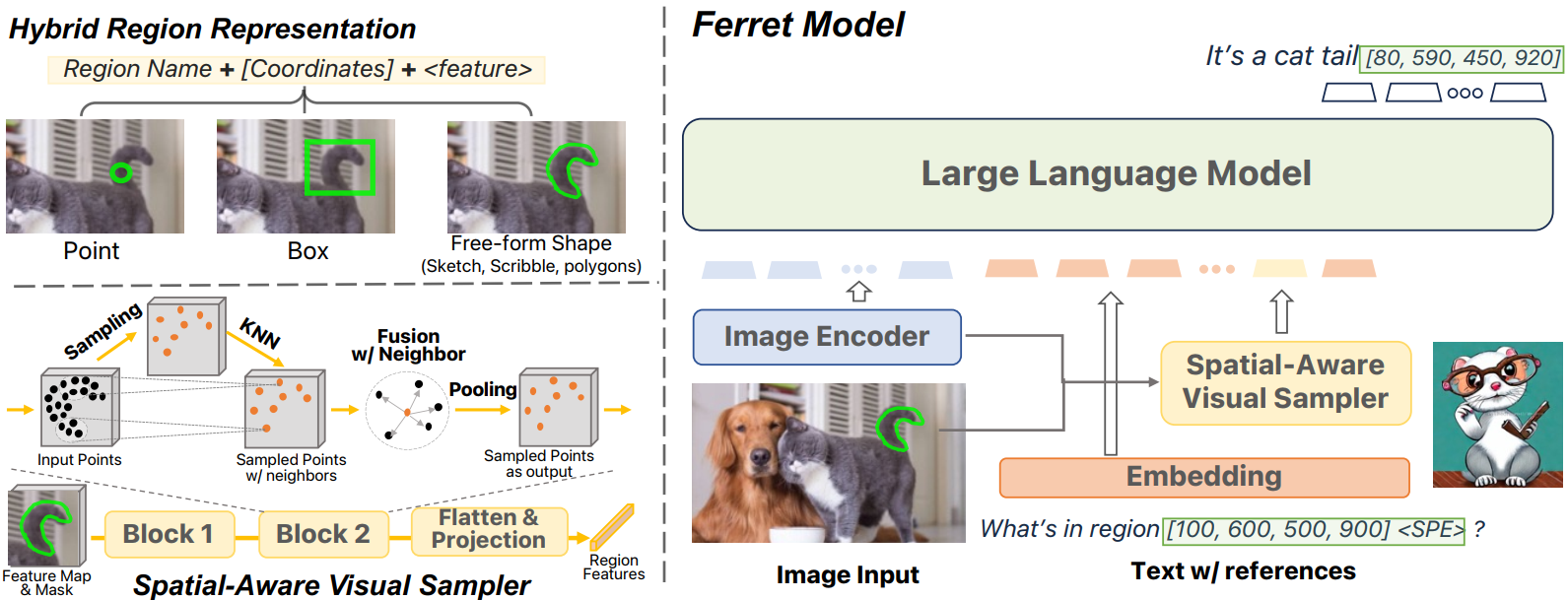}
    \caption{Illustration of the Ferret~\cite{you2023ferret}. 
    Figure is from~\cite{you2023ferret}.}
    \label{fig:VG_FERRET}
\end{figure}

Ferret~\cite{you2023ferret} introduces a Multimodal Large Language Model (MLLM) that can understand spatial referring and accurately ground open-vocabulary descriptions within an image. 
Ferret proposes a novel hybrid region representation that combines discrete coordinates with continuous visual features to refer to regions of various shapes and formats within an image, 
This representation allows Ferret to flexibly handle inputs that mix referred regions with free-form texts and accurately ground the mentioned objects in its outputs.
As shown in Figure~\ref{fig:VG_FERRET}, the model architecture of Ferret consists of an image encoder to extract image embeddings, a spatial-aware visual sampler to extract regional continuous features, and a Large Language Model (LLM) to jointly model image, text, and region features. This architecture enables Ferret to process diverse region inputs, such as points, bounding boxes, and free-form shapes, and accurately ground open-vocabulary descriptions.
Ferret demonstrates superior performance in various tasks and reduces object hallucination.

\begin{figure}[t]
    \centering
    \includegraphics[width=0.48\textwidth]{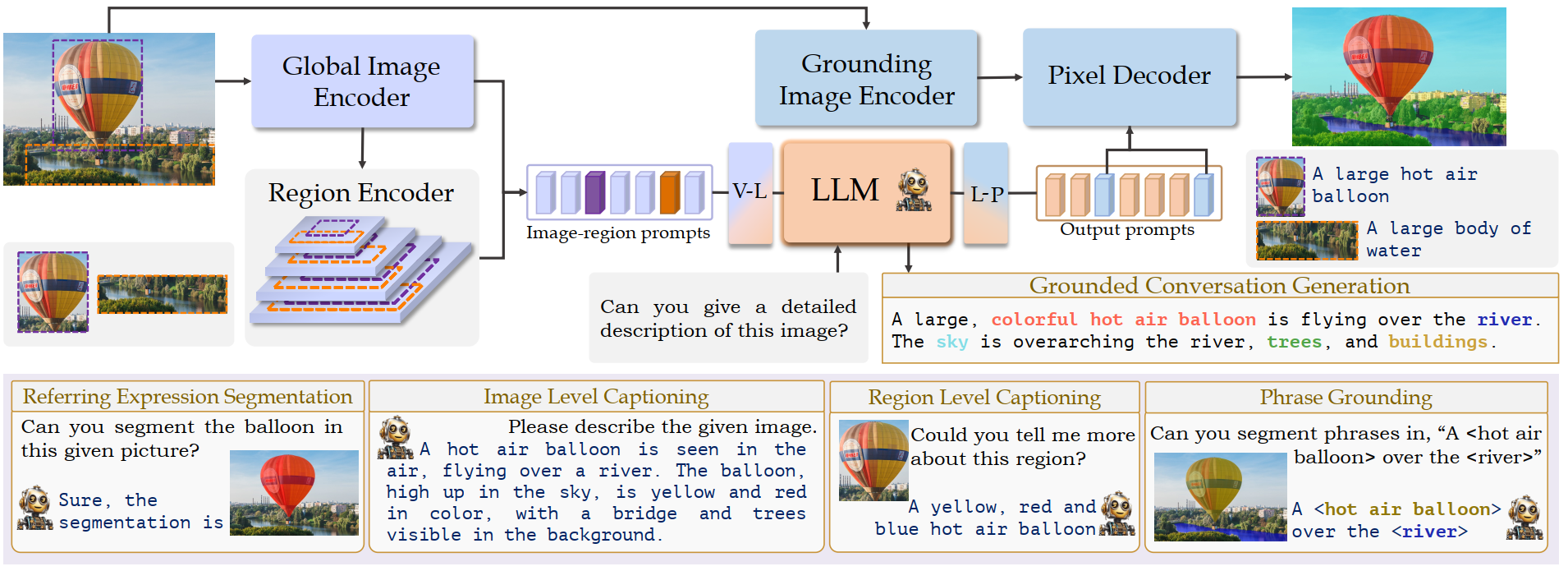}
    \caption{Illustration of the GLaMM~\cite{rasheed2023glamm}. 
    Figure is from~\cite{rasheed2023glamm}.}
    \label{fig:VG_GLaMM}
\end{figure}

GLaMM~\cite{rasheed2023glamm} introduces a new task called Grounded Conversation Generation (GCG), aiming to generate natural language responses that are seamlessly integrated with object segmentation masks.
It requires generating image descriptions with the phrases or words that are linked to the corresponding segmentation masks, thereby bridging the gap between textual and visual understanding. 
Moreover, it proposes GLaMM, which is first model that has the ability to generate natural language responses that involves segmentation masks. 
As shown in Figure~\ref{fig:VG_GLaMM}, GLaMM consists of five core components: a global image encoder, a regional image encoder, a large language model (LLM), a grounding image encoder, and a pixel decoder. 
With the above modules, it enables the model to accept text and visual inputs, and interact at multiple levels of granularity and generate grounded textual outputs accordingly. 
In summary, this architecture enables image-level, region-level and pixel-level understand and perception.
GLaMM demonstrates superior performance on its created Grounding-anything Dataset (GranD) and designed evaluation protocol.

\subsection{Instruction-based Image Learning for Generative Tasks}
Instruction-based learning for generative tasks in multimodal models has gained significant attention. This approach involves constructing high-quality instruction-following datasets and designing instruction-tuning methods to enhance large language models. These models acquire multi-turn, interleaved multimodal instruction-following capabilities, enabling them to perform advanced multimodal tasks, including image generation and editing.

\subsubsection{Image Generation}

\begin{figure}[t]
    \centering
    \includegraphics[width=0.48\textwidth]{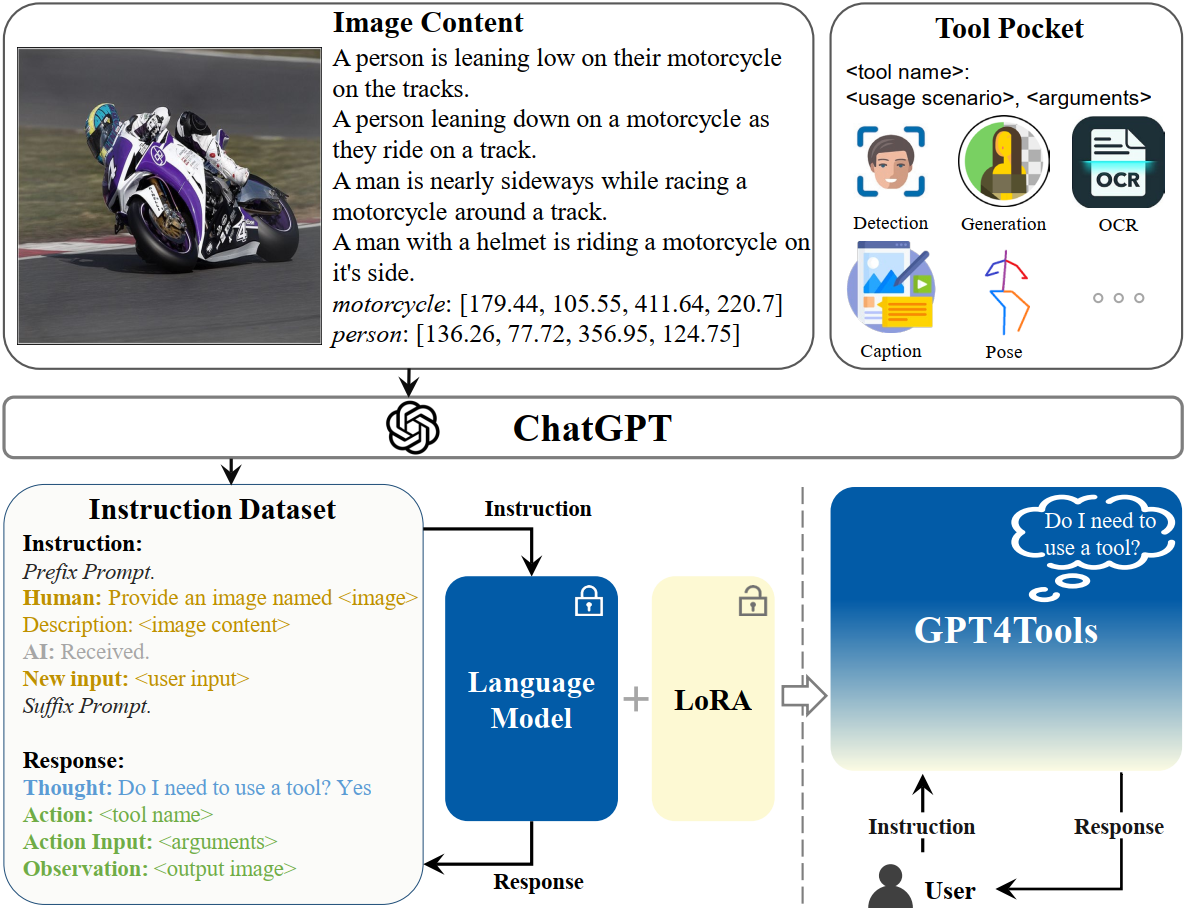}
    \caption{Illustration of the GPT4Tools~\cite{yang2023gpt4tools}. 
    Figure is from~\cite{yang2023gpt4tools}.}
    \label{fig:IG_GPT4Tools}
\end{figure}

GPT4Tools~\cite{yang2023gpt4tools} enables open-source language models to effectively use multimodal tools. It constructs a tool-related instructional dataset from advanced language models and utilizes Low-Rank Adaptation (LoRA) optimization to enhance the language models' tool-usage capabilities. Additionally, it proposes a benchmark to evaluate the accuracy of language models in using tools, demonstrating significant improvements in tool usage across various visual tasks.
As shown in Figure~\ref{fig:IG_GPT4Tools}, the GPT4Tools framework involves constructing a tool-related instruction dataset by prompting an advanced language model with various multimodal contexts. This dataset is then used to fine-tune open-source language models using Low-Rank Adaptation (LoRA) optimization, enabling them to effectively use tools for visual tasks such as comprehension and image generation. Additionally, the framework includes a benchmark to evaluate the language models' ability to use tools, showcasing significant improvements in tool usage accuracy.

\begin{figure}[t]
    \centering
    \includegraphics[width=0.48\textwidth]{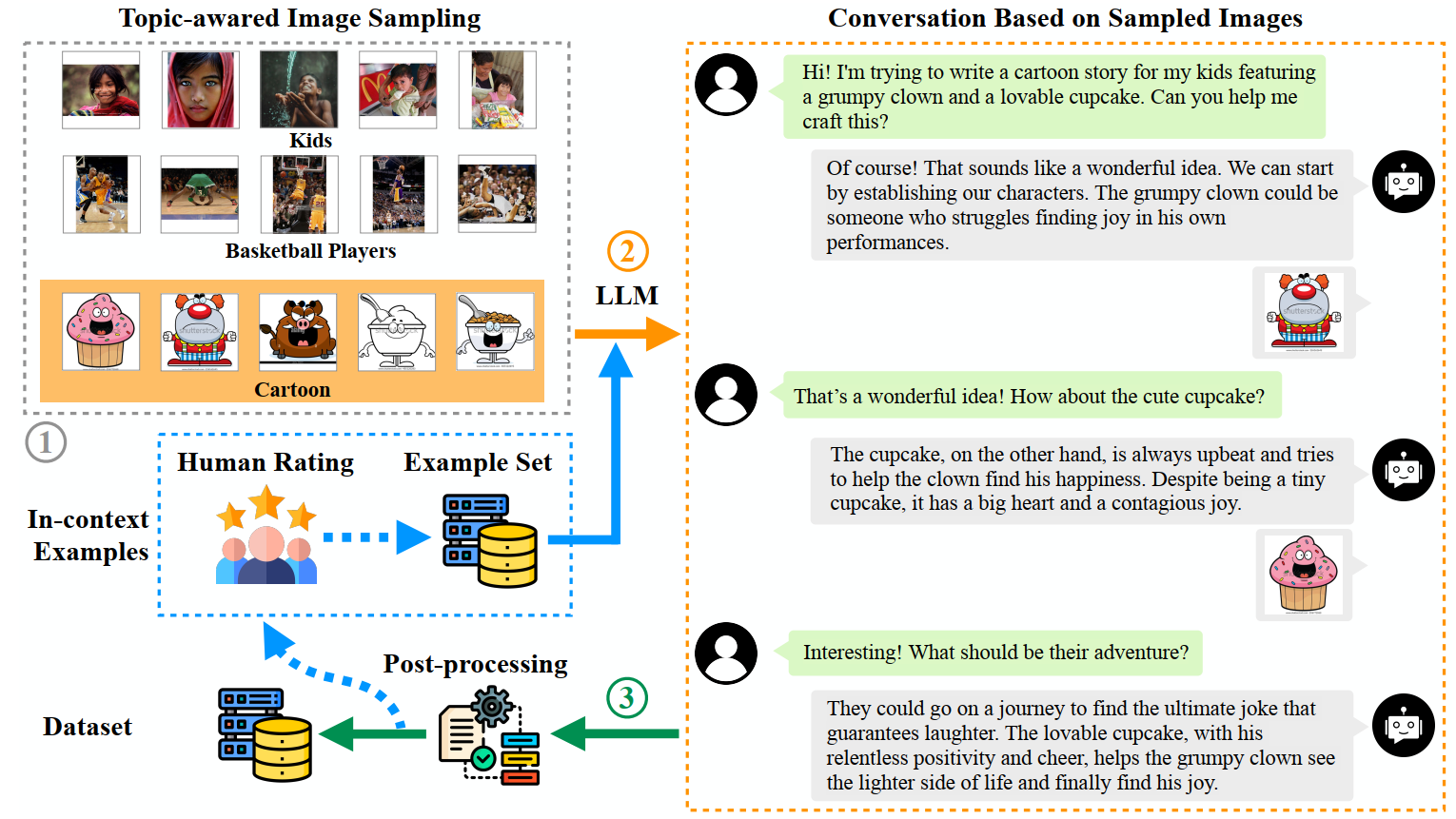}
    \caption{Illustration of the GPT4Tools~\cite{li2023textbind}. 
    Figure is from~\cite{li2023textbind}.}
    \label{fig:IG_TEXTBIND}
\end{figure}

TextBind~\cite{li2023textbind} enhances large language models with multi-turn interleaved multimodal instruction-following capabilities. It significantly reduces the need for high-quality exemplar data, making it more accessible and scalable for real-world tasks. The proposed model, MIM, trained on TextBind, outperforms recent baselines in open-world multimodal conversations, demonstrating remarkable performance in textual response generation, image generation, and overall multimodal instruction-following.
As shown in Figure~\ref{fig:IG_TEXTBIND}, MIM seamlessly integrates image encoder and decoder models to accommodate interleaved image-text inputs and outputs. It supplements large language models with visual input and output modules, enabling the model to process multi-turn interleaved multimodal instructions and generate coherent responses. The architecture is trained in two stages, focusing on aligning the feature spaces of vision and language models and further improving instruction-following capabilities.

\subsubsection{Image Editing}

\begin{figure}[t]
    \centering
    \includegraphics[width=0.3\textwidth]{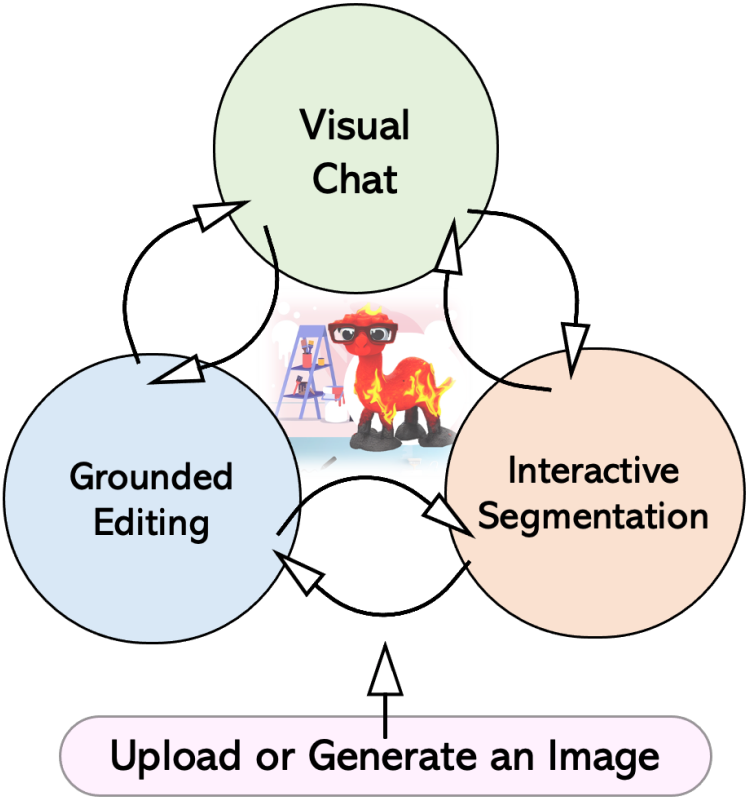}
    \caption{Illustration of the LLaVA-Interactive~\cite{chen2023llava}. 
    Figure is from~\cite{chen2023llava}.}
    \label{fig:IE_LLaVA_Interactive}
\end{figure}

LLaVA-Interactive makes significant contributions to the field of multimodal human-AI interaction by providing a cost-efficient and versatile system for multi-turn dialogues with human users. It combines visual and language prompts, enabling sophisticated multimodal tasks such as image editing, segmentation, and generation. Additionally, LLaVA-Interactive addresses technical challenges in system development and demonstrates its capabilities across a wide range of real-world application scenarios, showcasing its potential for performing new, complex tasks in various domains.
As shown in Figure~\ref{fig:IE_LLaVA_Interactive}, the workflow of LLaVA-Interactive involves several key steps for visual creation processes. It begins with image input, where users can upload an image or generate one by providing a language caption and drawing bounding boxes to establish the spatial arrangement of objects. Users can then engage in visual chat, interactive segmentation, and grounded editing to iteratively refine their visual creations. This multi-turn interaction allows users to ask questions, create object masks, place new objects on the image, and make adjustments to achieve their intended visual outcomes.

\subsection{Instruction-based Image Learning for Complex Reasoning Tasks}

\subsubsection{Image Captioning}
Image captioning involves training models to understand the content of an image and generate a natural language description that accurately represents the visual content. This task requires integrating computer vision techniques for image understanding with natural language processing methods for language generation. The goal is to enable machines to describe the visual content of an image in a human-like manner, allowing for better understanding and interpretation of visual information.
Visual instruction tuning improves the task of image captioning by providing a fine-tuning process with specifically devised and fine-grained multimodal instruction sets.  This allows the model to associate system instructions and text queries with input multimodal contexts, enhancing its ability to generate accurate and relevant captions for images.

\begin{figure}[t]
    \centering
    \includegraphics[width=0.48\textwidth]{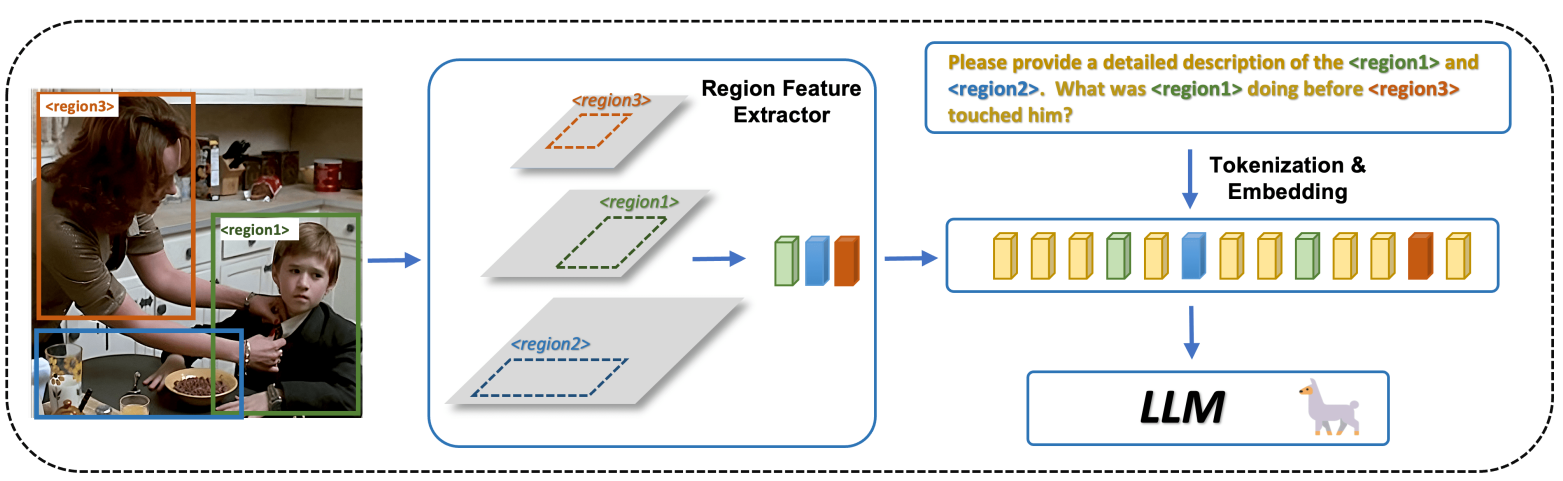}
    \caption{Illustration of the GPT4RoI~\cite{zhang2023gpt4roi}. 
    Figure is from~\cite{zhang2023gpt4roi}.}
    \label{fig:IC_GPT4ROI}
\end{figure}

GPT4RoI introduces spatial instruction tuning for large language models on region-of-interest (RoI) in image-text pairs.  This model allows users to interact with both language and drawing bounding boxes to adjust referring granularity, and it can mine a variety of attribute information within each RoI.  GPT4RoI is trained on 7 region-text pair datasets and brings an unprecedented interactive and conversational experience compared to previous image-level models, enhancing fine-grained multimodal understanding.
As shown in Figure~\ref{fig:IC_GPT4ROI}, GPT4RoI assist the task of image captioning by allowing models to incorporate references to specific regions of interest (RoI) in the image.  This enables the models to generate captions that are more detailed and specific to particular regions within the image.  By aligning language instructions with RoI features, visual instruction tuning enhances the model's ability to understand and describe fine-grained visual details, leading to more accurate and informative image captions.

MiniGPT-4 is a model that aligns visual embedding space with a popular LLM, Vicuna, to achieve advanced vision-language abilities. The model demonstrates the ability to generate detailed image descriptions, create websites from hand-drawn drafts, write stories and poems inspired by images, and provide cooking recipes from food photos. MiniGPT-4 also highlights the importance of fine-tuning the model with a detailed image description dataset to enhance the naturalness of the produced languages and their usability.

\begin{figure}[t]
    \centering
    \includegraphics[width=0.48\textwidth]{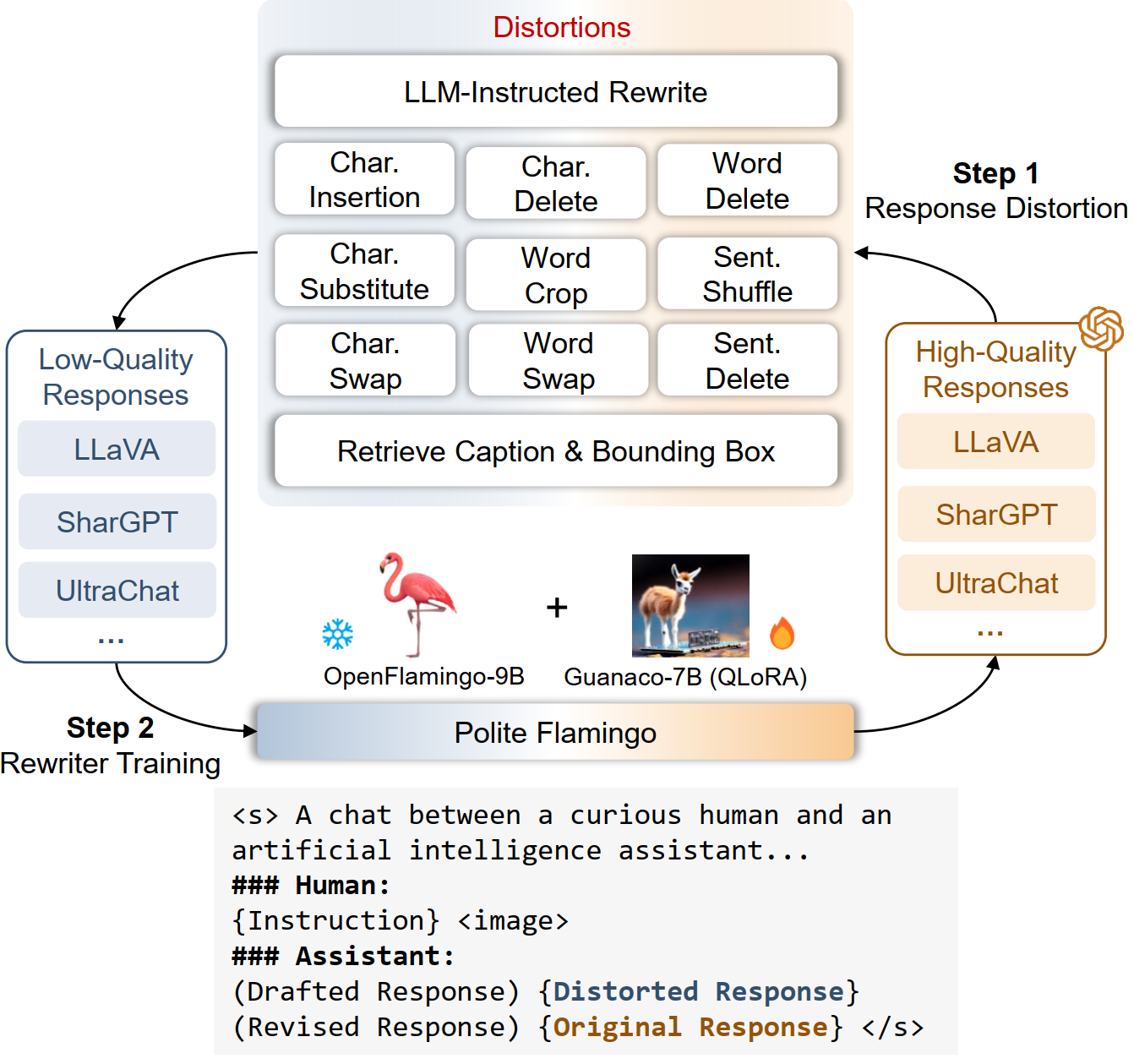}
    \caption{Illustration of the Clever Flamingo~\cite{chen2023visual}. 
    Figure is from~\cite{chen2023visual}.}
    \label{fig:IC_CleverFlamingo}
\end{figure}

Clever Flamingo a novel method to curate raw vision-language datasets into visual instruction tuning data, reducing the ``multimodal alignment tax''.  It constructs a large-scale visual instruction tuning dataset based on response rewriting and introduces a U-shaped multi-stage visual instruction tuning approach.  It also demonstrates the advantages of the resulting model in terms of both multimodal understanding and response politeness.
As shown in Figure~\ref{fig:IC_CleverFlamingo}, the U-shaped multi-stage visual instruction tuning approach involves three stages.  In Stage 1, the focus is on improving the instruction-following ability by tuning only the language model.  Stage 2 shifts to improving the visual understanding capability by exclusively tuning the connector.  Finally, in Stage 3, the model is fine-tuned to recover the optimal politeness of the responses.  This approach aims to enhance the model's multimodal understanding and response politeness efficiently.

DreamLLM is a learning framework that introduces a versatile Multimodal Large Language Model (MLLM) capable of generating free-form interleaved content and excelling at zero-shot or in-context vision-language comprehension and synthesis tasks. It operates on the principles of generative modeling of language and image posteriors, as well as fostering the generation of raw, interleaved documents, allowing it to learn all conditional, marginal, and joint multimodal distributions effectively. The contribution of DreamLLM lies in demonstrating the effectiveness of achieving enhanced learning synergy between multimodal content understanding and creation, paving the way for further research in the multimodal machine learning field.

AnyMAL is a unified model designed to reason over diverse input modality signals and generate textual responses. It presents an efficient and scalable solution for building Multimodal LLMs, fine-tuning the model with a multimodal instruction set covering diverse tasks, and achieving strong zero-shot performance in both automatic and human evaluations on various multimodal tasks. Additionally, AnyMAL extends previous approaches by allowing for diverse input modalities beyond vision signals and scaling the LLM parameters to 70B via an efficient pre-training approach.

\subsubsection{Visual Question Answering}
Visual Question Answering (VQA) combines image understanding and natural language processing to answer questions about the content of a given image. In this task, a user presents an image along with a question in natural language that refers to some aspect of the image. The VQA model then analyzes the visual data to understand the scene, identifies relevant components, and processes the text of the question. Finally, it generates an accurate and relevant answer based on the synthesis of these two streams of information. The challenge for the VQA is to correctly interpret the visual cues and the context of the question, which requires a deep understanding of both the visual elements in the image and the semantics of the question.
Visual instruction tuning improves the performance of the VQA model by enabling efficient adaptation of large language models to effectively process and integrate visual instructions, leading to enhanced reasoning ability and accurate responses in VQA tasks.

\begin{figure}[t]
    \centering
    \includegraphics[width=0.48\textwidth]{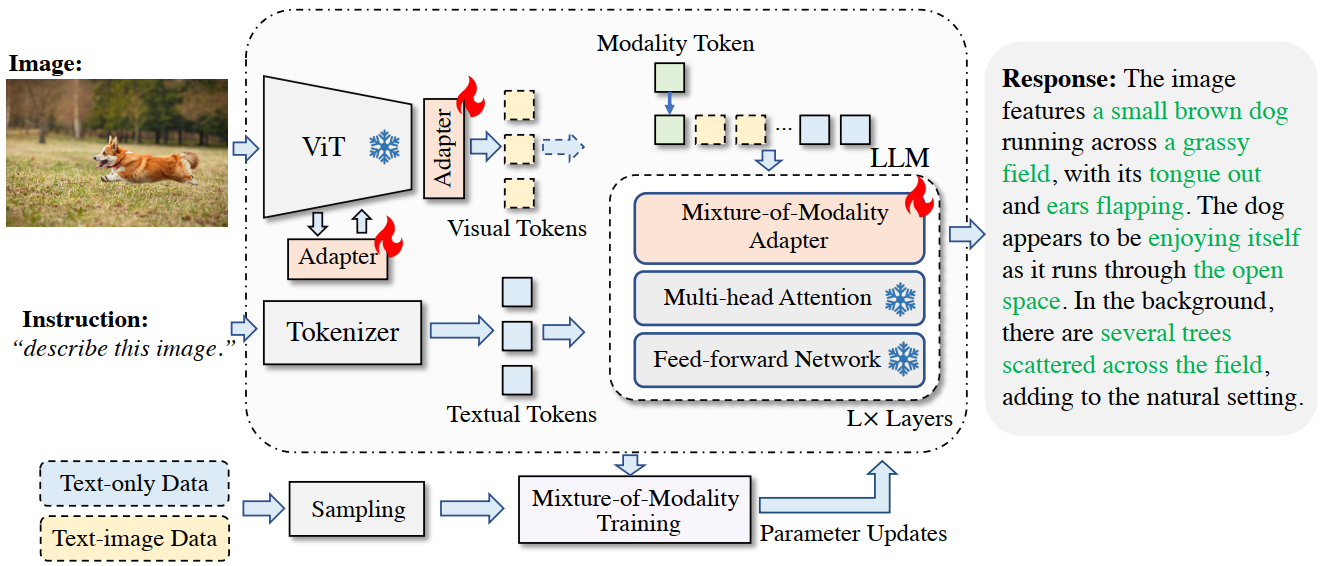}
    \caption{Illustration of LaVIN~\cite{luo2023cheap}. 
    Figure is from~\cite{luo2023cheap}.}
    \label{fig:VQA_LaVIN}
\end{figure}

LaVIN proposes a novel and efficient solution for vision-language instruction tuning called Mixture-of-Modality Adaptation (MMA). This approach enables the joint optimization of multimodal large language models (LLMs) with a small number of parameters, significantly reducing training costs. The proposed MMA equips LLMs with lightweight adapters and a routing scheme to dynamically choose adaptation paths for different modalities, resulting in a large vision-language instructed model called LaVIN. 
As shown in Figure~\ref{fig:VQA_LaVIN}, LaVIN employs a simplified and lightweight architecture that incorporates Mixture-of-Modality Adapters (MM-Adapters) to process instructions from different modalities. These MM-Adapters connect the large language model (LLM) with the image encoder, enabling efficient adaptation to vision-language tasks. The architecture is optimized through Mixture-of-Modality Training (MMT) in an end-to-end manner, allowing LaVIN to effectively execute input instructions from various modalities while demonstrating superior performance in vision-language tasks.

SciTune focuses on aligning large language models (LLMs) with scientific disciplines, concepts, and goals. The framework includes two stages: scientific concept alignment and scientific instruction tuning. By training LLaMA-SciTune models on science-focused multimodal tasks, the paper demonstrates improved performance in visual grounded language understanding and multimodal reasoning, surpassing human performance in the ScienceQA benchmark. Additionally, the paper emphasizes the use of human-generated scientific multimodal instructions to align LLMs with natural scientific concepts and true human intent.

MultiInstruct leverages instruction tuning to improve the generalizability of Vision-Language pretrained models on multimodal and vision tasks. It introduces new metrics such as Sensitivity to measure the model's capability to consistently produce results regardless of slight variations in instructions. MultiInstruct demonstrates strong zero-shot performance on various unseen multimodal tasks and highlights the potential benefits of larger text-only instruction datasets for multimodal instruction tuning.

LMEye is a human-like eye with a play-and-plug interactive perception network designed to enable dynamic interaction between Large Language Models (LLMs) and external vision information. LMEye significantly improves zero-shot multimodal performances for various scales and types of LLMs, demonstrating superior performance on evaluation benchmarks for multimodal LLMs, visual question answering, in-detail image description, and multimodal reasoning tasks. Additionally, LMEye addresses the limitations and challenges associated with MLLMs, such as generating toxic or biased content, and proposes potential improvement solutions.

VPG-C aims to enhance the ability of Multimodal Large Language Models (MLLMs) to comprehend demonstrative instructions with interleaved multimodal context. The proposed VPG-C module infers and completes missing visual details, and it also introduces a synthetic discriminative training strategy to fine-tune VPG-C without the need for supervised demonstrative instruction data. Additionally, it introduces a comprehensive benchmark called DEMON for evaluating MLLMs on 31 tasks with complex vision-language demonstrative context. The results show that VPG-C achieves notable zero-shot performance on the DEMON benchmark and demonstrates superior performance on established benchmarks like MME and OwlEval.

BLIVA is a multimodal Large Language Model that leverages learned query embeddings and encoded patch embeddings to enhance text-image visual perception and understanding. BLIVA demonstrates superior performance in both general and text-rich Visual Question Answering (VQA) benchmarks, showcasing exceptional OCR capabilities and robust localization ability. The model's innovative design bolsters performance in academic benchmarks and real-world examples, highlighting its effectiveness in handling text-rich visual questions.

MiniGPT-v2 is a unified interface for vision-language multi-task learning. It is designed to effectively handle various vision-language tasks, such as image description, visual question answering, and visual grounding, using a single architecture. Its key innovations include the use of unique identifiers for different tasks during training, enabling the model to distinguish and learn multiple tasks efficiently, and achieving state-of-the-art results on diverse vision-language benchmarks.

The mPLUG-Owl2 is a multimodal foundation model that revolutionizes large language models by incorporating modality collaboration and interference mitigation. It features a modularized network design, a modality-adaptive module, and a two-stage training paradigm to effectively manage multimodal signals. It achieves state-of-the-art performance on vision-language benchmarks, demonstrates adaptability in zero-shot multimodal tasks, and also excels in pure-text benchmarks. It also provides in-depth analysis and validation of the impact of modality collaboration and offers insights into the effectiveness of the proposed training paradigm for future multimodal foundation models.

\begin{figure}[t]
    \centering
    \includegraphics[width=0.48\textwidth]{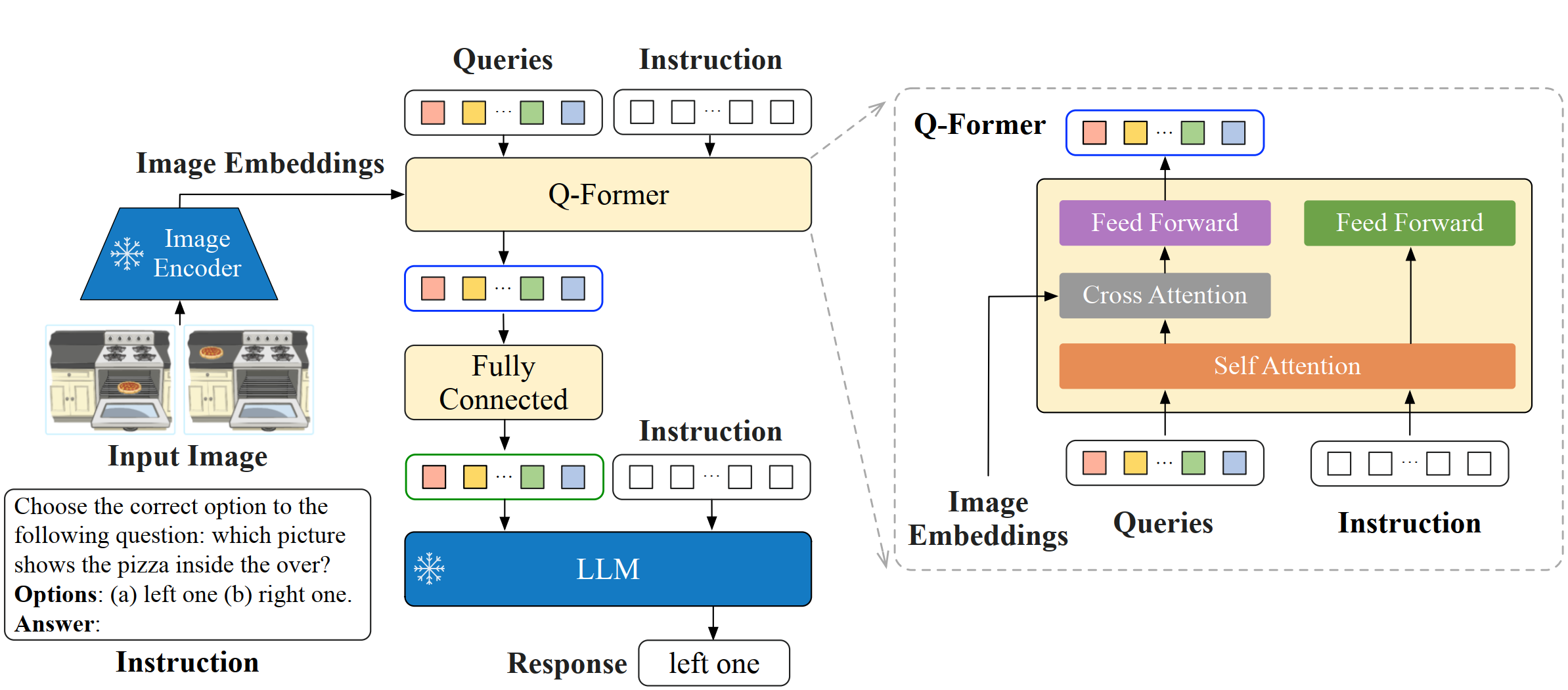}
    \caption{Illustration of InstructBLIP~\cite{dai2023instructblip}. 
    Figure is from~\cite{dai2023instructblip}.}
    \label{fig:VQA_InstructBLIP}
\end{figure}

InstructBLIP~\cite{dai2023instructblip} is a visual instruction tuning pipeline, which help construct a general-purpose multimodal model that can handle a broad range of vision tasks via a universal task interface with languages as task instructions. 
As shown in Figure~\ref{fig:VQA_InstructBLIP}, InstructBLIP consists of a Query Transformer (Q-Former) that extracts instruction-aware visual features from the output embeddings of a frozen image encoder. These visual features are then fed as soft prompt input to a frozen Language Model (LLM). During instruction tuning, the Q-Former is finetuned while the image encoder and LLM remain frozen. This architecture allows for the extraction of task-relevant visual features based on the given instructions, enhancing the model's ability to follow instructions and generate responses.
With a comprehensive study on vision-language instruction tuning, it demonstrates the effectiveness of InstructBLIP on zero-shot generalization to unseen tasks. The framework achieves state-of-the-art performance on a diverse set of vision-language tasks and provides novel techniques for instruction-aware visual feature extraction and balanced dataset sampling.

InternLM-XComposer is a vision-language large model that excels in advanced image-text comprehension and composition. Its key innovations lie in three main areas: 1) Interleaved Text-Image Composition, allowing seamless integration of images into coherent articles, 2) Comprehension with Rich Multilingual Knowledge, enabling deep understanding of visual content across diverse domains, and 3) State-of-the-art Performance, consistently achieving top results in various vision-language benchmarks. Additionally, it introduces a novel evaluation procedure for assessing the quality of interleaved text-image articles. 

\subsubsection{Visual Assistant}
Visual assistant typically refers to a system or application that uses computer vision and machine learning algorithms to understand and process visual information, such as images, in conjunction with language. It is capable of interpreting visual content and responding to queries or instructions related to the visual input. 
Instruction-based Image Learning enhances the ability of visual assistants to understand and follow multimodal vision-and-language instructions, improving the adaptability to user instructions, and ultimately leading to performance improvements in multimodal tasks and instruction-following capabilities. This process contributes to the development of a more capable and adaptable visual assistant, enabling it to effectively process and respond to both visual and language-based instructions.

\begin{figure}[t]
    \centering
    \includegraphics[width=0.45\textwidth]{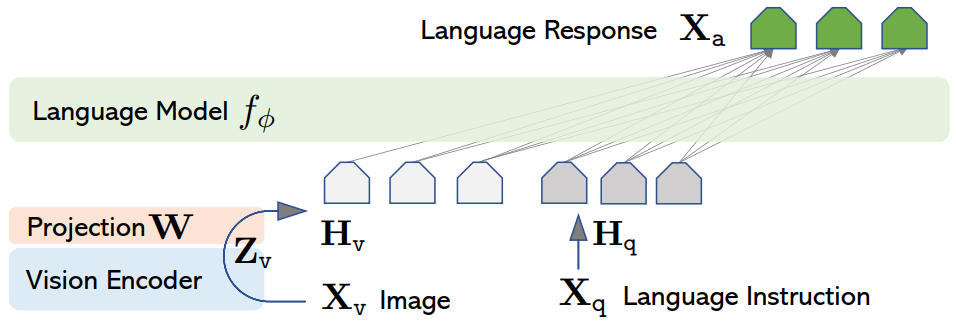}
    \caption{Illustration of the LLaVA~\cite{liu2023visual}. 
    Figure is from~\cite{liu2023visual}.}
    \label{fig:VA_LLaVA}
\end{figure}

LLaVA first introduces visual instruction tuning, extending the concept of instruction tuning to the language-image multimodal space. It presents the LLaVA model, which demonstrates impressive multimodal chat abilities and achieves state-of-the-art accuracy when fine-tuned on Science QA. Additionally, the paper constructs two evaluation benchmarks for visual instruction following and makes the model, data, and code publicly available, contributing to the research community.
The LLaVA architecture is shown in Figure~\ref{fig:VA_LLaVA}, which leverages a vision encoder, specifically the CLIP visual encoder ViT-L/14, to provide visual features for input images. These visual features are then processed by a language model termed Vicuna. The architecture consists of two stages: visual feature alignment and fine-tuning end-to-end, where the visual encoder weights are frozen, and the pre-trained weights of the projection layer and LLM in LLaVA are updated. This architecture enables the model to effectively leverage the capabilities of both the pre-trained language model and the visual encoder for general-purpose visual and language understanding.

Otter introduces the multimodal In-Context Instruction Tuning (MIMIC-IT) dataset, which consists of instruction-image-answer triplets and in-context examples. Otter itself is a multimodal model with in-context instruction tuning based on OpenFlamingo, showcasing improved instruction-following ability and in-context learning. Additionally, it optimizes OpenFlamingo's implementation, reducing the training requirements and integrating it into Hugging Face Transformers for easier use by researchers.

LLaVA-1.5 improves baselines for large multimodal models (LMMs) with visual instruction tuning. The authors demonstrate that simple modifications to the LLaVA framework, such as using an MLP cross-modal connector, incorporating academic task-related data, introducing response formatting prompts to balance short- and long-form VQA, scaling up the input image resolution, and including additional visual knowledge sources, result in stronger and more feasible baselines. These improvements lead to state-of-the-art performance across 11 benchmarks, using significantly less training data and compute resources compared to existing methods. The work provides a fully-reproducible and affordable baseline for future research in open-source LMMs.

SVIT introduces a large-scale dataset called SVIT, containing 4.2 million instruction tuning data generated by prompting GPT-4 with manual annotations of images. The dataset aims to enhance visual instruction tuning for multimodal models, leading to better performance in visual perception, reasoning, and planning tasks. The experiments demonstrate that training multimodal models on the SVIT dataset achieves superior performance compared to training on smaller datasets.

\begin{figure}[t]
    \centering
    \includegraphics[width=0.45\textwidth]{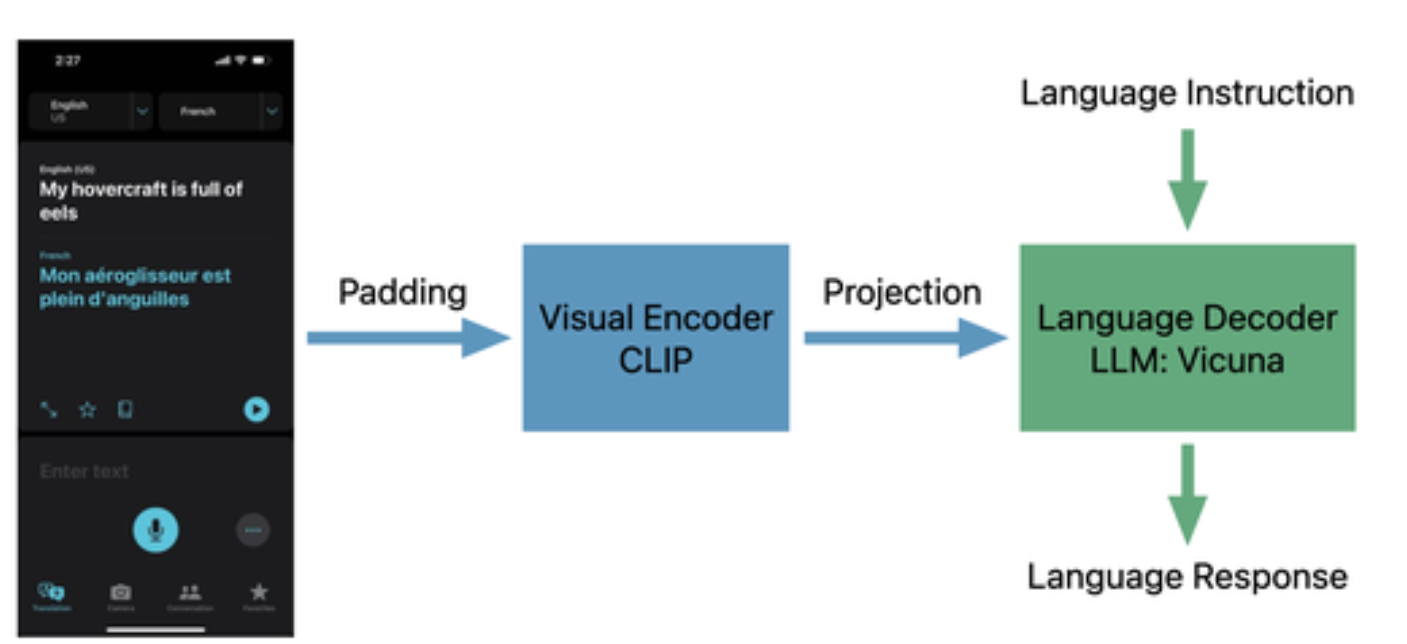}
    \caption{Illustration of ILuvUI~\cite{jiang2023iluvui}. 
    Figure is from~\cite{jiang2023iluvui}.}
    \label{fig:VA_ILuvUI}
\end{figure}

As shown in Figure~\ref{fig:VA_ILuvUI}, ILuvUI introduces a Vision-Language Model (VLM) specifically tailored for understanding and interacting with user interfaces (UIs). The model is trained using a dataset of image-instruction pairs generated from UI screenshots, and it demonstrates the ability to describe UI elements, provide contextual help, and plan multi-step interactions. The paper also benchmarks ILuvUI against existing models, highlighting its effectiveness in UI understanding tasks and its potential for enhancing UI accessibility. Additionally, the paper discusses the need for standardized benchmarks to evaluate VLMs in the context of UI tasks.

AssistGPT introduces a multimodal AI assistant system called AssistGPT, which integrates multiple models to handle complex visual tasks.
AssistGPT utilizes an interleaved language and code reasoning approach called Plan, Execute, Inspect, and Learn (PEIL). It consists of four core modules: Planner, Executor, Inspector, and Learner. The Planner controls the reasoning process, the Executor executes external tools, the Inspector manages input and intermediate results, and the Learner assesses system performance and records successful trials as in-context examples. 
The system showcases its capabilities in processing complex images and videos, understanding high-level queries, and handling flexible inputs, demonstrating its effectiveness beyond benchmark results.

StableLLaVA introduces a novel data collection methodology for enhancing visual instruction tuning in multimodal Large Language Models (LLMs). The proposed approach synthesizes both images and associated dialogues, addressing limitations encountered with benchmark datasets including noise and domain bias. 
The research showcases the flexibility of the pipeline by generating a large-scale dataset covering more than ten useful capabilities and demonstrates significant improvements in model performance across these capabilities.

\begin{figure}[t]
    \centering
    \includegraphics[width=0.45\textwidth]{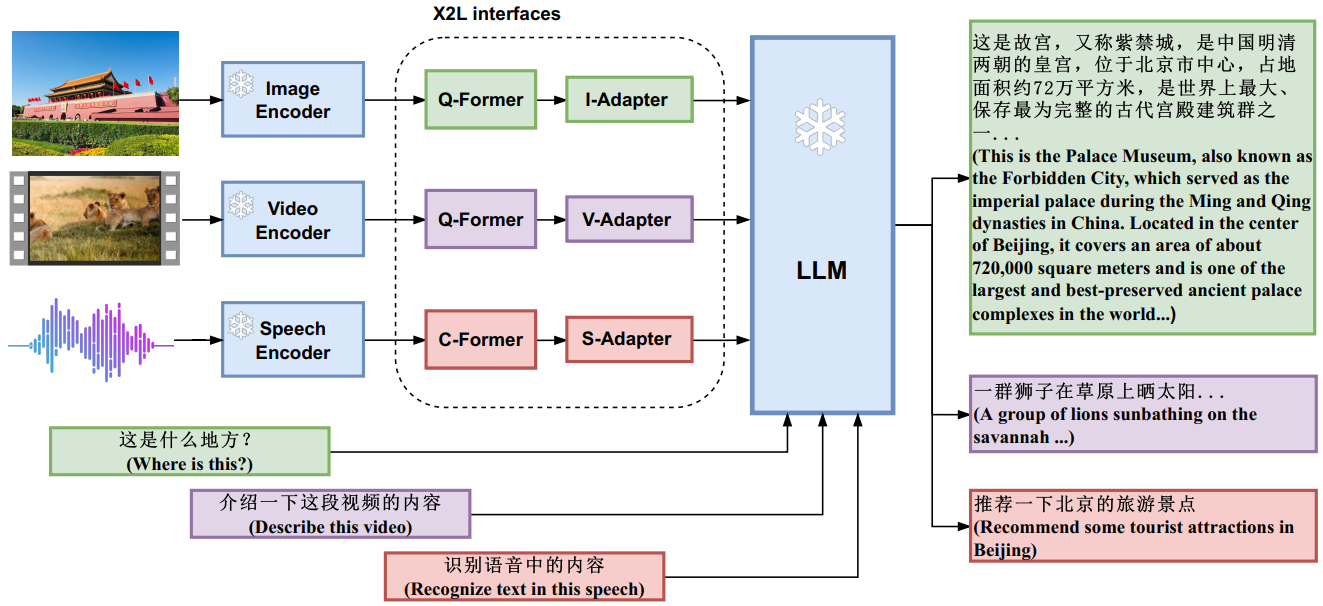}
    \caption{Illustration of X-LMM~\cite{chen2305bootstrapping}. 
    Figure is from~\cite{chen2305bootstrapping}.}
    \label{fig:VA_X-LLM}
\end{figure}

X-LLM is a Multimodal Large Language Model, which integrates multiple modalities such as images, speech, and videos into a large language model through X2L interfaces. The framework demonstrates impressive capabilities in tasks like visual spoken question answering and multimodal machine translation. Additionally, the paper introduces a three-stage training method for X-LLM and constructs a high-quality multimodal instruction dataset to further enhance its performance. Overall, the contributions include the development of a powerful multimodal language model and the exploration of joint multimodal instruction data to improve its capabilities.
As shown in Figure~\ref{fig:VA_X-LLM}, X-LLM's network architecture consists of multiple frozen single-modal encoders, including image, video, and speech encoders, aligned with a large language model (ChatGLM) through X2L interfaces. These interfaces, such as the image interface, video interface, and speech interface, convert multimodal information into foreign languages using Q-Formers and Adapter modules. The training process involves three stages, focusing on converting multimodal information, aligning representations with the LLM, and integrating multiple modalities. Overall, the architecture enables the integration of diverse modalities into a large language model for multimodal understanding and response generation.

PandaGPT is a model that integrates multimodal encoders from ImageBind and language models from Vicuna to perform instruction-following tasks across six modalities: image/video, text, audio, depth, thermal, and IMU. It demonstrates the ability to connect information from different modalities and compose their semantics naturally, enabling tasks such as image description generation, story writing inspired by videos, and answering questions about audios. PandaGPT's training on aligned image-text pairs allows it to display emergent cross-modal capabilities for data other than image and text, paving the way for holistic understanding of inputs across different modalities.

LAMM introduces the Language-Assisted multimodal (LAMM) dataset, framework, and benchmark, aiming to facilitate the training and evaluation of multimodal large language models (MLLMs). The main contributions include the comprehensive dataset and benchmark covering a wide range of vision tasks for 2D and 3D vision, a detailed methodology for constructing multimodal instruction tuning datasets, and a primary MLLM training framework optimized for modality extension. Additionally, the paper provides baseline models, extensive experimental observations, and analysis to accelerate future research in the field of multimodal language models.

LLaVAR is an enhanced visual instruction-tuned model for text-rich image understanding. It also collects noisy and high-quality instruction-following data to augment visual instruction tuning, significantly improving text understanding within images. The model's enhanced capability allows for end-to-end interactions based on various forms of online content combining text and images, and the authors open-source the training and evaluation data together with the model checkpoints.

Qwen-VL is a versatile vision-language model that integrates image understanding, text reading, localization, and multi-round dialogue capabilities. It addresses the limitations of large language models by incorporating visual signals and demonstrates superior performance in tasks such as image captioning, visual question answering, refer expression comprehension, and text-oriented tasks. The model's multi-task pre-training data and its ability to handle diverse style tasks make it a valuable contribution to multimodal research.

\begin{figure}[t]
    \centering
    \includegraphics[width=0.45\textwidth]{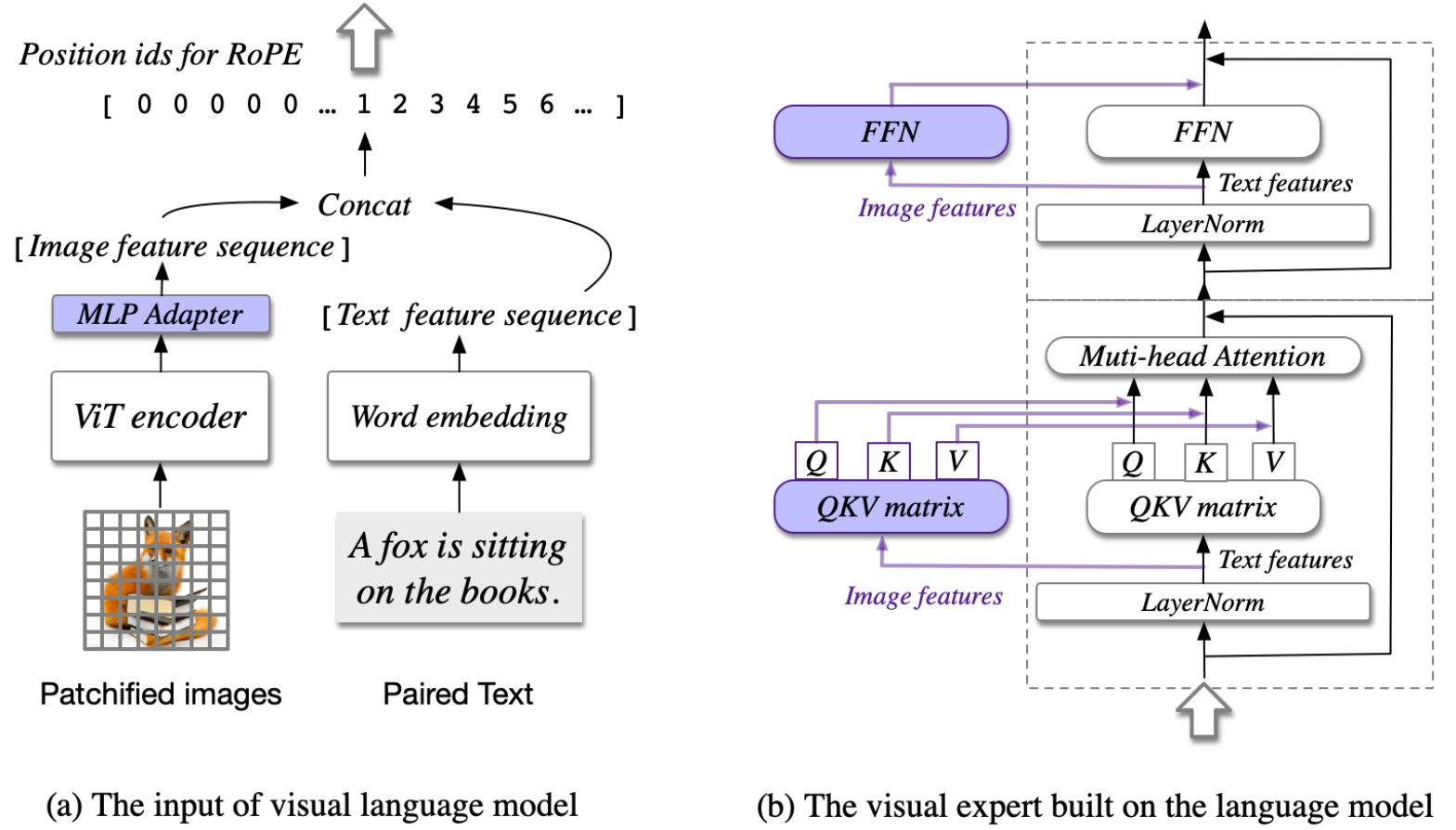}
    \caption{Illustration of CogVLM~\cite{wang2023cogvlm}. 
    Figure is from~\cite{wang2023cogvlm}.}
    \label{fig:VA_CogVLM}
\end{figure}

CogVLM is a powerful open-source visual language model that excels in a broad range of multimodal tasks such as image captioning, visual question answering, and visual grounding. The model's superior performance and robust generalization are rigorously validated through quantitative evaluations on various benchmarks, showcasing its remarkable capability and robustness. Additionally, the paper presents qualitative examples generated by CogVLM, demonstrating its effectiveness in real-world applications. 
As shown in Figure~\ref{fig:VA_CogVLM}, CogVLM comprises four fundamental components: a vision transformer (ViT) encoder, an MLP adapter, a pretrained large language model (GPT), and a visual expert module. The ViT encoder processes the image, the MLP adapter maps the output of ViT into the same space as the text features, and the pretrained large language model forms the base for further training. The visual expert module is added to each layer to enable deep visual-language feature alignment, consisting of a QKV matrix and an MLP in each layer. This architecture allows for deep fusion of vision and language information, resulting in state-of-the-art performance on multimodal tasks.

\begin{figure}[t]
    \centering
    \includegraphics[width=0.45\textwidth]{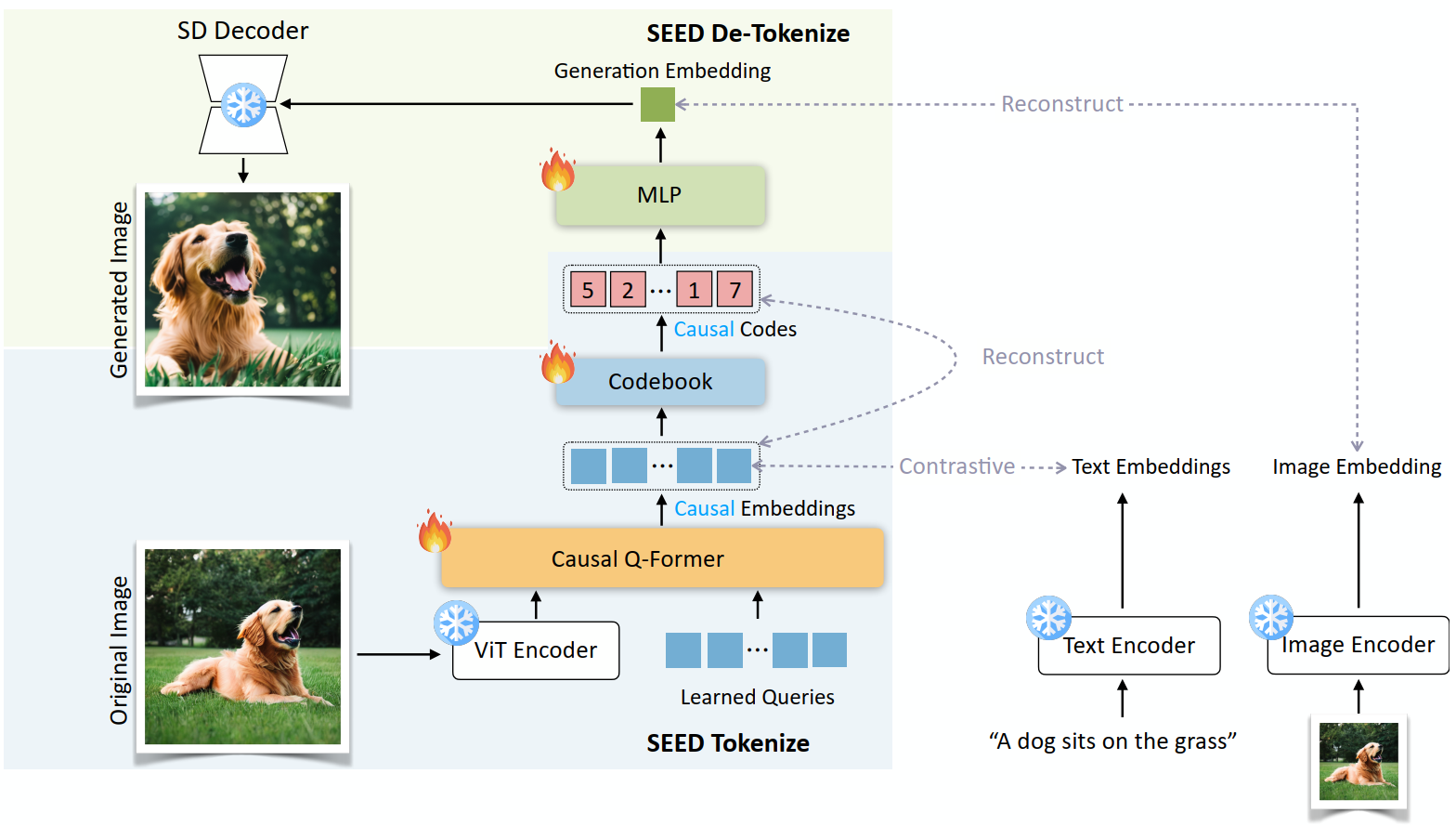}
    \caption{Illustration of SEED-LLaMA~\cite{ge2023making}. 
    Figure is from~\cite{ge2023making}.}
    \label{fig:VA_SEED_LLaMA}
\end{figure}

SEED-LLaMA introduces SEED, a discrete image tokenizer designed to enable Large Language Models (LLMs) to process and generate text and images interchangeably. SEED-LLaMA, a multimodal AI assistant, is produced by pretraining and instruction tuning on interleaved visual and textual data with SEED tokenizer. It demonstrates impressive performance in multimodal comprehension and generation tasks, as well as compositional emergent abilities such as multi-turn in-context multimodal generation. The key contribution lies in enabling LLMs to perform scalable multimodal autoregression under its original training recipe, thus advancing the potential of multimodality in AI.
As shown in Figure~\ref{fig:VA_SEED_LLaMA}, SEED is a discrete image tokenizer that converts 2D raster-ordered features into a sequence of causal semantic embeddings, which are further discretized into quantized visual codes with causal dependency.  These visual codes are then decoded into generation embeddings aligned with the latent space of a pre-trained model, allowing for the generation of realistic images.  SEED enables Large Language Models to perform scalable multimodal autoregression on interleaved visual and textual data, thus unifying multimodal comprehension and generation tasks within a single framework.

OtterHD introduces OtterHD-8B model, which addresses the limitations of fixed-resolution inputs in Large Multimodal Models (LMMs). It leverages the Fuyu-8B architecture to process images of varying resolutions, demonstrating enhanced performance in discerning fine details in complex scenes. The model's contribution lies in its ability to effectively handle high-resolution images and its performance on the MagnifierBench benchmark, highlighting the importance of resolution flexibility in contemporary LMMs.

\begin{figure}[t]
    \centering
    \includegraphics[width=0.45\textwidth]{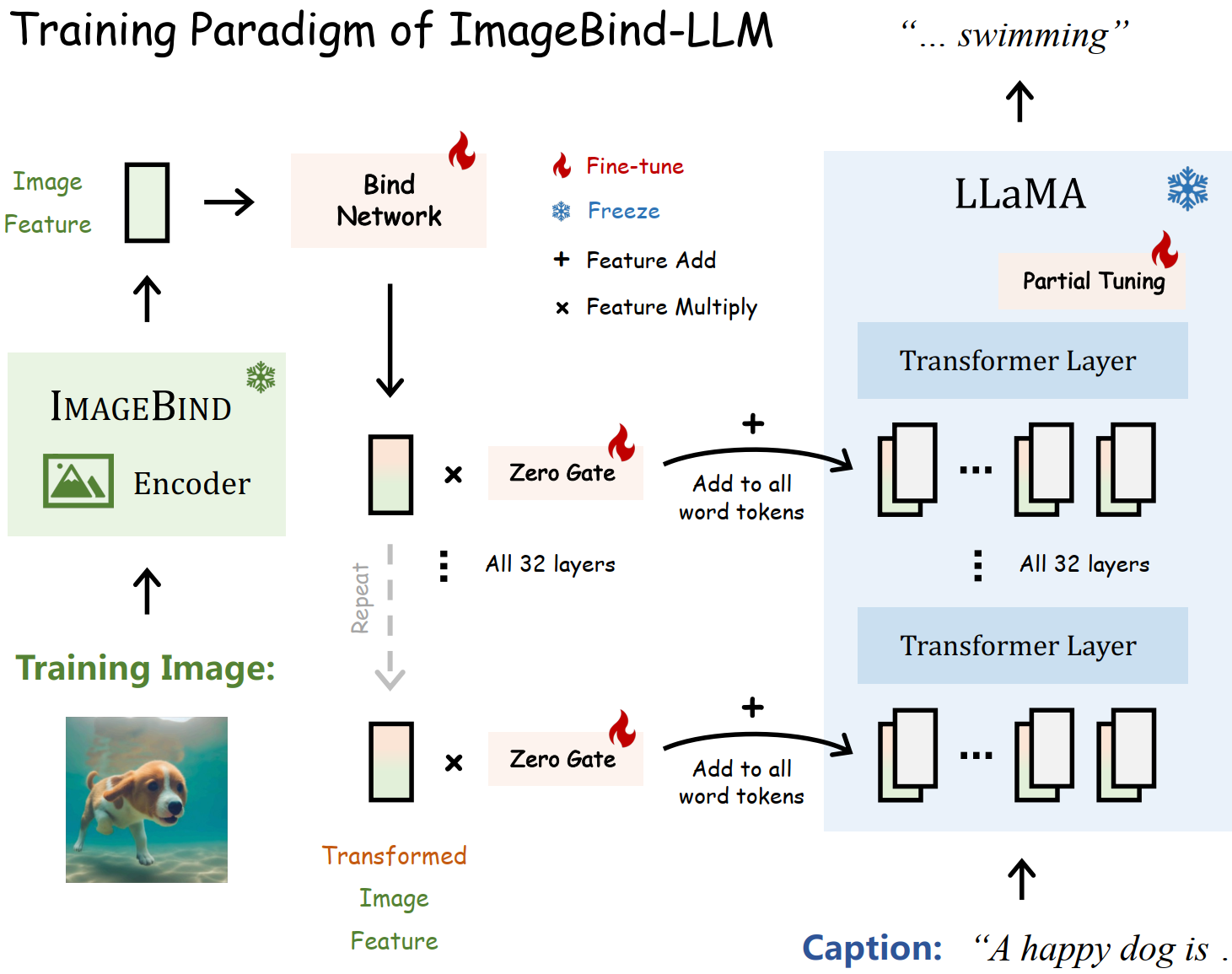}
    \caption{The training paradigm of of ImageBind-LLM~\cite{han2023imagebind}. 
    Figure is from~\cite{han2023imagebind}.}
    \label{fig:VA_Imagebind-llm}
\end{figure}

ImageBind-LLM is a model that enhances multimodality instruction tuning and cache-enhanced inference. It revisits prior works such as ImageBind and LLaMA-Adapter, and evaluates the proposed ImageBind-LLM on a new benchmark, MME. The model demonstrates strong performance in perception tasks and showcases its multimodal instruction capabilities through qualitative analysis. Overall, the paper contributes to the development of robust and versatile language models with enhanced multimodality understanding and performance.
As shown in Figure~\ref{fig:VA_Imagebind-llm}, the training paradigm of ImageBind-LLM involves a two-stage training pipeline. In the first stage, the model is pre-trained on large-scale image-caption data to learn image-conditioned response capacity. This stage involves aligning the joint embedding space of ImageBind with LLaMA using a learnable bind network and an attention-free zero-initialized mechanism for visual knowledge injection. In the second stage, the model is fine-tuned on a mixture of language instruction data and visual instruction data to equip it with both language and visual instruction-following abilities. Additionally, a training-free visual cache model is proposed to mitigate the modality discrepancy between training and inference.

\subsection{Instruction-based Video Learning}
Instruction-based video learning improves the performance of the video comprehensionby enabling efficient adaptation of large language models (LLMs) to effectively devise, process, and integrate video-centric instruction tuning datasets, leading to enhanced spatiotemporal reasoning and causal relationship inferencing ability and accurate responses in visual question answering tasks.  

\subsubsection{Visual Assistant}

EmbodiedGPT is an end-to-end multimodal foundation model for embodied AI with a "chain-of-thought" capability, enabling embodied agents to interact with the physical world more naturally. It also develops two datasets, EgoCOT and EgoVQA, and proposes a cost-effective training approach for extracting task-relevant features from planning queries. The approach demonstrates state-of-the-art or comparable performance on multiple embodied tasks, including embodied control, planning, video captioning, and video QA, outperforming existing models on benchmark tasks.

ChatBridge is a novel multimodal language model that leverages large language models to bridge the gap between various modalities. It proposes a two-stage training approach to align different modalities with language and introduces a new multimodal instruction tuning dataset called MULTIS.

\begin{figure}[t]
    \centering
    \includegraphics[width=0.45\textwidth]{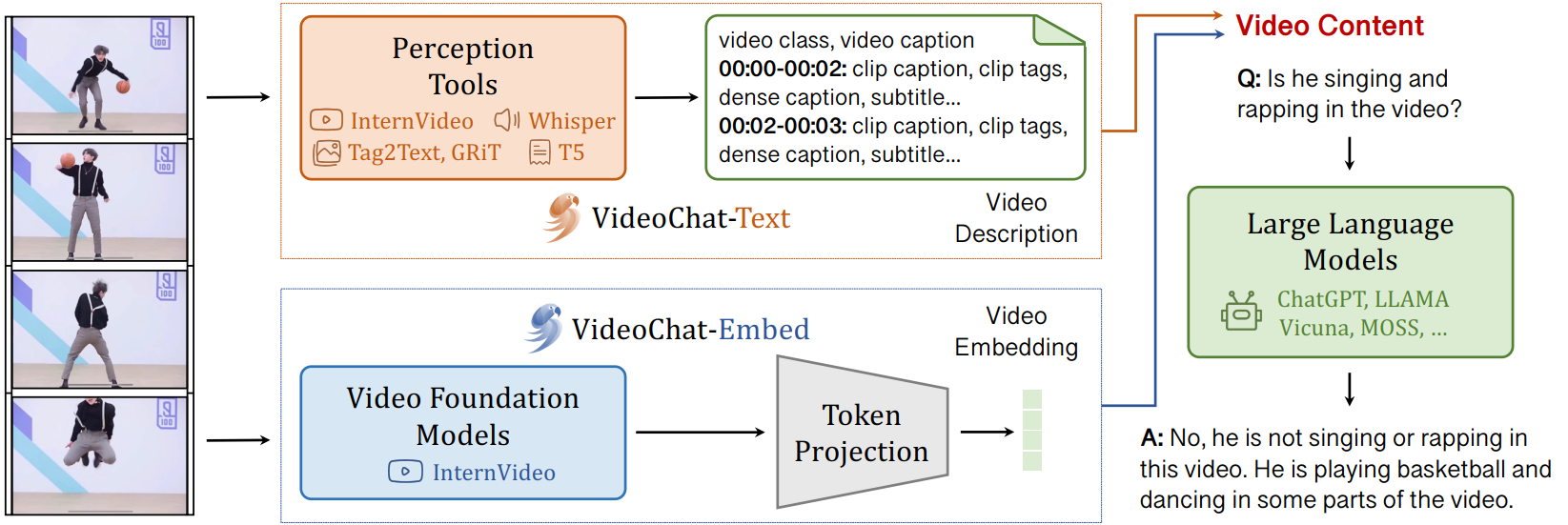}
    \caption{Illustration of VideoChat~\cite{li2023videochat}. 
    Figure is from~\cite{li2023videochat}.}
    \label{fig:VA_VideoChat}
\end{figure}

VideoChat is a chat-centric video understanding system that integrates video foundation models and large language models. It proposes a video-centric instruction dataset emphasizing spatiotemporal reasoning and causal relationships, providing a valuable asset for training chat-centric video understanding systems. It also presents qualitative experiments showcasing the system's potential across various video applications and sets a standard for future research in the field of video understanding.
As shown in Figure~\ref{fig:VA_VideoChat}, the framework of VideoChat consists of two main components: VideoChat-Text and VideoChat-Embed. VideoChat-Text textualizes videos in stream by converting visual data into textual format using various vision models and prompts, allowing a pretrained large language model to address user-specified tasks based on the video text descriptions. On the other hand, VideoChat-Embed encodes videos as embeddings and combines video and language foundation models with a Video-Language Token Interface (VLTF) to optimize cross-modality, enabling the model to effectively communicate with users through a large language model.

\begin{figure}[t]
    \centering
    \includegraphics[width=0.45\textwidth]{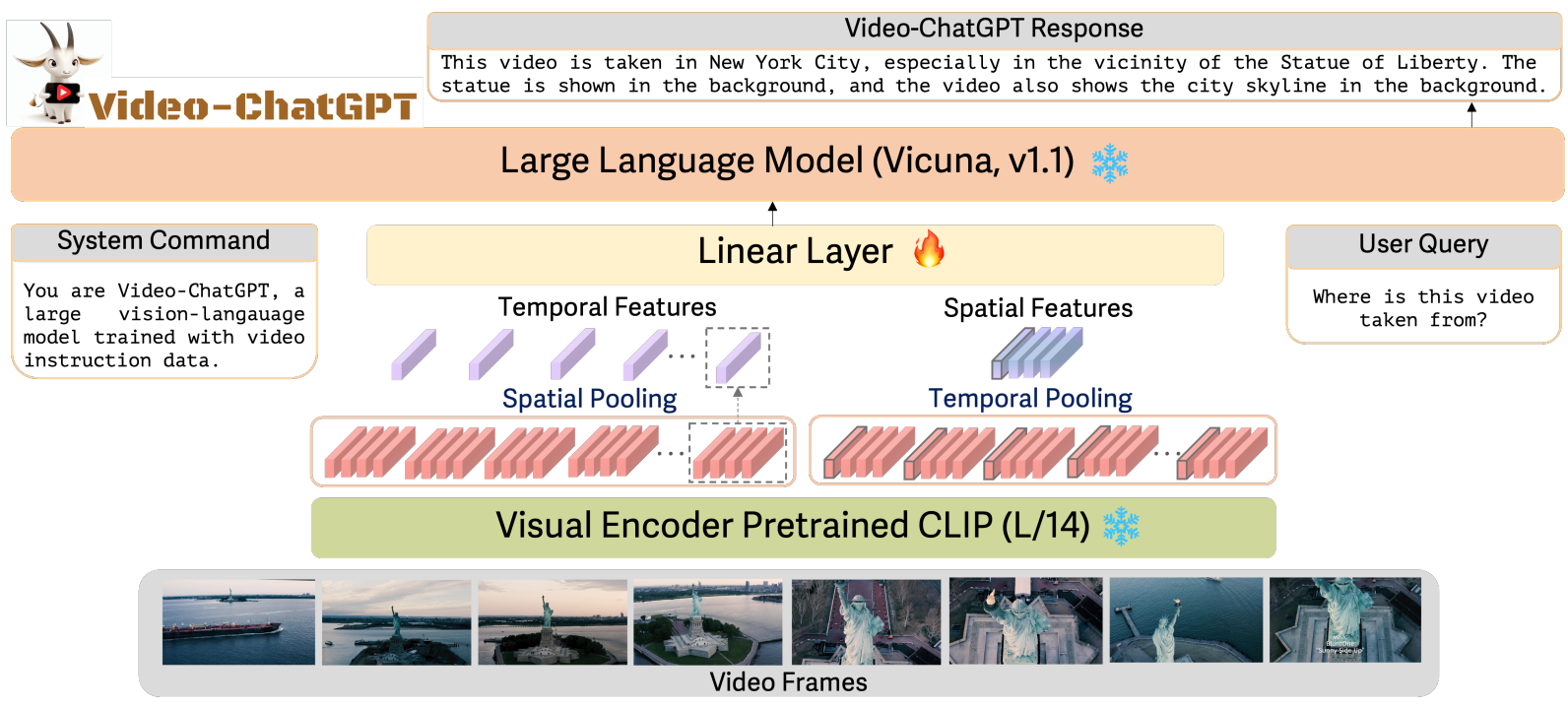}
    \caption{Illustration of Video-ChatGPT~\cite{maaz2023video}. 
    Figure is from~\cite{maaz2023video}.}
    \label{fig:VA_Video-ChatGPT}
\end{figure}

Video-ChatGPT is a multimodal model that merges a pretrained visual encoder with a Large Language Model (LLM) to understand and generate detailed conversations about videos. It presents a new dataset of 100,000 video-instruction pairs and develops a quantitative evaluation framework for video-based dialogue models. The model's architecture, training process, and evaluation results are thoroughly described, showcasing its competence in video understanding and conversation generation. Additionally, the paper proposes a novel human-assisted and semi-automatic annotation framework for generating high-quality video instruction data.
As shown in Figure~\ref{fig:VA_Video-ChatGPT}, the architecture of Video-ChatGPT leverages a pretrained visual encoder, CLIP ViT-L/14, to extract both spatial and temporal video features. These features are then projected into the input space of a Large Language Model (LLM) using a learnable linear layer. The resulting model is capable of understanding and generating detailed conversations about videos, showcasing proficiency in video reasoning, creativity, spatial understanding, action recognition, and temporal understanding.

\begin{figure}[t]
    \centering
    \includegraphics[width=0.45\textwidth]{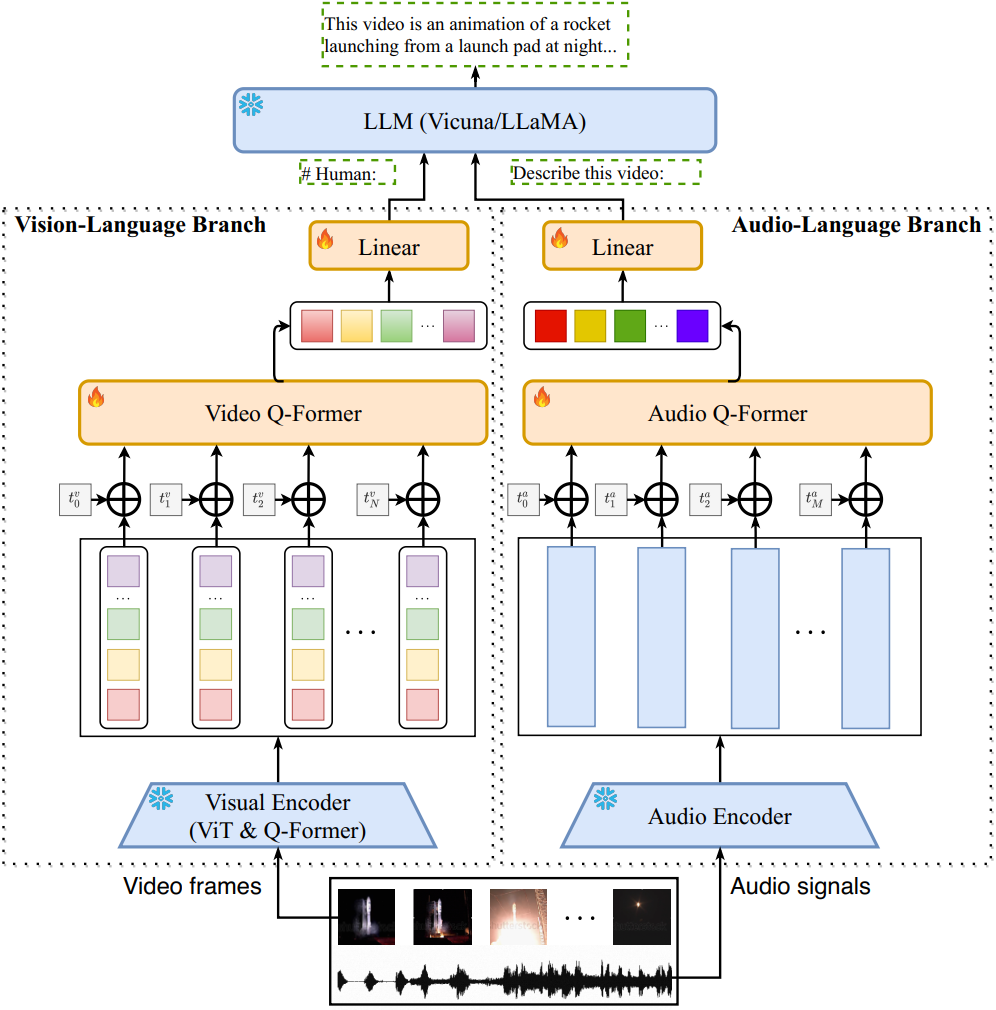}
    \caption{Illustration of Video-LLaMA~\cite{zhang2023video}. 
    Figure is from~\cite{zhang2023video}.}
    \label{fig:VA_Video-LLaMA}
\end{figure}

Video-LLaMA focus on empowering Large Language Models (LLMs) with the capability to understand both visual and auditory content in videos. It aims to enable LLMs to comprehend and generate meaningful responses grounded in the visual and auditory information presented in the videos.
As shown in Figure~\ref{fig:VA_Video-LLaMA}, the architecture of Video-LLaMA consists of two main branches: the Vision-Language Branch and the Audio-Language Branch. The Vision-Language Branch includes a frozen pre-trained image encoder, a position embedding layer, a video Q-former, and a linear layer to transform video representations into the same dimension as the text embeddings of LLMs. The Audio-Language Branch includes a pre-trained audio encoder, a position embedding layer, an audio Q-former, and a linear layer to map audio features to the embedding space of LLMs. These branches enable Video-LLaMA to process both visual and auditory content within a single framework.

Valley aims to develop a multimodal foundation model capable of comprehending video, image, and language within a general framework. Valley aims to function as a highly effective video assistant that can make complex video understanding scenarios easy. It focuses on creating a seamless interaction between humans and machines, enabling natural and intuitive conversations while engaging in various tasks related to video understanding.

MACAW-LLM is a novel architecture for multimodal language modeling, integrating image, audio, video, and text data. 
It also introduces the MACAW-LLM instruction dataset, which covers diverse instructional tasks and modalities. MACAW-LLM involves a simplified one-step instruction fine-tuning process, a multimodal dataset for instruction-tuned language models, and an architecture that aligns multimodal features with textual features for generating output sequences.

\subsection{Instruction-based 3D Vision Learning}

3D vision tasks involve the analysis and interpretation of visual data to reconstruct and understand the three-dimensional structure of the environment, including depth estimation, 3D reconstruction, object recognition, and scene comprehension. These tasks enable machines to interact with the physical world in a more human-like manner, supporting applications in robotics, augmented reality, autonomous vehicles, and more.
The increasing demand for natural language interactions with 3D content includes scenarios such as verbally commanding robots to manipulate objects and interactively creating and editing 3D content through natural language. Existing efforts with 2D images face challenges such as depth ambiguity and viewpoint dependency, making it essential to empower LLMs to comprehend 3D structures accurately and effectively. 
This capability opens up new avenues for natural language interactions with 3D objects and environments.

\subsubsection{Visual Assistant}
\begin{figure}[t]
    \centering
    \includegraphics[width=0.45\textwidth]{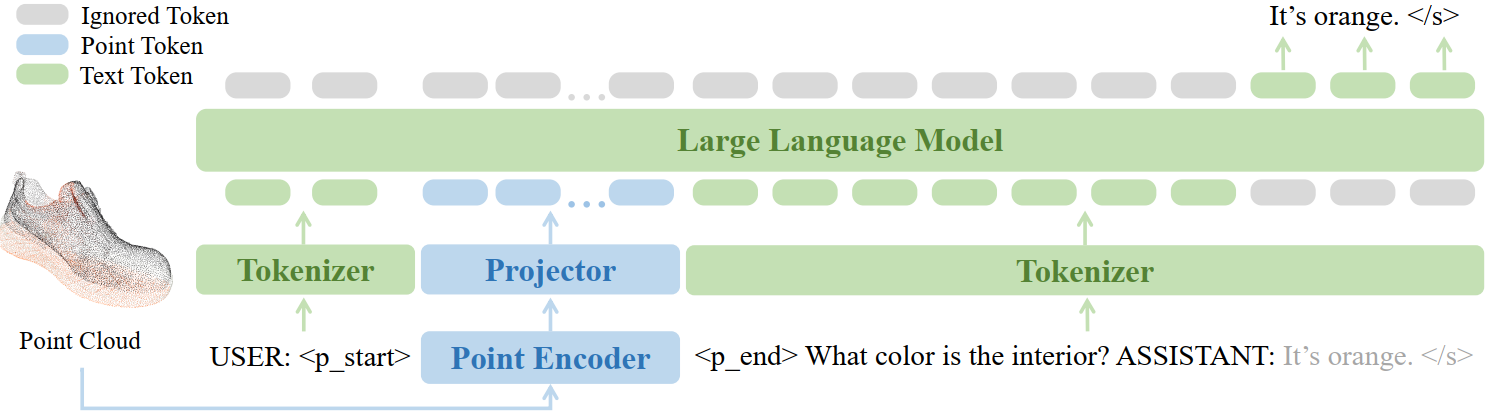}
    \caption{Illustration of PointLLM~\cite{xu2023pointllm}. 
    Figure is from~\cite{xu2023pointllm}.}
    \label{fig:3DVis_PointLLM}
\end{figure}

PointLLM is a large language model specifically designed for understanding 3D object point clouds. It provides a comprehensive evaluation suite, including benchmarks and a large-scale dataset, which will be open-source for community use. It also addresses the limitations of traditional metrics in evaluating language models and emphasizes the need for more comprehensive and reliable measures. Additionally, it explores the potential of PointLLM in tasks such as text-to-3D generation, demonstrating its capacity to generate detailed and accurate captions for 3D models. 
As shown in Figure~\ref{fig:3DVis_PointLLM}, the architecture of PointLLM consists of three main components: a pre-trained point cloud encoder, a large language model (LLM) backbone, and a multimodal projection layer. The point cloud encoder encodes point clouds into tokens, which are then combined with text tokens and fed into the LLM backbone. The LLM backbone, based on transformer architecture, processes the combined sequence of tokens to generate responses. The model is trained using a two-stage strategy, aligning the latent spaces between the encoder and the LLM, followed by instruction-based fine-tuning.

\begin{figure}[t]
    \centering
    \includegraphics[width=0.45\textwidth]{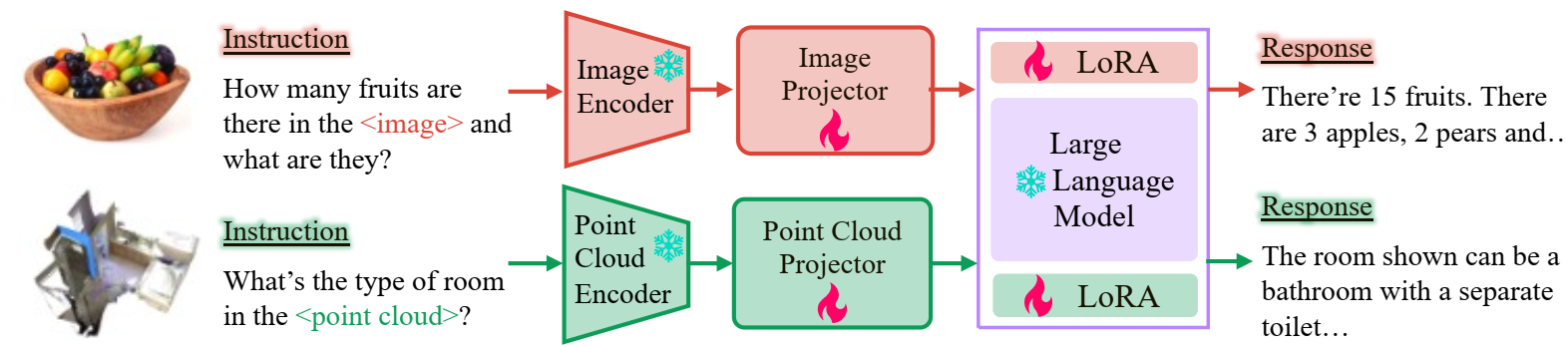}
    \caption{Illustration of LAMM~\cite{yin2023lamm}. 
    Figure is from~\cite{yin2023lamm}.}
    \label{fig:3DVis_LAMM}
\end{figure}

LAMM is an open-source endeavor focused on multimodal Large Language Models (MLLMs). The main focus of LAMM include the introduction of a comprehensive dataset and benchmark covering a wide range of vision tasks for 2D and 3D vision, the methodology for constructing multimodal instruction tuning datasets and benchmarks for MLLMs, and the provision of a primary but potential MLLM training framework optimized for modality extension. Additionally,it provides baseline models, extensive experimental observations, and analysis to accelerate future research in the field of MLLMs.
As shown in Figure~\ref{fig:3DVis_LAMM}, the framework of the multimodality language model (MLLM) in LAMM involves encoding each modality, such as image or point cloud, using corresponding pre-trained encoders. The encoded features are then projected to the same feature space as the text embeddings by a trainable projection layer. Instructions are tokenized and concatenated with vision and text tokens to feed into the MLLM model. The model is trained in a one-stage end-to-end fashion with trainable projection layers and LoRA modules, allowing for the extension to cover more modalities and tasks, such as video understanding and image synthesis.

\subsection{Instruction-based Medical Vision Learning}

\subsubsection{Medical Visual Question Answering}
Medical Visual Question Answering (MedVQA) tasks involve answering natural language questions about medical visual content, where the goal is to aid in the interpretation of medical images with vital clinic-relevant information.

For example, PMC-VQA introduces a generative model, MedVInT, for Medical Visual Question Answering (MedVQA) and establishes a scalable pipeline to construct a large-scale MedVQA dataset that covers various modalities and diseases. Additionally, it proposes a more challenging benchmark for evaluating VQA methods in the medical domain.

\subsubsection{Visual Assistant}

Medical visual assistant is a vision-language conversational assistant specifically designed for biomedical applications, which is trained to understand and converse about biomedical images, providing open-ended responses to inquiries about the content of biomedical images. 
Instruction-based Medical Vision Learning aims to transfer a general purposed multimodal large language model as medical visual assistant by carefully captured large-scale medical visual instruction dataset as well as specifically designed visual instruction tuning methods. 
The medical visual assistant is capable of following diverse instructions and completing tasks in a conversational manner, making it a valuable tool for biomedical visual question answering and providing informed advice in biomedical-related fields.

\begin{figure}[t]
    \centering
    \includegraphics[width=0.45\textwidth]{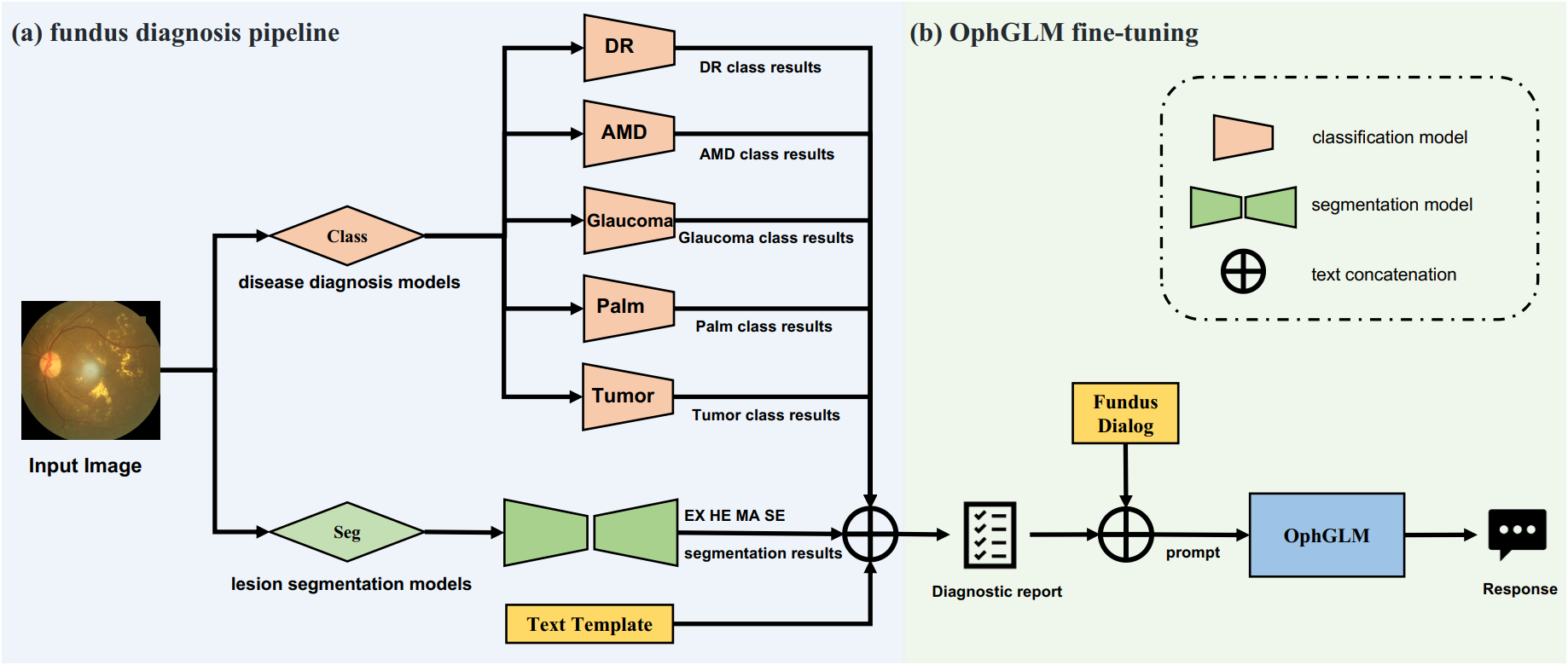}
    \caption{Illustration of OphGLM~\cite{gao2023ophglm}. 
    Figure is from~\cite{gao2023ophglm}.}
    \label{fig:MDE_VA_OphGLM}
\end{figure}

OphGLM is an ophthalmology large language-and-vision assistant, which integrates visual models with large language models in ophthalmology. 
It constructs a fine-tuned dataset for ophthalmic diseases, develop disease diagnosis models based on fundus images, and create a novel ophthalmology large language-and-vision assistant. 
The experimental results demonstrate the potential of OphGLM in clinical applications in ophthalmology.
As shown in Figure~\ref{fig:MDE_VA_OphGLM}, OphGLM consists of two main modules: the fundus diagnosis pipeline and the OphGLM pipeline. The fundus diagnosis pipeline includes disease diagnosis and lesion segmentation models based on fundus images. The OphGLM pipeline integrates the fundus image diagnostic report with the fundus dialogue, ultimately generating high-quality responses. This architecture allows OphGLM to accept fundus images as input and provide accurate and detailed medical information.

\subsection{Instruction-based Document Vision Learning}
The document understanding model is designed to automatically extract, analyze, and comprehend information from various types of digital documents. 
It aims to understand and interpret complex relationships between visual text and objects in diverse types of images, such as diagrams, documents, and webpages. 
Instruction tuning for document learning involves enhance the general purpose model comprehend and interpret visual information in various types of documents by designed visual instruction tuning strategy for visual-text understanding tasks and the captured datasets that facilitate multimodal document understanding. 

\begin{figure}[t]
    \centering
    \includegraphics[width=0.45\textwidth]{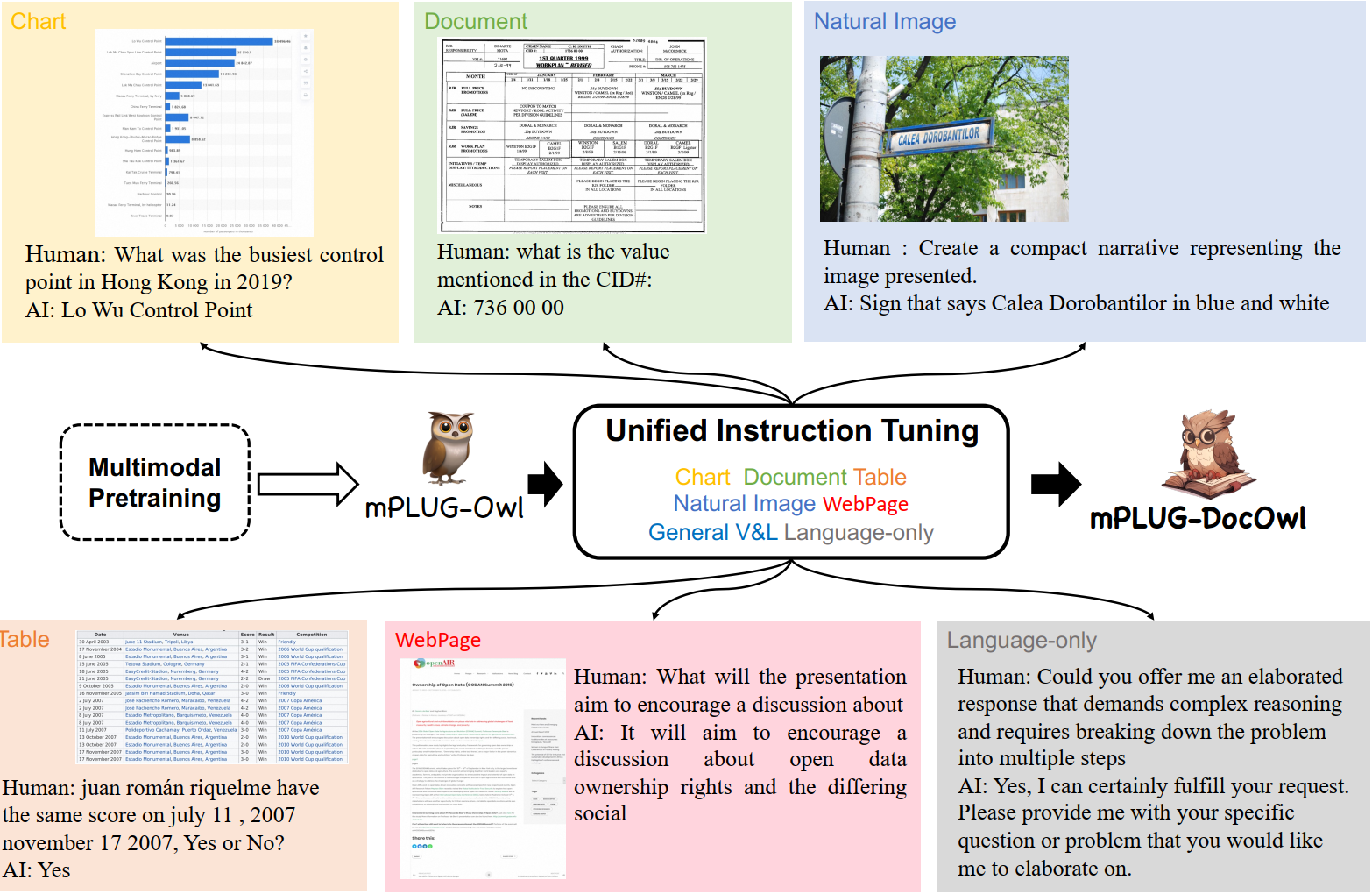}
    \caption{Illustration of mPLUG-DocOwl~\cite{ye2023mplugdoc}. 
    Figure is from~\cite{ye2023mplugdoc}.}
    \label{fig:DOC_mPLUG-DocOwl}
\end{figure}

\subsubsection{Visual Assistant}

mPLUG-DocOwl is a modularized Multimodal Large Language Model designed for OCR-free document understanding. It proposes a unified instruction tuning strategy to balance language-only, general vision-and-language, and document understanding. 
As shown in Figure~\ref{fig:DOC_mPLUG-DocOwl}, the instruction tuning paradigm of mPLUG-DocOwl involves the integration of diverse document understanding tasks into a unified format for training. It includes tasks such as Visual Question Answering, Information Extraction, Natural Language Inference, and Image Captioning. 
It outperforms existing multimodal models in document understanding and demonstrates strong generalization on various downstream tasks without specific fine-tuning. It also provides a carefully constructed evaluation set, LLMDoc, for assessing diverse document understanding capabilities, and conducts human evaluation to compare the performance of mPLUG-DocOwl with other models.

\begin{figure}[t]
    \centering
    \includegraphics[width=0.35\textwidth]{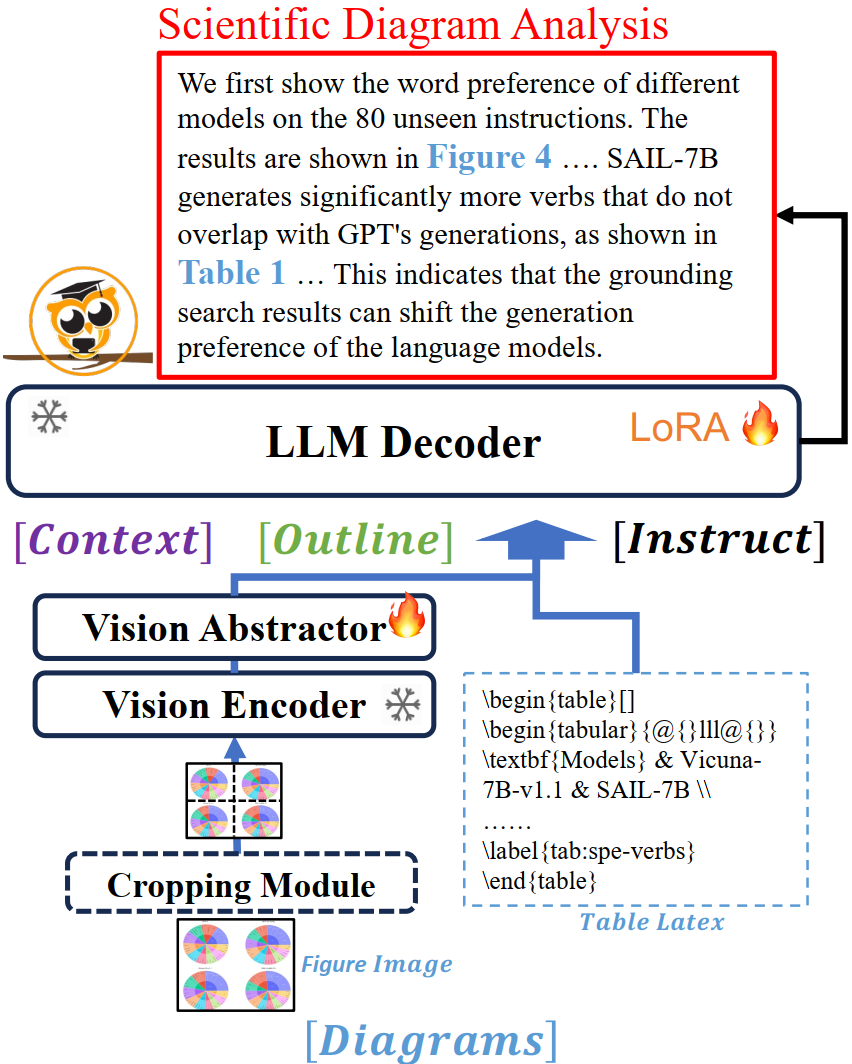}
    \caption{Illustration of mPLUG-PaperOwl~\cite{hu2023mplug}. 
    Figure is from~\cite{hu2023mplug}.}
    \label{fig:DOC_mPLUG-PaperOwl}
\end{figure}

mPLUG-PaperOwl focuses on strengthening the multimodal diagram analysis ability of Multimodal Large Language Models (LLMs) to assist in academic paper writing. It introduces the M-Paper dataset, which supports the joint comprehension of multiple scientific diagrams, including figures and tables in the format of images or Latex codes. It also proposes three multimodal tasks and a GPT-based metric to measure the paragraph analysis quality, and it validates the effectiveness of multimodal inputs and training strategies through comprehensive experiments.
As shown in Figure~\ref{fig:DOC_mPLUG-PaperOwl}, the overall architecture of mPLUG-PaperOwl follows a three-module framework, consisting of a vision encoder, a vision abstractor, and a Large Language Model as the language decoder. The vision encoder is fine-tuned to better learn how to filter useful visual diagram information for generating analysis, while the vision abstractor is fine-tuned to improve the model's ability to understand and describe diagrams. The model is trained on an ensemble of training data from three multimodal tasks to enhance its performance.

\section{Conclusion}

Visual instruction tuning fine-tunes a large vision model with language as task instructions, ultimately learning from a wide range of vision tasks described by language instructions a general-purpose multimodal model that can follow arbitrary instructions and thus solve arbitrary tasks specified by the user.
In this survey, we extensively review visual instruction tuning studies from
different perspectives, ranging from background to foundations, datasets, methodology, benchmarks, and current
research challenges and open research directions.
We summarize visual instruction tuning datasets, methods, and performances in tabular forms, aiming to offer a comprehensive overview on what accomplishments we have achieved, what challenges we currently faced, and what we could further achieved in visual instruction tuning research.

\ifCLASSOPTIONcaptionsoff
  \newpage
\fi

{\small
\bibliographystyle{IEEEtran}
\bibliography{ref}
}

\end{document}